\documentclass[11pt]{article}

\usepackage[preprint]{acl}

\usepackage{times}
\usepackage{latexsym}

\usepackage[T1]{fontenc}

\usepackage[utf8]{inputenc}

\usepackage{microtype}

\usepackage{inconsolata}

\usepackage{graphicx}

\usepackage{booktabs}    
\usepackage{multirow}    
\usepackage{makecell}    
\usepackage{adjustbox}   
\usepackage[table]{xcolor} 

\usepackage{graphicx}    

\usepackage{array}       

\usepackage{amsfonts}  
\usepackage{mathtools} 

\usepackage{bm}
\usepackage{hyperref}
\usepackage{url}
\usepackage{mathabx}
\usepackage{amsmath}
\usepackage{amssymb}
\usepackage{graphicx}      
\usepackage{subcaption}    
\usepackage[title]{appendix}
\usepackage{algorithm}
\usepackage{algorithmic}
\newcommand{\mycomment}[1]{\hspace{\fill} \textcolor{black}{// #1}}
\usepackage{amsmath,amsfonts}

\usepackage{tikz}
\usetikzlibrary{matrix,calc}
\usepackage{placeins}
\usepackage{caption}

\usepackage{multirow}
\usepackage{amsthm}
\usepackage{mathtools}

\usepackage{wrapfig}
\usepackage{xcolor}
\definecolor{customblue}{HTML}{4A69BD}
\definecolor{customred}{HTML}{D40202}
\definecolor{customgreen}{HTML}{22AB2F}
\usepackage{fontawesome5}   
\theoremstyle{plain}
\newtheorem{theorem}{Theorem}[section]
\newtheorem{proposition}[theorem]{Proposition}
\newtheorem{lemma}[theorem]{Lemma}

\theoremstyle{definition}
\newtheorem{definition}[theorem]{Definition}
\newtheorem{assumption}[theorem]{Assumption}
\newtheorem{remark}[theorem]{Remark}

\newcommand{\ourmethod}{\text{VERL}}
\newcommand{\ourER}{\text{ER}}
\newcommand{\ourV}{\text{ERV}}
\newcommand{\ourAcc}{\text{ERA}}

\definecolor{positive}{RGB}{25, 140, 25}
\definecolor{negative}{RGB}{200, 30, 45}
\definecolor{neutral}{RGB}{128, 128, 128}  

\newcommand{\pos}[1]{\textcolor{positive}{+#1}}
\newcommand{\negv}[1]{\textcolor{negative}{$-$#1}}
\newcommand{\zero}{\textcolor{neutral}{+0.0}}

\newcommand{\sectionwithouttoc}[1]{%
  \refstepcounter{section}%
  \sectionmark{#1}%
  \section*{\thesection\quad #1}%
}

\newcommand{\subsectionwithouttoc}[1]{%
  \refstepcounter{subsection}%
  \subsectionmark{#1}%
  \subsection*{\thesubsection\quad #1}%
}

\usepackage{dblfloatfix} 

\usepackage[dvipsnames]{xcolor}

\definecolor{FigQuestion}{HTML}{B36A2E}  
\definecolor{FigGT}{HTML}{2E62B8}        
\definecolor{FigVanilla}{gray}{0.35}    
\definecolor{FigOurs}{HTML}{8B1E3F}     

\title{Semantic-Space Exploration and Exploitation in RLVR for LLM Reasoning}

\renewcommand{\thefootnote}{\fnsymbol{footnote}}

\renewcommand{\thefootnote}{\fnsymbol{footnote}}

\author{
{\bfseries
Fanding Huang$^{1}$\footnotemark[1],
Guanbo Huang$^{1}$\footnotemark[1],
Xiao Fan$^{1}$,
Yi He$^{1}$,
Xiao Liang$^{2}$,
Xiao Chen$^{1}$}
\\
{\bfseries
Qinting Jiang$^{1}$,
Faisal Nadeem Khan$^{1}$,
Jingyan Jiang$^{3}$\footnotemark[2],
Zhi Wang$^{1}$\footnotemark[2]}
\\
$^{1}$Tsinghua Shenzhen International Graduate School, Tsinghua University \\
$^{2}$University of California, Los Angeles \quad
$^{3}$Shenzhen Technology University \\
\small \texttt{jiangjingyan@sztu.edu.cn, wangzhi@sz.tsinghua.edu.cn}
}

\begin{document}
\maketitle

\footnotetext[1]{Equal contribution: Fanding Huang and Guanbo Huang.}
\footnotetext[2]{Corresponding authors: Jingyan Jiang and Zhi Wang.}
\setcounter{footnote}{0}
\renewcommand{\thefootnote}{\arabic{footnote}}

\begin{abstract}
Reinforcement Learning with Verifiable Rewards (RLVR) for LLM reasoning is often framed as balancing exploration and exploitation in action space, typically operationalized with token-level proxies (e.g., output entropy or confidence). We argue that this apparent trade-off is largely a measurement artifact: token-level statistics reflect next-token uncertainty rather than how reasoning progresses over multi-token semantic structures. We therefore study exploration and exploitation in the hidden-state space of response trajectories. We use \textit{Effective Rank} (ER) to quantify representational exploration and introduce its temporal derivatives, \textit{Effective Rank Velocity} (ERV) and \textit{Effective Rank Acceleration} (ERA), to characterize exploitative refinement dynamics. Empirically and theoretically, ER and ERV exhibit near-zero correlation in semantic space, suggesting the two capacities can be improved simultaneously. Motivated by this, we propose \textit{Velocity-Exploiting Rank Learning} (VERL), which shapes the RLVR advantage with an auxiliary signal derived from ER/ERV and uses the more stable ERA as a meta-control variable to adaptively balance the incentives. Across multiple base models, RLVR algorithms, and reasoning benchmarks, VERL yields consistent improvements, including large gains on challenging tasks (e.g., 21.4\% in  Gaokao 2024). The code is available at \url{https://github.com/hf618/VERL}.
\end{abstract}

\sectionwithouttoc{Introduction}

\begin{figure}[!t] 
    \centering
    \includegraphics[width=0.47\textwidth]{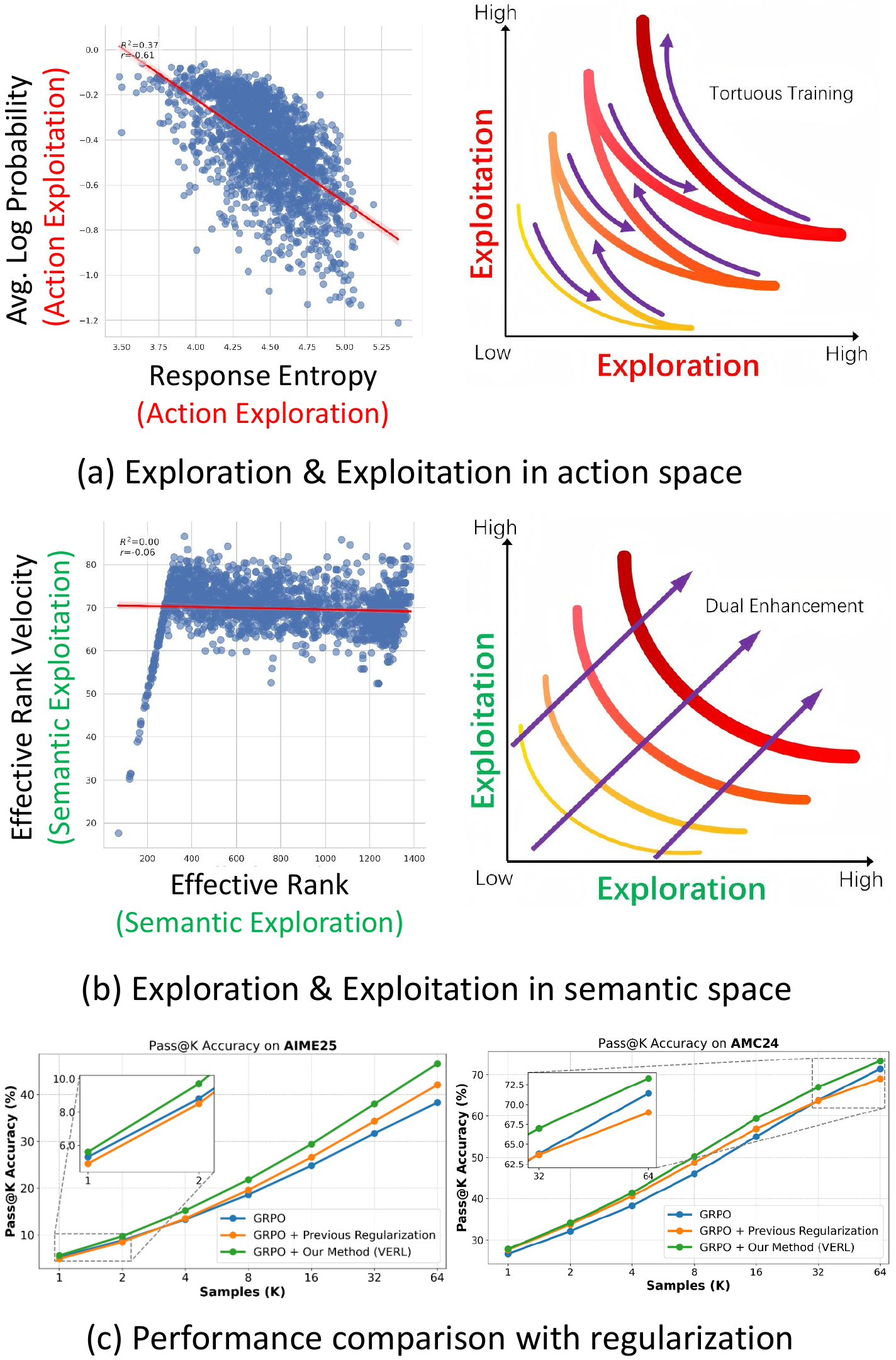}
    \caption{Comparative analysis of responses from DeepSeek-R1-Distill-Qwen-7B on simpleRL test set \citep{zeng2025simplerlzoo}. (a) Traditional metrics for exploitation \& exploration constrained by negative coupling, leading to meandering progress for both capabilities. (b) Our hidden-state metrics are structurally decoupled, capturing complementary aspects of semantic exploration and exploitation. (c) Training regularization with our metrics demonstrates stronger performance in both exploitation (small K) and exploration (large K).}
    \label{first_fig}
    \vspace{-4mm}
\end{figure}

Recent advancements in Reinforcement Learning with Verifiable Rewards (RLVR) have substantially improved the reasoning abilities of Large Language Models (LLMs). A dominant narrative in recent work~\citep{chen2025pass,yue2025does,deng2025trial,agarwal2025unreasonable} interprets this progress through the lens of balancing \textit{exploration} (seeking diverse reasoning paths) and \textit{exploitation} (refining the most promising known strategies). In practice, however, these notions are almost exclusively operationalized in the \emph{token-level action space}: exploration is associated with high-entropy next-token distributions, while exploitation is linked to high-confidence ones. These token-level proxies are convenient to compute, yet they primarily measure uncertainty about the \emph{next token}. As a result, the widely discussed ``exploration--exploitation trade-off'' (Fig.~\ref{first_fig}a) in reasoning may reflect the limitations of these proxies, instead of an inherent property of reasoning behavior.

This mismatch becomes more evident when distinguishing the \emph{action space} from the \emph{semantic space} in LLM reasoning. Reasoning progress is better viewed as transitions between \emph{semantic states}, traversing concepts, subgoals, and partial derivations that span many tokens~\citep{wei2022chain,yao2023tree}, while token sampling occurs in the action space. Token-level proxies (e.g., next-token entropy) can therefore misattribute local sampling uncertainty to semantic progress, leading to two failure modes: higher entropy may only amplify lexical variability without reaching new semantic directions, whereas lower entropy may collapse the trajectory too early into a familiar semantic region~\citep{fu2025deep,qiao2025concise,agarwal2025unreasonable}.

Prior work has shown that pre-trained LLM hidden states naturally encode rich semantic structure~\citep{huang25cosmic,jing2025lingualens} and reasoning dynamics~\citep{cheng2026reasoning,deng2025token}. These findings suggest that hidden states can serve as a continuous representation of the model's evolving semantic state. Thus, this motivates that exploration and exploitation should be defined over trajectories in semantic space (how broadly the representation spans semantic directions, and how it refines within a region), rather than over token distributions in action space. 
We therefore ask: \textit{Is the exploration--exploitation trade-off intrinsic to reasoning, or largely an artifact of measuring it in the token-level action space?}

To answer this, we step out of the action space and study exploration and exploitation directly in a \emph{semantic space} induced by transformer hidden-state representations. In this semantic space, we quantify \textit{representational exploration} by Effective Rank (\ourER{}), which measures how broadly a hidden-state trajectory spans semantic directions~\citep{valeriani2023geometry,matthews2024semantics,jing2025lingualens}. Following this intuition, we quantify \textit{representational exploitation} by the temporal evolution of this span: Effective Rank Velocity (\ourV{}) captures how semantic complexity is refined along the trajectory, and Effective Rank Acceleration (\ourAcc{}) captures whether this refinement is sustaining, accelerating, or saturating. Equipped with these semantic-trajectory measures, we uncover a striking empirical result: \ourER{} and \ourV{} exhibit near-zero correlation (Fig.~\ref{first_fig}b). This provides evidence that the commonly perceived exploration--exploitation trade-off is not intrinsic to LLM reasoning, but largely an artifact of measuring both notions in the token-level action space. Moreover, the two capacities can be \emph{decoupled and improved simultaneously} in semantic space, as shown in Fig.~\ref{first_fig}c.

Building on this insight, we propose \emph{Velocity-Exploiting Rank Learning} (\ourmethod{}), a plug-and-play advantage-shaping method that optimizes exploration and exploitation \emph{in semantic space}. \ourmethod{} augments the RL advantage with auxiliary incentives derived from \ourER{} (encouraging broader semantic directions to avoid premature collapse) and \ourV{} (reinforcing productive semantic refinement). Crucially, \ourmethod{} uses \ourAcc{} as a meta-control signal to adapt the strength of these incentives over training, as \ourAcc{} tracks the trend of semantic refinement (i.e., whether exploitation is accelerating or saturating) and  is theoretically more stable under short-term fluctuations (Sec.~\ref{Sec Metrics}). Across diverse base models, RL algorithms, and reasoning benchmarks, \ourmethod{} yields consistent improvements across evaluation settings, including up to 21.4\% absolute accuracy gain on the challenging Gaokao 2024 benchmark.

Our contributions are summarized as follows:
\begin{itemize}
    \item We recast exploration and exploitation from the token-level action space to a hidden-state semantic space, introducing \ourER{}, \ourV{}, and \ourAcc{} to measure semantic trajectories.
    \item We show that \ourER{} and \ourV{} are nearly uncorrelated in the semantic space. This evidence suggests that the commonly reported trade-off is largely induced by token-level measurements rather than being an intrinsic constraint of reasoning.
    \item We present \ourmethod{}, a plug-and-play method that improves both exploration and exploitation in the semantic space, and show consistent gains across models, RL algorithms, and benchmarks.
\end{itemize}

\sectionwithouttoc{Preliminaries}
\label{Preliminaries}
\subsectionwithouttoc{Problem Formulation and Notations}

We adopt an RLVR perspective for LLM reasoning. Given a prompt $x\sim\mathcal{P}_x$, an LLM policy $\pi_\theta(\cdot\mid x)$ generates a trajectory $y_{0:T}=(y_0,\ldots,y_T)$ with $T\le L_{\max}$. A scalar reward function $r(x,y)$ is used to evaluate the quality of the completed trajectory $y$ given the prompt $x$. We aim to learn the parameters $\theta$ by maximizing the expected reward, where $\phi$ represents the optimal solution:

\begin{equation}
\begin{aligned}
\phi = \underset{\theta}{\operatorname{argmax}}\,\mathbb{E}_{x \sim \mathcal{P}_x}\mathbb{E}_{y\sim\pi_\theta(\cdot|x)}\left[r(x,y)\right].
\end{aligned}
\end{equation}

A common view~\citep{chen2025pass,cheng2026reasoning,deng2025trial} is that optimizing this objective entails an exploration--exploitation trade-off: the policy must explore diverse paths to discover high-reward solutions while exploiting reliable strategies that consistently succeed.

\subsectionwithouttoc{Reinforcement Learning Baselines}
\textbf{Proximal Policy Optimization (PPO)}~\citep{schulman2017proximal} is an RL algorithm that seeks to maximize a clipped surrogate objective, which prevents excessively large policy updates that may destabilize training. The objective is defined as:

\begin{equation}
\label{eq:ppo}
\begin{aligned}
\mathcal{L}_{\mathrm{PPO}}(\theta)
&= \mathbb{E}_{x,y}\Bigl[\sum_{t=1}^{|y|}
\min\bigl(\rho_t(\theta)A_t,\;\bar\rho_t(\theta)A_t\bigr)\Bigr],\\
\bar\rho_t(\theta)
&:= \operatorname{clip}_{\,1-\epsilon_{\mathrm{low}}}^{\,1+\epsilon_{\mathrm{high}}}\!\bigl(\rho_t(\theta)\bigr),
\end{aligned}
\end{equation}
where $\rho_t(\theta)\vcentcolon=\frac{\pi_\theta(y_t|x,y_{<t})}{\pi_{\theta_{\mathrm{old}}}(y_t|x,y_{<t})}$ is the probability ratio between the current and old policies, and ${A}_t$ is the estimated advantage, often calculated using Generalized Advantage Estimation (GAE)~\citep{schulman2015high}, with clipping thresholds $\epsilon$.

\textbf{Group Relative Policy Optimization (GRPO)} from \citep{deepseek-math} computes a baseline directly from the rewards of multiple trajectories. For a given prompt, GRPO samples a group of $G$ responses, obtains their rewards $\{r_i\}_{i=1}^G$, and normalizes them to compute the group-relative advantage for each response:
\begin{equation}
A_i \vcentcolon=
\bigl(r_i-\operatorname{mean}(\{r_j\}_{j=1}^G)\bigr)
\big/
\operatorname{std}(\{r_j\}_{j=1}^G).
\end{equation}

This response-level advantage is then uniformly propagated to all tokens in the output sequence and optimized using the PPO-style objective in Eq.~\ref{eq:ppo}.

\subsectionwithouttoc{Hidden State Representations}
\label{Hidden State}
\textbf{Response hidden states.}
For an autoregressive response of length $T$, let $z_t \in \mathbb{R}^D$ denote the final hidden state at token $t$.
Stacking them yields $\mathbf{Z}=[z_1;\ldots;z_T]\in\mathbb{R}^{T\times D}$, which we view as a semantic trajectory. \textbf{Dataset hidden states.}
Given $N$ prompts, we summarize the $i$-th response by $\bar z_i=\frac{1}{T_i}\sum_{t=1}^{T_i} z_{i,t}$~\citep{skean2025layer}, and stack them into
$\bar{\mathbf{Z}}=[\bar z_1;\ldots;\bar z_N]\in\mathbb{R}^{N\times D}$ to represent the dataset-level semantic distribution.

\sectionwithouttoc{A Hidden-State Perspective on Representational Dynamics}
\label{Sec Metrics}

\paragraph{Assumption.}
Building on prior studies~\citep{skean2025layer,jing2025lingualens}, which show that linguistic and factual properties can be recovered from hidden states, our analysis assumes that semantic factors are approximately linearly decodable from hidden states. Our theoretical analysis is based on this assumption about representation.

\subsectionwithouttoc{Semantic Exploration: Effective Rank (\ourER{})}
\label{sec:static metric}
We use Effective Rank (\ourER{}) as a proxy for semantic diversity, as it increases when hidden states are more evenly distributed across non-redundant latent directions and decreases when they collapse onto a few dominant ones. Following \citep{roy2007effective}, we define the Effective Rank of a response hidden-state matrix $\mathbf{Z}\in\mathbb{R}^{T\times D}$ using its (non-padding) singular values $\{\sigma_j\}$.
Let $p_j=\sigma_j/\sum_k \sigma_k$, then

\begin{equation}
\ourER{}(\mathbf{Z}) \vcentcolon= \mathrm{erank}(\mathbf{Z})
= \exp\!\Bigl[-\,\textstyle\sum_j p_j \log p_j\Bigr].
\end{equation}

We interpret \ourER{} as \emph{representational breadth} in the hidden-state semantic space: larger \ourER{} indicates that the trajectory spans a broader set of semantic directions, while smaller \ourER{} suggests concentration in fewer directions.

\begin{theorem}
\label{the_erank}
Suppose we have a matrix of embeddings $\mathbf{Z}\in\mathbb{R}^{T \times D}$. Then the \ourER{} of $\mathbf{Z}$ is a lower bound of conventional rank of $\mathbf{Z}$:
\begin{equation}
1 \le \mathrm{erank}(\mathbf{Z}) \le \mathrm{rank}(\mathbf{Z}) \le \min\{T,D\}.
\end{equation}
\end{theorem}

\begin{remark}
While rank is discrete and blind to spectral concentration, \ourER{} is a continuous effective-dimensionality measure that grows as the singular spectrum flattens. Details are in Appendix.~\ref{app:er_exploration}.
\end{remark}

\begin{figure*}[!t]

    \centering 

    \includegraphics[width=\textwidth]{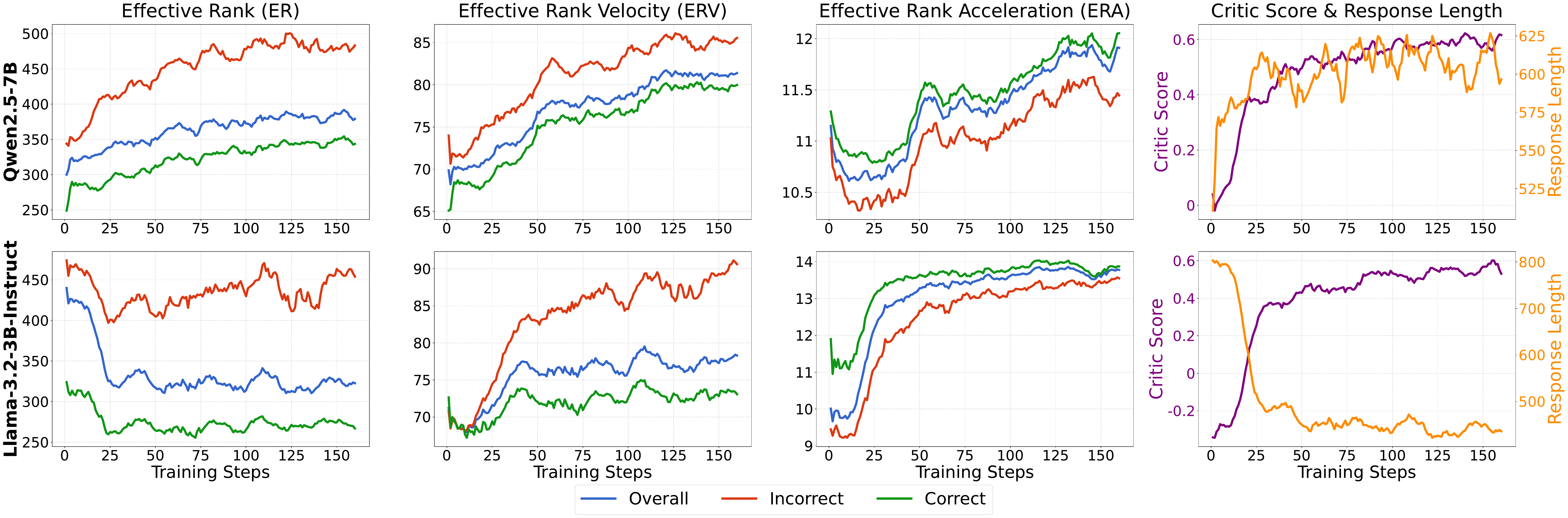}

    \caption{Response-level metrics during GRPO post-training, smoothed with a 10-step rolling window. Metrics are shown for the \textcolor{blue}{Overall} batch, as well as for subsets of \textcolor[RGB]{0,140,0}{Correct} and \textcolor{red}{Incorrect} samples. The rightmost column displays the average \textcolor[RGB]{107,30,120}{Critic Score} (reward) and \textcolor{orange}{Response Length} per batch.}
    \label{observation1}
\end{figure*}

\subsectionwithouttoc{Semantic Exploitation: Effective Rank Velocity (\ourV{}) and Acceleration (\ourAcc{})}
\label{sec:Dynamic Metrics}

\ourER{} provides a \emph{static} measure of how broadly a hidden-state trajectory spans semantic directions. To capture how this semantic breadth evolves during reasoning—whether the trajectory continues to expand into new directions, steadily refines within a region, or gradually saturates—we introduce two temporal-difference metrics: Effective Rank Velocity (\ourV{}) and Effective Rank Acceleration (\ourAcc{}).
Importantly, these dynamics are defined \emph{in semantic space} through hidden-state representations rather than in the token-level action space. This choice aligns with our view that exploitation reflects \emph{productive semantic refinement}, which is more than just low-entropy sampling.

\noindent\textbf{Setup. }
Let $M$ be any trajectory metric computed on hidden states (like $M=\ourER{}$). Given a response of length $T$, let $m_t$ denote the value of $M$ computed on the prefix hidden-state matrix up to position $t$. To reduce sensitivity to short-range token fluctuations and control computation, we evaluate the metric every $s$ tokens, where $s$ is the temporal stride. Then, define the evaluation indices as $\mathcal{T} = \{s, 2s, \ldots, Ks\}$, where $K = \lfloor (T-1)/s \rfloor$.

\begin{definition}[\ourV{}, denoted by $\Delta_M^{(1)}$, as the \textit{average historical-baseline deviation}]

\label{def_velo}

For each step $j \in \mathcal{T}$, we define the per-step historical-baseline deviation from the historical-prefix average:

\begin{equation}
\delta_{j\cdot s}\vcentcolon=
m_{j\cdot s}-\frac{1}{j-1}\sum_{k=1}^{j-1} m_{k\cdot s}.
\end{equation}

It measures whether current prefixes tend to exceed earlier-prefix averages, rather than the one-step change $m_t-m_{t-1}$. Thus, this form contrasts the current semantic breadth with the historical average over earlier prefixes. This tracks net semantic progress against running history, aligning with exploitation as greedy trajectory refinement under the current policy guidance (Appendix.~\ref{app:erv_exploitation}). A large positive $\delta_{j\cdot s}$ indicates that the trajectory at step $j\cdot s$ expands semantic breadth beyond its previous trend; a small value suggests that semantic expansion is no longer outpacing the past.

Then, we average these $\delta_{j\cdot s}$ of the response to obtain the \textit{average historical-baseline deviation} for the whole reasoning trajectory:

\begin{equation}
\Delta_M^{(1)} \vcentcolon=
\frac{1}{K-1}\sum_{j=2}^{K}\delta_{j\cdot s}.
\end{equation}
\end{definition}

We specialize it as $\ourV{}\vcentcolon=\Delta^{(1)}_{\ourER{}}$, and interpret it as a measure of \emph{exploitation in semantic space} during reasoning: sustained positive \ourV{} reflects consistent representational refinement along the current reasoning path, whereas small \ourV{} indicates diminishing returns and potential stagnation.

To quantify the changes in semantic breadth across different steps, we define the consecutive-step increments as $\Delta m_{r \cdot s} \vcentcolon= m_{r \cdot s} - m_{(r-1) \cdot s}$, where $r \in \mathcal{T}$. This represents the difference in the metric $M$ at every $s$-th step of the trajectory. We can then rearrange $\delta_{j \cdot s}$, the historical-baseline deviation, as follows:
\begin{equation}
\delta_{j \cdot s}
= \frac{1}{j-1} \sum_{r=2}^j (r-1)\,\Delta m_{r \cdot s}, 
\quad j \ge 2,
\end{equation}
which shows that $\delta_{j \cdot s}$ is a time-weighted average of local increments, giving more weight to recent $\Delta m_{r \cdot s}$. This provides a dynamic measure of semantic progress, emphasizing how the trajectory evolves over time.

\begin{figure*}[!t]

    \centering 

    \includegraphics[width=\textwidth]{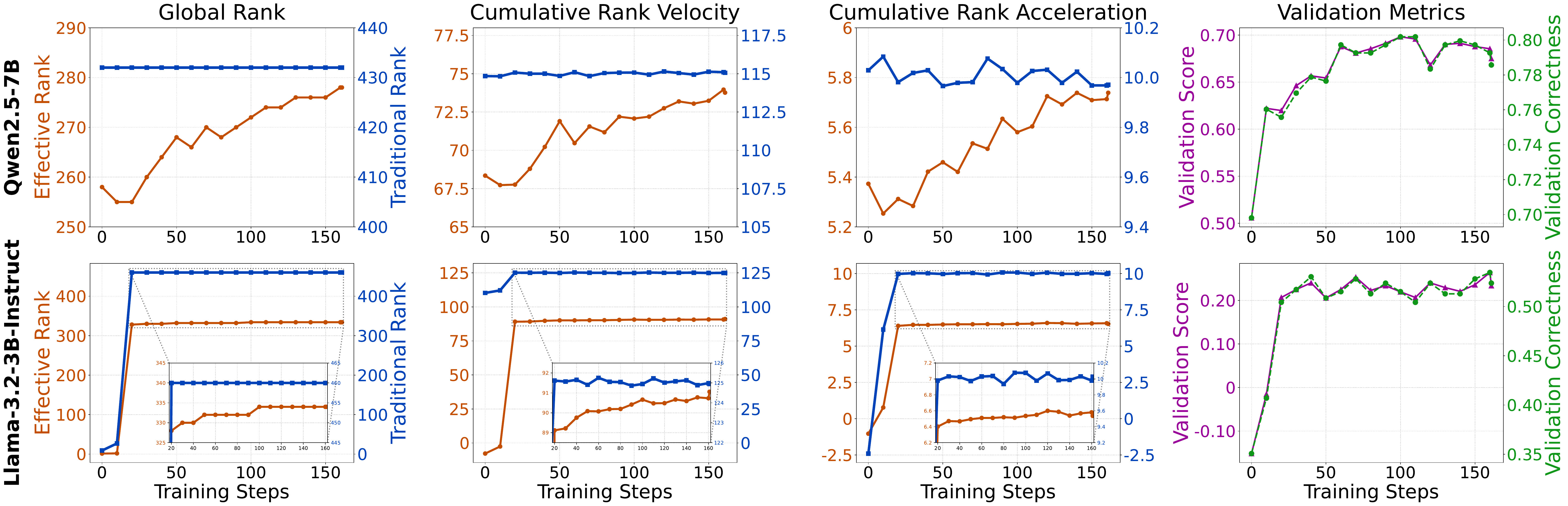}

    \caption{Visualization of dataset-level metrics during GRPO post-training. The figure compares \textcolor[RGB]{36,65,171}{traditional rank-based metrics} with our \textcolor{brown}{effective-rank-based metrics}. Also shown are the \textcolor[RGB]{107,30,120}{validation score} and sample \textcolor[RGB]{0,140,0}{correctness}, both averaged over the validation dataset.}

    \label{observation2}
\end{figure*}

\begin{definition}[\ourAcc{}, denoted by $\Delta_M^{(2)}$, as the \textit{average change of historical-baseline deviation}]

\label{def_accel}
To capture whether the refinement dynamics are \emph{accelerating} or \emph{saturating}, we define the \textit{average change of historical-baseline deviation}:
\begin{equation}
\Delta_M^{(2)} \vcentcolon= \frac{1}{K-2} \sum_{j=3}^{K} \big[\delta_{j \cdot s} - \delta_{(j-1) \cdot s}\big].
\end{equation}
\end{definition}

\noindent\textbf{Interpretation.}
We specialize $\ourAcc{}\vcentcolon=\Delta^{(2)}_{\ourER{}}$. Positive \ourAcc{} indicates that the semantic refinement reflected by $\delta_{j\cdot s}$ is strengthening over time (accelerating refinement), while negative \ourAcc{} suggests that refinement is weakening, consistent with stabilization or saturation in a narrow semantic region.
In later sections, we leverage this stability-oriented signal to modulate training incentives in a principled and adaptive manner.


\subsectionwithouttoc{Scaling Properties of Semantic-Space Dynamics}
\label{sec:scaling}

In Sec.~\ref{sec:static metric} and Sec.~\ref{sec:Dynamic Metrics}, we defined \ourER{}, \ourV{}, and \ourAcc{} on the hidden-state trajectory of a single response, i.e., in the model's semantic space.
We now study how these representational dynamics scale at two levels: (i) \textbf{dataset-level} scaling as the number of responses $N$ grows, and (ii) \textbf{trajectory-level} scaling as the reasoning length $T$ increases.
The key observation is that both settings can be analyzed through the same lens: the effective number of approximately orthogonal semantic directions, denoted by $k$. The following proposition provides a unified framework.

\begin{proposition}
\label{pro_ON}
\textit{Assume a hidden-state matrix has effective support on $k$ approximately orthogonal directions. Then \ourER{} and its corresponding \ourV{} are upper-bounded linearly by $k$:
$\ourER{}=\mathcal{O}(k)$ and $\Delta_{\ourER{}}^{(1)}=\mathcal{O}(k)$.
In contrast, the corresponding \ourAcc{} does not scale with $k$:
$\Delta_{\ourER{}}^{(2)}=\mathcal{O}(1)$.}
\end{proposition}

\begin{remark}
Prop.~\ref{pro_ON} admits two interpretations, depending on what the rows of $\mathbf{Z}$ represent.
\textbf{Dataset level:} with $\mathbf{Z}=\bar{\mathbf{Z}}\in\mathbb{R}^{N\times D}$ (rows are response embeddings), $k$ measures the number of distinct semantic modes covered by the dataset, so \ourER{}/\ourV{} grow with diversity of reasoning across questions while \ourAcc{} stays $\mathcal{O}(1)$.
\textbf{Response level:} with $\mathbf{Z}\in\mathbb{R}^{T\times D}$ (rows are step-wise hidden states along one trace), $k$ measures how many reasoning steps add non-redundant semantic directions for a given question, so \ourER{}/\ourV{} indicates novelty of reasoning within a response and $\mathcal{O}(1)$ \ourAcc{} indicates non-saturating refinement.
\end{remark}

\sectionwithouttoc{Empirical Evidence for Decoupling in Semantic Space}
\label{Sec Experimental Study}
We empirically study how exploration and exploitation evolve \emph{in the hidden-state semantic space} during RLVR training. Concretely, we track both conventional rank and our semantic-space metrics (\ourER{}, \ourV{}, \ourAcc{}) throughout GRPO post-training. We experiment with Qwen~\citep{hui2024qwen2} and Llama~\citep{grattafiori2024llama} under the GRPO paradigm~\citep{deepseek-math}, using the training configuration of \citet{zeng2025simplerlzoo}: 8k hard MATH problems (level 3-5), each paired with a verifiable reference answer. Complementary theoretical analysis is available in Appendix.~\ref{app:prop_old_new}.

\begin{figure*}[!t]
    \centering 
    \includegraphics[width=\textwidth]{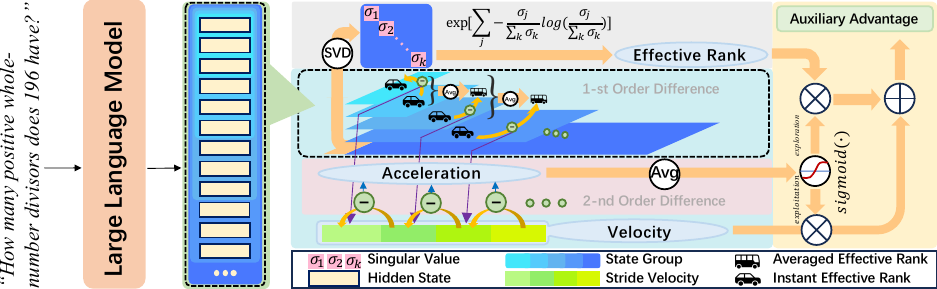}
    \caption{Overview of Velocity-Exploiting Rank-Learning. \textcolor[RGB]{100,100,100}{Exploration} is quantified by computing the \textcolor[RGB]{100,100,100}{\ourER{}} of the prefix hidden states via SVD, while \textcolor[RGB]{50,150,200}{exploitation} is captured through EMA-smoothed first-order difference (\textcolor[RGB]{50,150,200}{\ourV{}}) on per-step prefix hidden-state matrices and extended to second-order difference (\textcolor[RGB]{200,50,100}{\ourAcc{}}). Finally, exploration and exploitation are adaptively integrated to derive the \textcolor[RGB]{255,200,50}{auxiliary advantage}.}
    \label{Overview}
\end{figure*}

\subsectionwithouttoc{Analysis of Response-Level Metrics}
\label{Main_Response_Analysis}

At each training step, we compute response-level metrics on the current batch (Overall / Correct / Incorrect) as shown in Fig.~\ref{observation1}, with additional analyses deferred to Appendix.~\ref{Appendix_Response_Analysis}.

\textbf{\ourER{} varies by model, while \ourV{} increases consistently across models.}
Although RL consistently improves task performance, it induces different exploration patterns across base models in semantic space. As shown in the first column of Fig.~\ref{observation1}, Qwen exhibits an increasing \ourER{}, suggesting broader coverage over semantic directions, whereas Llama shows a decreasing \ourER{}, consistent with a more concentrated trajectory. In contrast, \ourV{} (second column) exhibits a consistent upward trend for both models, indicating that RL fine-tuning reliably strengthens \emph{semantic refinement dynamics}. Concretely, representations evolve in a way that increasingly outpaces their historical trend, even when the models differ in their exploration breadth.

\textbf{\ourAcc{} is more strongly aligned with outcome-relevant refinement than ER or ERV alone.}
We further compare correct vs. incorrect trajectories in Fig.~\ref{observation1}. Interestingly, incorrect samples often exhibit higher \ourER{} and \ourV{} than correct ones, suggesting that larger semantic breadth (high \ourER{}) or faster growth relative to the past (high \ourV{}) can reflect \emph{unproductive expansion} that drifts away from a correct solution. In contrast, \ourAcc{} is consistently higher for correct trajectories, indicating that correct reasoning is characterized not merely by being ``large'' or ``fast'', but by an \emph{increasingly coherent refinement} where the refinement dynamics strengthen over time. This suggests that \ourAcc{} is a more reliable indicator of robust reasoning than zero- or first-order quantities of the \ourER{} dynamics.

\subsectionwithouttoc{Analysis of Dataset-level Metrics}
\label{Main_Dataset_Analysis}

Following Sec.~\ref{Preliminaries}, we extend the analysis from individual trajectories to the validation set by constructing the dataset hidden-state matrix $\overline{\mathbf{Z}}$. We then track the zero-, first-, and second-order dynamics of $\mathrm{erank}(\overline{\mathbf{Z}})$ during training (Fig.~\ref{observation2}), with additional studies in Appendix.~\ref{Appendix_Dataset_Analysis}.

\textbf{Policy improvement correlates with expanding dataset-level semantic diversity.}
Across training, we observe that validation performance (accuracy and score) co-evolves with dataset-level \ourER{} dynamics. As optimization proceeds, $\mathrm{erank}(\overline{\mathbf{Z}})$ and its temporal differences tend to increase, suggesting that the learned policy occupies a richer set of semantic directions across the same problem set.
The upward trends of \ourV{} and \ourAcc{} further indicate that the policy becomes progressively more effective at refining and expanding this dataset-level semantic space toward correct solutions.

\textbf{\ourER{} reveals refinement even when conventional rank plateaus.}
In late-stage training, conventional rank may plateau, suggesting that the number of linearly independent directions in $\overline{\mathbf{Z}}$ no longer increases. However, \ourER{} can continue to rise, revealing a subtler improvement: the singular spectrum becomes more balanced, meaning the model utilizes existing semantic directions more uniformly rather than concentrating energy on a few dominant ones. This reduced redundancy indicates refinement within the established semantic space, beyond what discrete rank alone can capture.

\begin{table*}[!t]
\centering
\caption{Performance comparison of models on mathematical reasoning benchmarks (Pass@1). 
“+ GRPO” and “+ PPO” denote RL fine-tuning using the GRPO and PPO frameworks, respectively. 
“w/ \ourmethod{}” indicates incorporating our \ourmethod{} into the corresponding RL algorithm. 
$\Delta$ represents the performance contrast between original RL method and its \ourmethod{} variant. See Appendix.~\ref{Appendix_Pass@1} for full details.}
\begin{adjustbox}{width=\textwidth,center}
\begin{tabular}{lcccccccccccccccc}
\toprule
\textbf{Model} & 
\makecell[c]{\textbf{AIME24}} & 
\makecell[c]{\textbf{AIME25}} & 
\makecell[c]{\textbf{AMC23}} & 
\makecell[c]{\textbf{AMC24}} & 
\makecell[c]{\textbf{ASDiv}} & 
\makecell[c]{\textbf{Carp\_En}} & 
\makecell[c]{\textbf{CMATH}} & 
\makecell[c]{\textbf{Gaokao}\\ \textbf{2024\_I}} & 
\makecell[c]{\textbf{Gaokao}\\ \textbf{2024\_Mix}} & 
\makecell[c]{\textbf{Gaokao}\\ \textbf{MathCloze}} &
\makecell[c]{\textbf{GSM8K}} & 
\makecell[c]{\textbf{MAWPS}} & 
\makecell[c]{\textbf{Olympiad}\\ \textbf{Bench}} & 
\makecell[c]{\textbf{SVAMP}} & 
\makecell[c]{\textbf{TabMWP}} & 
\makecell[c]{\textbf{Avg.}} \\
\midrule

\textbf{Llama-3.2-3B-Instruct} 
& 0.0 & 0.0 & 25.0 & 11.1 & 74.6 & 26.5 & 10.2 & 14.3 & 14.3 & 6.8 & 66.6 & 86.9 & 12.7 & 74.1 & 41.4 & 31.0 \\
+ GRPO 
& 3.3 & 0.0 & 27.5 & 8.9 & 88.8 & 45.0 & 28.3 & 21.4 & 20.9 & 23.7 & 80.7 & 96.0 & 16.7 & 87.7 & 71.7 & 41.4 \\
\rowcolor{blue!10} + GRPO w/ \ourmethod{} 
& 13.3 & 6.7 & 25.0 & 11.1 & 89.3 & 45.4 & 46.2 & 14.3 & 22.0 & 22.9 & 81.7 & 96.0 & 17.6 & 87.8 & 72.3 & 43.4 \\
$\Delta_{\text{GRPO}}$ 
& \textcolor{positive}{+10.0} & \textcolor{positive}{+6.7} & \textcolor{negative}{-2.5} & \textcolor{positive}{+2.2} & \textcolor{positive}{+0.5} & \textcolor{positive}{+0.4} & \textcolor{positive}{+17.9} & \textcolor{negative}{-7.1} & \textcolor{positive}{+1.1} & \textcolor{negative}{-0.8} & \textcolor{positive}{+1.0} & \textcolor{neutral}{+0.0} & \textcolor{positive}{+0.9} & \textcolor{positive}{+0.1} & \textcolor{positive}{+0.6} & \textcolor{positive}{+2.0} \\
+ PPO 
& 10.0 & 3.3 & 22.5 & 13.3 & 87.9 & 46.4 & 21.2 & 7.1 & 16.5 & 20.3 & 81.4 & 95.5 & 17.8 & 86.8 & 71.0 & 40.1 \\
\rowcolor{blue!10} + PPO w/ \ourmethod{} 
& 10.0 & 3.3 & 25.0 & 11.1 & 88.7 & 46.0 & 30.7 & 14.3 & 19.8 & 27.1 & 82.9 & 95.7 & 17.3 & 85.8 & 71.3 & 41.9 \\
$\Delta_{\text{PPO}}$ 
& \textcolor{neutral}{+0.0} & \textcolor{neutral}{+0.0} & \textcolor{positive}{+2.5} & \textcolor{negative}{-2.2} & \textcolor{positive}{+0.8} & \textcolor{negative}{-0.4} & \textcolor{positive}{+9.5} & \textcolor{positive}{+7.2} & \textcolor{positive}{+3.3} & \textcolor{positive}{+6.8} & \textcolor{positive}{+1.5} & \textcolor{positive}{+0.2} & \textcolor{negative}{-0.5} & \textcolor{negative}{-1.0} & \textcolor{positive}{+0.3} & \textcolor{positive}{+1.9} \\
\midrule

\textbf{Qwen2.5-7B} 
& 6.7 & 0.0 & 45.0 & 15.6 & 91.4 & 55.8 & 86.7 & 42.9 & 33.0 & 49.2 & 85.8 & 95.4 & 25.8 & 88.5 & 82.8 & 53.6 \\
+ GRPO 
& 10.0 & 6.7 & 55.0 & 26.7 & 94.8 & 60.2 & 91.7 & 14.3 & 34.1 & 64.4 & 90.2 & 97.6 & 36.1 & 92.8 & 91.3 & 57.7 \\
\rowcolor{blue!10} + GRPO w/ \ourmethod{} 
& 13.3 & 10.0 & 50.0 & 28.9 & 95.0 & 60.8 & 90.7 & 35.7 & 35.2 & 69.5 & 89.2 & 97.7 & 35.4 & 92.9 & 91.9 & 59.8 \\
$\Delta_{\text{GRPO}}$ 
& \textcolor{positive}{+3.3} & \textcolor{positive}{+3.3} & \textcolor{negative}{-5.0} & \textcolor{positive}{+2.2} & \textcolor{positive}{+0.2} & \textcolor{positive}{+0.6} & \textcolor{negative}{-1.0} & \textcolor{positive}{+21.4} & \textcolor{positive}{+1.1} & \textcolor{positive}{+5.1} & \textcolor{negative}{-1.0} & \textcolor{positive}{+0.1} & \textcolor{negative}{-0.7} & \textcolor{positive}{+0.1} & \textcolor{positive}{+0.6} & \textcolor{positive}{+2.1} \\
+ PPO 
& 6.7 & 3.3 & 50.0 & 33.3 & 94.9 & 59.6 & 89.8 & 28.6 & 31.9 & 63.6 & 89.1 & 97.3 & 36.1 & 92.8 & 90.8 & 57.9 \\
\rowcolor{blue!10} + PPO w/ \ourmethod{} 
& 10.0 & 6.7 & 52.5 & 33.3 & 94.8 & 60.0 & 90.3 & 28.6 & 34.1 & 66.9 & 90.2 & 97.8 & 36.1 & 92.5 & 90.6 & 59.0 \\
$\Delta_{\text{PPO}}$ 
& \textcolor{positive}{+3.3} & \textcolor{positive}{+3.3} & \textcolor{positive}{+2.5} & \textcolor{neutral}{+0.0} & \textcolor{negative}{-0.1} & \textcolor{positive}{+0.4} & \textcolor{positive}{+0.5} & \textcolor{neutral}{+0.0} & \textcolor{positive}{+2.2} & \textcolor{positive}{+3.3} & \textcolor{positive}{+1.1} & \textcolor{positive}{+0.5} & \textcolor{neutral}{+0.0} & \textcolor{negative}{-0.3} & \textcolor{negative}{-0.2} & \textcolor{positive}{+1.1} \\
\bottomrule
\end{tabular}
\end{adjustbox}
\label{tab:pass_1}
\end{table*}

\begin{table*}[!t]
\centering
\caption{Performance comparison of models on mathematical reasoning benchmarks (Pass@$k$). 
“+ GRPO” and “+ PPO” denote RL fine-tuning using the GRPO and PPO frameworks, respectively. 
“w/ \ourmethod{}” indicates incorporating our \ourmethod{} into the corresponding RL algorithm. 
$\Delta$ represents the performance contrast between original RL method and its \ourmethod{} variant. See Appendix.~\ref{Appendix_Pass@k} for full details.}
\setlength{\tabcolsep}{4pt} 
\renewcommand{\arraystretch}{0.8} 
\begin{adjustbox}{width=\textwidth,center}
\footnotesize 
\begin{tabular}{lcccccc ccccc ccccc}
\toprule
\textbf{Model} &
\makecell[c]{\textbf{MATH500} \textbf{(Pass@16)}} &
\makecell[c]{ } &
\makecell[c]{ } &
\makecell[c]{\textbf{AMC23} \textbf{(Pass@128)}} &
\makecell[c]{ } &
\makecell[c]{ } &
\makecell[c]{\textbf{AMC24} \textbf{(Pass@128)}} &
\makecell[c]{ } &
\makecell[c]{ } &
\makecell[c]{\textbf{AIME24} \textbf{(Pass@256)}} &
\makecell[c]{ } &
\makecell[c]{ } &
\makecell[c]{\textbf{AIME25} \textbf{(Pass@256)}} &
\makecell[c]{ } &
\makecell[c]{ } &
\makecell[c]{\textbf{Avg.}} \\
\midrule

\textbf{Llama-3.2-3B-Instruct} 
& 79.8 &&& 93.5 &&& 51.5 &&& 40.0 &&& 30.0 &&& 58.96 \\
+ GRPO 
& 80.2 &&& 95.4 &&& 60.6 &&& 40.0 &&& 30.0 &&& 61.24 \\
\rowcolor{blue!10} + GRPO w/ \ourmethod{} 
& 80.6 &&& 95.7 &&& 59.0 &&& 50.0 &&& 36.7 &&& 64.40 \\
$\Delta_{\text{GRPO}}$ 
& \pos{0.4} &&& \pos{0.3} &&& \negv{1.6} &&& \pos{10.0} &&& \pos{6.7} &&& \pos{3.16} \\
+ PPO 
& 82.2 &&& 94.5 &&& 57.0 &&& 46.7 &&& 36.7 &&& 63.42 \\
\rowcolor{blue!10} + PPO w/ \ourmethod{} 
& 82.4 &&& 94.7 &&& 57.8 &&& 46.7 &&& 40.0 &&& 64.32 \\
$\Delta_{\text{PPO}}$ 
& \pos{0.2} &&& \pos{0.2} &&& \pos{0.8} &&& \zero &&& \pos{3.3} &&& \pos{0.90} \\
\midrule
\textbf{Qwen2.5-7B} 
& 90.6 &&& 98.4 &&& 73.7 &&& 60.0 &&& 60.0 &&& 76.54 \\
+ GRPO 
& 90.8 &&& 97.8 &&& 78.3 &&& 56.7 &&& 50.0 &&& 74.72 \\
\rowcolor{blue!10} + GRPO w/ \ourmethod{} 
& 91.4 &&& 98.3 &&& 79.0 &&& 63.3 &&& 60.0 &&& 78.40 \\
$\Delta_{\text{GRPO}}$ 
& \pos{0.6} &&& \pos{0.5} &&& \pos{0.7} &&& \pos{6.6} &&& \pos{10.0} &&& \pos{3.68} \\
+ PPO 
& 91.2 &&& 98.6 &&& 74.3 &&& 53.3 &&& 56.7 &&& 74.82 \\
\rowcolor{blue!10} + PPO w/ \ourmethod{} 
& 91.4 &&& 98.0 &&& 74.4 &&& 56.7 &&& 66.7 &&& 77.44 \\
$\Delta_{\text{PPO}}$ 
& \pos{0.2} &&& \negv{0.6} &&& \pos{0.1} &&& \pos{3.4} &&& \pos{10.0} &&& \pos{2.62} \\
\bottomrule
\end{tabular}
\end{adjustbox}
\label{tab:pass_k}
\end{table*}


\sectionwithouttoc{Velocity-Exploiting Rank-Learning}
\label{sec:our_method}

Building on Sec.~\ref{Sec Metrics} and the empirical findings in Sec.~\ref{Sec Experimental Study}, we propose Velocity-Exploiting Rank Learning (\ourmethod{}), an advantage-shaping approach that injects \emph{semantic-space} signals derived from hidden-state dynamics into standard RL objectives. Conventional RLVR optimizes rewards in the token-level action space, but does not explicitly regulate how reasoning trajectories evolve in representation space. This can yield inefficient updates, e.g., unproductive semantic expansion or premature collapse into narrow reasoning patterns. \ourmethod{} addresses this by constructing an auxiliary advantage from \ourER{}, \ourV{}, and \ourAcc{}, encouraging both broader semantic coverage and productive refinement in a unified and stable manner.


\begin{table*}[!t]
    \centering
    \caption{Pass@1 performance under different $\beta$ settings. “Adaptive $\beta$” denotes $\beta\vcentcolon=\text{sigmoid}(d_2)$. Accuracy columns are percentages, whereas Score Avg reports the average validation score. \textbf{Boldface} indicates the best result.}
    \label{ablation_lambda}
    \resizebox{\textwidth}{!}{%
    \begin{tabular}{l c c c c c c c c c c c c c}
        \toprule
        \multirow{2}{*}[-1.5ex]{\makecell{\textbf{Training}\\ \textbf{Strategy}}} & \multirow{2}{*}[-1.5ex]{\makecell{\textbf{Score}\\ \textbf{Avg}}} & \multicolumn{3}{c}{\textbf{In Domain}} & \multicolumn{4}{c}{\textbf{Out of Domain}} & \multicolumn{5}{c}{\textbf{Hard Problems}} \\
        \cmidrule(lr){3-5} \cmidrule(lr){6-9} \cmidrule(lr){10-14}
        & & \textbf{MATH} & \textbf{MATH500} & \textbf{Avg} & \textbf{Gaokao} & \textbf{\makecell{CN Middle\\School}} & \textbf{CMATH} & \textbf{Avg} & \textbf{AIME24} & \textbf{AIME25} & \textbf{AMC23} & \textbf{AMC24} & \textbf{Avg} \\
        \midrule
        \textbf{GRPO} & 0.36 & \textbf{51.4} & 46.2 & 48.80 & \textbf{23.7} & 28.7 & 28.3 & 26.90 & 3.3 & 0.0 & \textbf{27.5} & 8.9 & 9.93 \\

        \textbf{GRPO+\ourmethod{}} {($\beta=0.5$)} & \textbf{0.38} & 51.2 & 47.2 & 49.20 & 21.2 & 36.6 & 38.7 & 32.17 & 10.0 & 0.0 & \textbf{27.5} & 8.9 & 11.60 \\
        \textbf{GRPO+\ourmethod{}} {(Adaptive $\beta$)} & \textbf{0.38} & 50.9 & \textbf{51.2} & \textbf{51.05} & 22.9 & \textbf{38.6} & \textbf{46.2} & \textbf{35.90} & \textbf{13.3} & \textbf{6.7} & 20.0 & \textbf{11.1} & \textbf{12.78} \\
        \bottomrule
    \end{tabular}
    }
    \label{tab:analysis_experiments}
\end{table*}
\begin{figure*}[!t]
    \centering
    \begin{subfigure}[b]{0.32\textwidth}
        \centering
        \includegraphics[width=\textwidth]{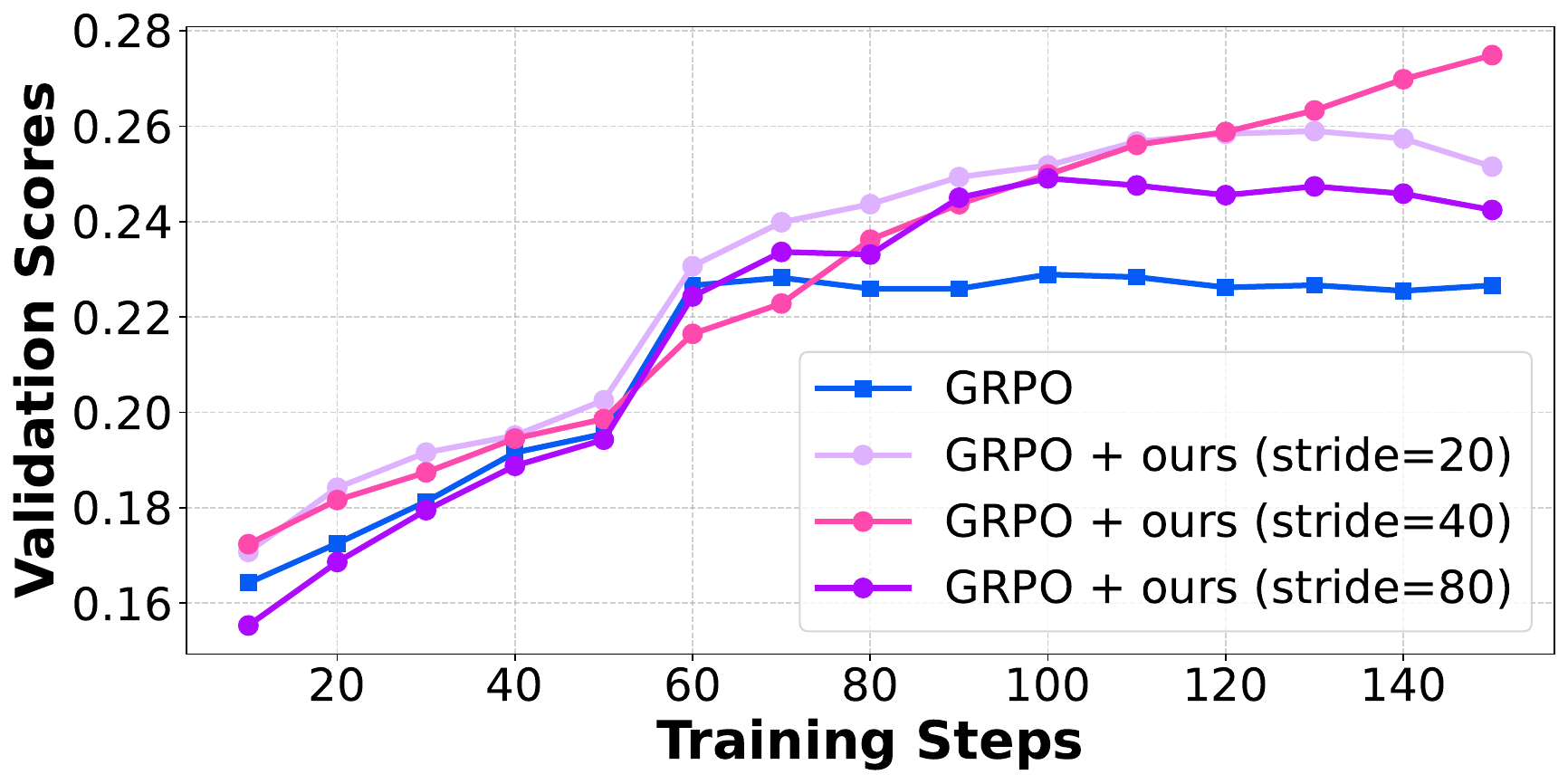}
        \caption{Different Strides (s).}
        \label{fig:sub_a}
    \end{subfigure}
    \hfill 
    \begin{subfigure}[b]{0.32\textwidth}
        \centering
        \includegraphics[width=\textwidth]{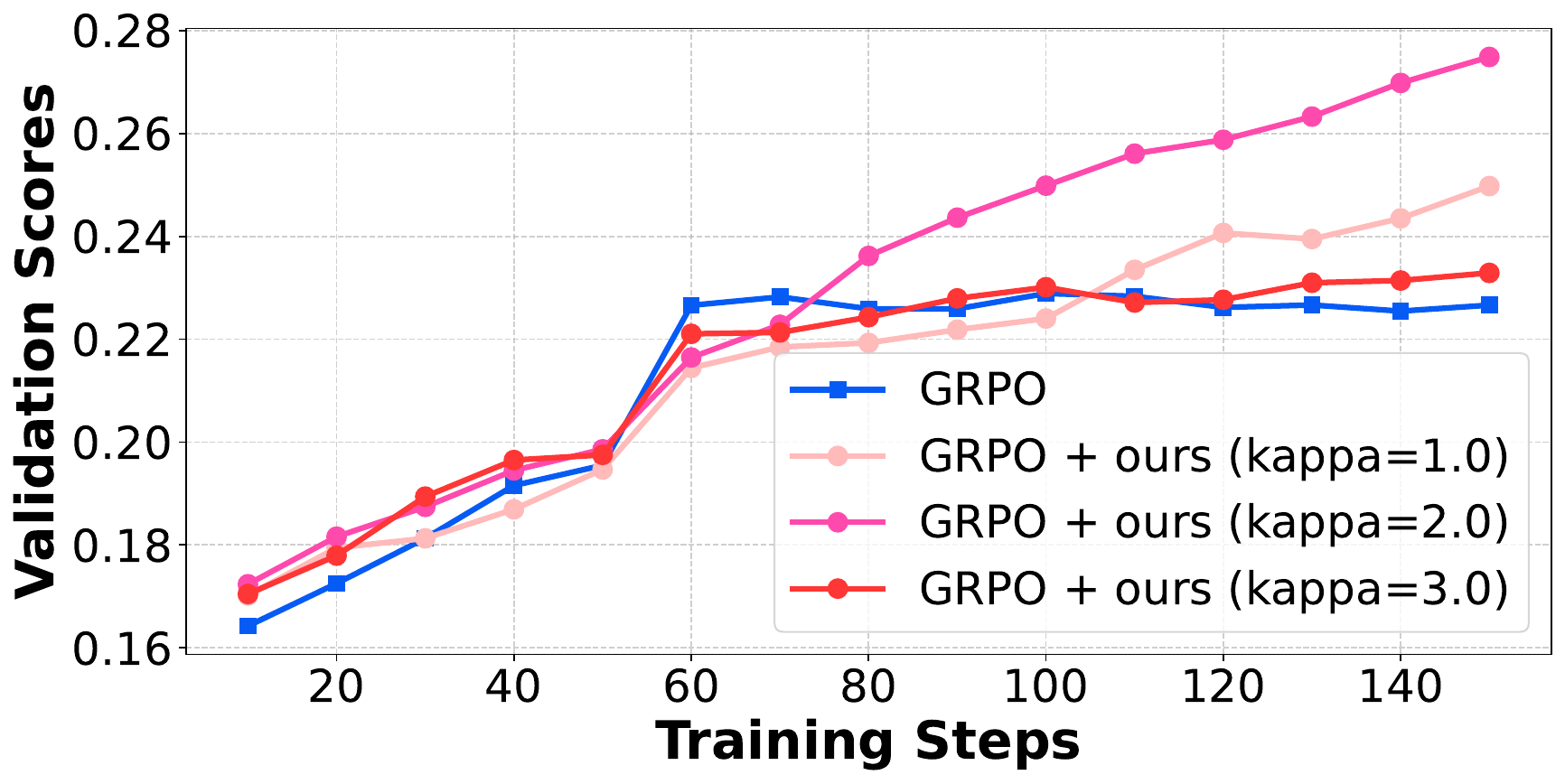}
        \caption{Different advantage clip ($\kappa$).}
        \label{fig:sub_b}
    \end{subfigure}
    \hfill 
    \begin{subfigure}[b]{0.32\textwidth}
        \centering
        \includegraphics[width=\textwidth]{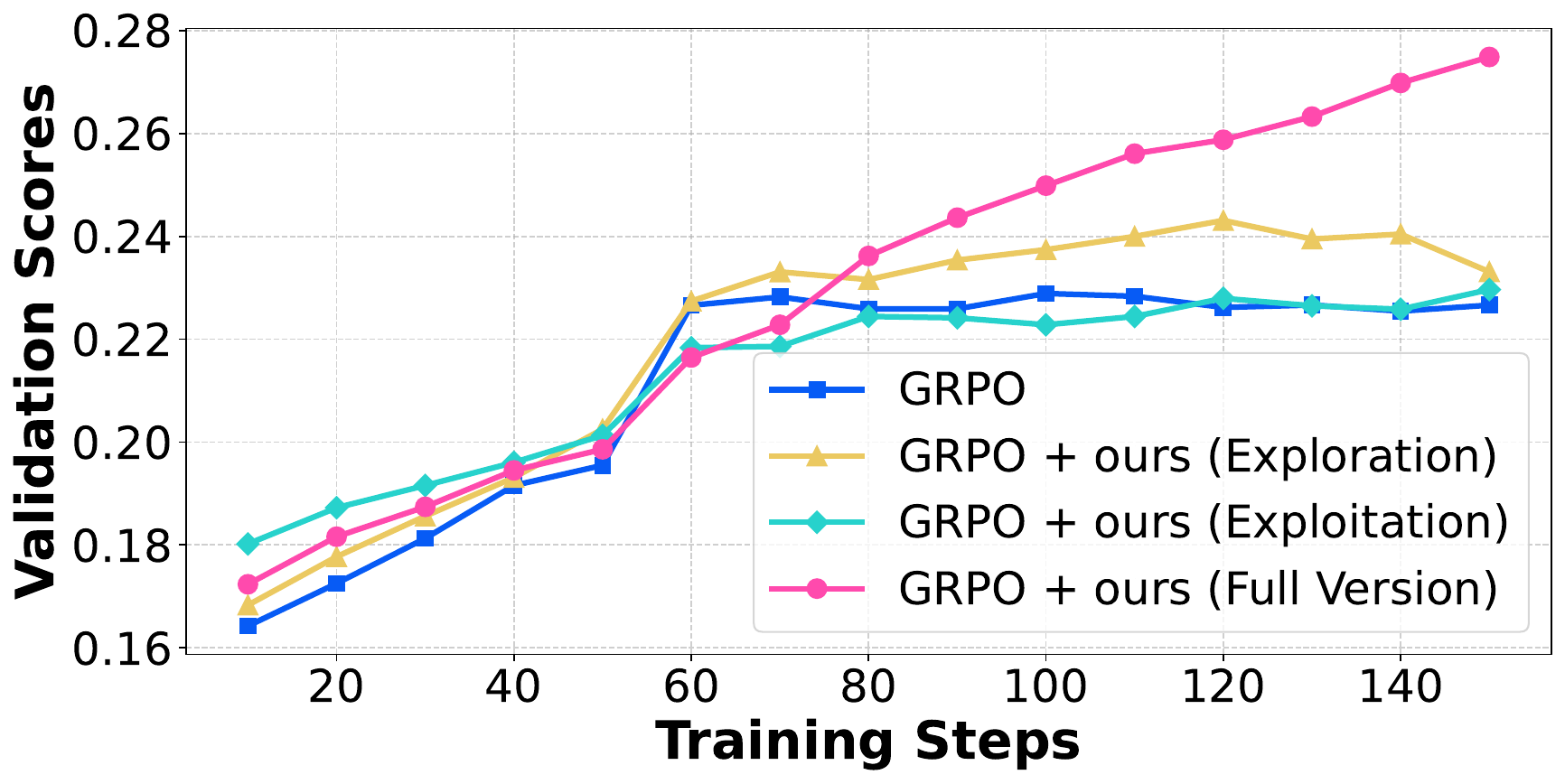} 
        \caption{Different shaping signals.} 
        \label{fig:sub_c}
    \end{subfigure}
    
    \caption{Hyperparameter ablations for Llama-3.2-3B-Instruct. Panels (a) and (b) show that VERL performs best with stride $s=40$ and clipping factor $\kappa = 2$, respectively. Panel (c) shows that using either the exploration or exploitation term alone is suboptimal, supporting the effectiveness of the combined signal.
    }
    \label{fig:hyperparameters}
\end{figure*}

\subsectionwithouttoc{Normalized Deviations in Semantic Space}

We compute three scalar metrics per trajectory, $\{m_0,m_1,m_2\}$, corresponding to \ourER{} (breadth), \ourV{} (first-order refinement dynamics), and \ourAcc{} (second-order refinement dynamics).
To obtain a stable, scale-invariant control signal across training, we normalize each metric by its recent trend using an exponential moving average (EMA) $\bar{\mu}_k$. We define the relative deviation:
\begin{equation}
d_k \vcentcolon= (m_k-\bar{\mu}_k) / |\bar{\mu}_k|
\quad k\in\{0,1,2\}.
\label{eq:deviation}
\end{equation}
Intuitively, $d_k$ measures how the current trajectory deviates from the policy's recent behavior in semantic space.


\subsectionwithouttoc{Adaptive Balancing via \ourAcc{} as a Meta-Control Signal}
\label{sec:dynamic-balancing}

Sec.~\ref{sec:static metric}--\ref{sec:Dynamic Metrics} motivates \ourER{} ($M_0$) as a proxy for exploration breadth in semantic space and \ourV{} ($M_1$) as a proxy for refinement dynamics (exploitation). Empirically, these two quantities are nearly uncorrelated (Fig.~\ref{first_fig}b), suggesting they can be jointly optimized rather than traded off. We therefore combine $M_0$ and $M_1$ into a single auxiliary signal, and use \ourAcc{} ($M_2$) as a meta-control variable to adapt their relative emphasis.

The key intuition is that \ourAcc{} captures the \emph{change} of refinement dynamics. When refinement accelerates strongly relative to its EMA trend ($d_2\gg0$), the policy may be at risk of collapsing into overly confident, narrow patterns that generalize poorly. When refinement decelerates or saturates ($d_2\le 0$), strengthening exploitation becomes more important to consolidate useful progress. This motivates an \ourAcc{}-controlled interpolation between exploration- and exploitation-oriented weights.

We define two basis weight vectors, $\mathbf{w}_{\text{explore}}=[1,0]$ for $M_0$ and $\mathbf{w}_{\text{exploit}}=[0,1]$ for $M_1$.
The dynamic weight is:
\begin{equation}
\mathbf{w}_{\text{dyn}} \vcentcolon= \beta\,\mathbf{w}_{\text{explore}} + (1-\beta)\,\mathbf{w}_{\text{exploit}},
\end{equation}
where $\beta \vcentcolon= \sigma(d_2)$ and $\sigma(\cdot)$ is the sigmoid. We then define the auxiliary trajectory-level signal:
\begin{equation}
\Phi \vcentcolon= w_{\text{dyn},0}\cdot\tanh(d_0) + w_{\text{dyn},1}\cdot\tanh(d_1),
\label{eq:shaping}
\end{equation}
where $\tanh$ bounds the auxiliary signal to improve stability. Intuitively, $\Phi$ rewards trajectories whose semantic breadth or refinement exceeds recent trends, while using $d_2$ to adaptively mitigate over-collapse and stagnation.

\subsectionwithouttoc{Advantage Shaping with Auxiliary Semantic-Space Signals}

We shape the learning signal by adding a clipped, non-negative bonus derived from $\Phi$. Let $A_t$ denote the original advantage (e.g., from GRPO or PPO with GAE), and let $\Phi_i$ be the trajectory-level auxiliary signal in Sec.~\ref{sec:dynamic-balancing}. We use the refined advantage:
\begin{equation}
\hat{A}_t
= A_t + \min\!\left(\operatorname{ReLU}(\Phi_i),\; |A_t|/\kappa\right),
\end{equation}
which preserves the sign of $A_t$ while bounding the shaping magnitude for stability.
For GRPO, $\Phi_i$ is applied once per trajectory; for PPO+GAE, the same $\Phi_i$ is broadcast to all tokens in the trajectory.

\sectionwithouttoc{Experiments}
\subsectionwithouttoc{Experiment Settings}
\label{sec:exp set}

\textit{(i)} \textbf{Datasets.} We use the same training and evaluation datasets as in Sec.~\ref{Sec Experimental Study}.
\textit{(ii)} \textbf{Reward.} We employ a rule-based reward that checks mathematical correctness and \verb|\boxed{}| formatting. A correct answer receives $+1.0$ if properly formatted and $+0.5$ otherwise; an incorrect answer receives $-0.5$ if formatted and $-1.0$ otherwise.
\textit{(iii)} \textbf{Training.} We build on \textsc{verl}~\citep{sheng2025hybridflow} with vLLM~\citep{kwon2023efficient}. We use batch size 48, generate 4 rollouts per prompt for GRPO and 1 rollout for PPO, and set the maximum generation length $L_{\max}=1536$.

\subsectionwithouttoc{Main Results}

\textbf{Generalization across benchmarks.}
Tab.~\ref{tab:pass_1} summarizes Pass@1 results on 15 mathematical reasoning benchmarks (full tables in Appendix.~\ref{Appendix_Pass@1}). Across both easy (e.g., ASDiv) and hard benchmarks (e.g., OlympiadBench), \ourmethod{} consistently improves over the corresponding RL baselines. Gains are most pronounced on multi-step reasoning benchmarks, including \textbf{+21.4} on Gaokao 2024\_I with Qwen2.5-7B and \textbf{+10.0} on AIME24 with Llama-3.2-3B-Instruct.

\textbf{Generalization across RL algorithms and base models.}
\ourmethod{} is plug-and-play: it can be paired with both GRPO and PPO, and improves average performance for both Llama and Qwen models in Tab.~\ref{tab:pass_1}. This indicates that semantic-space advantage shaping transfers across optimization algorithms and base-model families.

\textbf{Improving both exploration and exploitation proxies.}
Tab.~\ref{tab:pass_k} reports Pass@$k$ under diverse decoding budgets (Appendix.~\ref{Appendix_Pass@k}).
Pass@1 is commonly used as a proxy for strong single-shot performance, whereas Pass@$k$ reflects the ability to succeed given a larger search budget. We observe larger gains on Pass@$k$ than on Pass@1, especially on harder benchmarks, suggesting that \ourmethod{} improves not only single-trajectory refinement but also the coverage of successful reasoning trajectories under increased sampling.  See Sec.~\ref{sec:case_study} for qualitative analysis.


\textbf{Performance Degradation.}
As shown in Tab.~\ref{tab:pass_1}, VERL's degradations are localized rather than systematic. Under GRPO, the main drops are Llama on Gaokao 2024\_I (-7.1) and Qwen on AMC23/CMATH (-5.0/-1.0), yet the corresponding average Pass@1 still improves by +2.0 and +2.1 points, respectively. These drops are also offset by large gains on challenging benchmarks, such as AIME24 (+10.0) and CMATH (+17.9) for Llama, and Gaokao 2024\_I (+21.4) for Qwen. Moreover, the Llama drop on Gaokao 2024\_I is optimizer-dependent, as VERL improves the same benchmark by +7.2 under PPO. Overall, the occasional benchmark-level fluctuations are outweighed by consistent aggregate gains.

\subsectionwithouttoc{Ablations and Analysis}
We ablate key components and hyperparameters of \ourmethod{}, including the role of \ourAcc{}, the stride $s$ for temporal metrics, the advantage clip $\kappa$, and the composition of the shaping signal $\Phi$.

\textbf{Effectiveness of \ourAcc{}.}
Tab.~\ref{tab:analysis_experiments} shows that adding semantic-space shaping improves over GRPO. Moreover, adaptive mixing ($\beta=\sigma(d_2)$) outperforms fixed mixing ($\beta=0.5$), indicating that \ourAcc{} provides a useful meta-control signal for adjusting when to emphasize breadth (\ourER{}) versus refinement (\ourV{}). This suggests that VERL improves performance on OOD and challenging tasks.

\textbf{Temporal stride $s$.}
The stride $s$ controls the temporal granularity for computing \ourV{} and \ourAcc{}. As shown in Fig.~\ref{fig:hyperparameters}a, improvements are robust across a range of $s$, suggesting that the signal is not overly sensitive to sampling frequency.
We use $s=40$ in subsequent experiments as it yields the best validation reward while avoiding noisy token-level fluctuations.

\textbf{Advantage clip $\kappa$.}
The clip factor $\kappa$ bounds the shaping bonus so it complements, rather than dominates, the original advantage. As shown in Fig.~\ref{fig:hyperparameters}b, \ourmethod{} consistently improves performance across all tested $\kappa$, indicating robust behavior to this hyperparameter. The best performance is achieved at $\kappa=2$, which provides a sufficiently strong yet well-proportioned shaping signal.

\textbf{Shaping signal $\Phi$.}
Fig.~\ref{fig:hyperparameters}c compares variants of Eq.~\ref{eq:shaping}.
Using only the exploration term improves coverage but under-utilizes high-reward trajectories, leading to earlier bottlenecks.
Using only the exploitation term yields faster initial gains but plateaus due to insufficient breadth.
Combining both terms achieves the most stable training and the best final performance, consistent with our goal of improving both capacities in semantic space.

\sectionwithouttoc{Conclusion}
We argue that the exploration--exploitation trade-off in RLVR for LLM reasoning is largely a measurement artifact of token-level action statistics. Moving to hidden-state semantic space, we propose \ourER{}, \ourV{}, and \ourAcc{} to capture semantic breadth and refinement dynamics, and empirically find near-zero correlation between exploration and exploitation. Leveraging \ourAcc{} as a meta-control signal, \ourmethod{} performs semantic-space advantage shaping and consistently improves both in-domain performance and out-of-domain generalization on challenging reasoning benchmarks.

\section*{Limitations}
We have not systematically studied how training-data cleaning or curation affects VERL. While VERL shows clear advantages on math-focused training data with verifiable answers, extending these gains to other reasoning training corpora remains an open direction for future work. Furthermore, we have not addressed the computational costs associated with scaling VERL to larger, more diverse training datasets. 



\section*{Acknowledgments}
This work was supported by the National Natural Science Foundation of China (Grant Nos.~92467204 and 62472249), the Shenzhen Science and Technology Program (Grant No.~KJZD20240903102300001), the Shenzhen Science and Technology Program (Grant No.~JCYJ20250604145014018), and the Natural Science Foundation of Top Talent of SZTU (Grant No.~GDRC202413). The authors used AI assistants only for grammar and spelling correction; all scientific content was written and verified by the authors.

\bibliography{anthology}




\clearpage
\appendix
\renewcommand{\contentsname}{Contents of Appendix}
\tableofcontents  
\newpage
\onecolumn

\clearpage

\section{Notations}
\label{appendix:Notations}
\begin{table}[ht]
  \centering
  \label{tab:notations}
  \setlength{\tabcolsep}{20pt}
  \renewcommand{\arraystretch}{1.}
  \begin{adjustbox}{max width=\textwidth}
  \begin{tabular}{llr}
    \toprule
    \textbf{Symbol} & \textbf{Description} \\
    \midrule
    $\pi$ & Large language model policy \\
    $\theta$ & The parameter of large language model policy  \\
    $\phi$ & The parameter corresponding to the optimal policy \\

    $t$ & Time step \\
    $T$ & Response length \\
    $L_\text{max}$ & The maximum length of model's output \\

    $x$ & Prompt \\
    $\mathcal{P}_x$ & Distribution of prompts \\
    
    $y_t$ & The $t$-th step (token) of the model's response \\
    $y_{i:j}$ & Sequence of reasoning steps from $i$ to $j$ \\

    $D$ & Feature dimension of hidden states \\
    $N$ & The size of the dataset \\
    
    $z_t$& Hidden state corresponding to the $t$-th step of the output token \\
    $\bar{z}_i$ & Single vector representation for the $i$-th response by averaging its token hidden states \\
    $\mathbf{Z}_c$ & Mean-centered hidden state matrix \\
    $\overline{\mathbf{Z}}_{1:n}$ & Dataset-level hidden states matrix formed by the first $n$ prompts \\
    $\mathbf{Z}$ & Response-level hidden states matrix \\
    SVD$(\cdot)$ & The function to calculate the singular values \\
    $\lambda_j(\cdot)$ & The $j$-th eigenvalues of the given matrix \\
    $\sigma_j$ & The $j$-th singular value of matrix \\
    $p_j$& The $j$-th normalized singular values \\
    $\mathbf{p}$ & Singular value distribution \\

    $s$ & The stride for effective rank velocity calculation \\
    $\mathrm{rank}(\cdot)/\mathrm{erank}(\cdot)$ & Conventional rank/Effective rank function\\
    $\Delta_M^{(i)}$ & The $i$-order temporal difference for metric $M$\\
    $\mathcal{M}$ & Set of metrics derived from the hidden states\\
    $M_i$ & The $i$-order  temporal difference of \ourER, exactly the different metrics\\
    $\bar{\mu}_k$ & Exponential moving average of metric $M_k$ over training steps\\
    $m_t$ & Value of metric $M$ computed on the token sub-sequence from the start to position $t$\\
    $\delta^{(i)}_n$ & Instantaneous $i$-order difference for step $n$\\
    
    $\epsilon_{\mathrm{high}/\mathrm{low}}$ &Hyperparameter for the upper/lower bound used for clipping\\
    $\epsilon$ & Small constant for numerical stability\\

    

    $\mathbf{w}_{\text{explore}/\text{exploit}/\text{dyn}}$ & Exploration-focused profile/Exploitation-focused profile/Dynamic-weighted profile\\
    $w_{dyn,i}$ & The $i$-th scalar of $\mathbf{w}_{\text{dynamic}}$\\



    $r_j$ & Reward of the $j$-th response \\
    $A_{i,t}$ & Group-relative advantage for the $t$-th token in the $i$-th response in group\\
    $A$ & The original advantage estimation\\
    $\hat{A}$ & The reshaped advantage value\\
    $\Phi$ & Auxiliary advantage\\


    $\beta$ & Interpolation coefficient for VERL training \\

    $G$&Size of sampled group in GRPO\\

    $\rho_t$  & Probability ratio between the current and old policies for $t$-step of the output \\

    $\mathcal{L}_{\mathrm{PPO}}(\cdot)$& The optimization objective for PPO applied to policy\\

    
    \bottomrule
  \end{tabular}
  \end{adjustbox}
  \label{Notations1}
\end{table}

\clearpage
\onecolumn
\section{Algorithm}

\begin{algorithm}[H]
   \caption{\ourmethod{}: Training}
   \label{alg:ours_algorithm}
\begin{algorithmic}[1]
    \STATE {\bfseries Input:} dataset $\mathcal{D}$, prompt $x \in \mathcal{D}$, $|\mathcal{D}|=N$, policy model $\pi_\theta$, hidden-state dimension $D$ and number of rollouts per prompt $S$.
   
    \STATE {\bfseries Parameters:} EMA factor $\gamma$, relative deviation stabilizing factor $\epsilon \ll 1$, RL fine-tuning stabilizing factor $\kappa$.
   
    \STATE {\bfseries Initialize:} Randomly initialize policy parameters $\pi_\theta$, historical averages of metrics $\bar{\mu}_{\ourER{}} = \bar{\mu}_{\ourV{}} = \bar{\mu}_{\ourAcc{}} = 0$, exploration capacity profile$\mathbf{w}_{\text{explore}} = [1,0]$, exploitation capacity profile $\mathbf{w}_{\text{exploit}} = [0,1]$.
   
    \STATE {\bfseries Output:} A well-trained policy model  $\pi_\theta$.

    \STATE \textbf{repeat} 
    \STATE \hspace{1em} \textbf{for} $x \in \mathcal{D}$ \textbf{do}: \mycomment{Pick a sample from dataset}
    \STATE \hspace{2em} \textbf{for} 1 \textbf{to} $S$ \textbf{do}: \mycomment{Generate $S$ rollouts for one sample}
    \STATE \hspace{3em} $y_0 \gets x$, $\mathbf{Z}_0 \gets \emptyset$, $t\gets1$
    \STATE \hspace{3em} \textbf{repeat} \mycomment{Generation process}
    
    \STATE \hspace{4em} $y_{t}, \mathbf{z}_{t} \sim \pi_\theta(\cdot|y_{t-1})$ 
    \STATE \hspace{4em} $y_{t} \gets \left[y_{t-1}; y_{t}\right]$ \mycomment{Concatenate token sequence}
    
    \STATE \hspace{4em} $\mathbf{Z}_{t} \gets [\mathbf{Z}_{t-1}; \mathbf{z}_{t}] \in \mathbb{R}^{t \times D}$
    \STATE \hspace{4em} $\bm{\sigma}_{t} \gets \text{SVD}(\mathbf{Z}_{t})$
    \STATE \hspace{4em} $j \gets |\bm{\sigma}_{t}|$
    \STATE \hspace{4em} $p_{j,t} \gets \sigma_{j,t} / \sum_j \sigma_{j,t}$
    \STATE \hspace{4em} $\text{erank}_{t} \gets \exp\left(-\sum_j p_{j,t} \log p_{j,t}\right)$
    \STATE \hspace{4em} \textbf{If} $t > 1$ \textbf{then}: $\delta_{\ourV{},t} \gets \text{erank}_{t} - \frac{1}{t-1}\sum_{k=1}^{t-1} \text{erank}_{k}$
    \STATE \hspace{4em} \textbf{If} $t > 2$ \textbf{then}: $\delta_{\ourAcc{},t} \gets \delta_{\ourV{},t} - \delta_{\ourV{},t-1}$
    \STATE \hspace{4em} $t \gets t + 1$
    \STATE \hspace{3em} \textbf{until} response generation terminates; \mycomment{$t-1$ is the final timestep after generation}
    \STATE \hspace{3em} ${A}_{origin} \gets $ base RL evaluating on $y_{t-1}$
    \STATE \hspace{3em} $m_{\ourER} \gets \text{erank}_{t-1}$  \mycomment{Calculating \ourER{} metric}
    \STATE \hspace{3em} $m_{\ourV{}} \gets \frac{1}{t - 2}\sum_{t=2}^{t-1} \delta_{\ourV{}}^{t}$ \mycomment{Calculating \ourV{} metric}
    \STATE \hspace{3em} $m_{\ourAcc{}} \gets \frac{1}{t - 3}\sum_{t=3}^{t-1} \delta_{\ourAcc{}}^{t}$ \mycomment{Calculating \ourAcc{} metric}
    \STATE \hspace{3em} $\bar{\mu}_k \gets \gamma\bar{\mu}_k + (1-\gamma)m_k$, $k \in \{\ourER{},\ourV{},\ourAcc{}\}$
    \STATE \hspace{3em} $d_k \gets \frac{m_k - \bar{\mu}_k}{|\bar{\mu}_k| + \epsilon}$, $ k \in \{\ourER{},\ourV{},\ourAcc{}\}$
    \STATE \hspace{3em} $\beta \gets \text{sigmoid}(d_{\ourAcc{}})$
    \STATE \hspace{3em} $\mathbf{w}_{\text{dyn}} \gets \beta\mathbf{w}_{\text{explore}} + (1-\beta)\mathbf{w}_{\text{exploit}}$
    \STATE \hspace{3em} $w_{\text{dyn},\ourER{}} \gets$ the first scalar value of $\mathbf{w}_{\text{dyn}}$
    \STATE \hspace{3em} $w_{\text{dyn},\ourV{}{}} \gets$ the second scalar value of $\mathbf{w}_{\text{dyn}}$
    \STATE \hspace{3em} $\Phi \gets w_{\text{dyn},\ourER{}}\tanh(d_\ourER{}) + w_{\text{dyn},\ourV{}}\tanh(d_\ourV{})$
    \STATE \hspace{3em} $\hat{A} \gets A_{origin} + \min\left(\text{ReLU}(\Phi), \frac{|A_{origin}|}{\kappa}\right)$
    
    \STATE \hspace{2em} \textbf{end for}
    \STATE \hspace{1em} \textbf{end for}
    \STATE \hspace{1em} Update $\theta$ via base RL objective with $\hat{A}$
    \STATE \textbf{until} $\theta$ converges;
\end{algorithmic}
\end{algorithm}

\twocolumn

\section{Related Work}
\label{related_work}

\subsection{Reasoning with LLMs}
Reasoning with LLMs has become a central topic in recent large-model research, with surveys highlighting a shift from fast pattern to more deliberate reasoning behaviors~\citep{li2025system}. Prior work has studied mathematical reasoning and challenging specialized domains such as olympiad-level geometry solving~\citep{yu2025chain,duan2025gold,tian2025tabularmath}, while more recent efforts extend LLM reasoning to domain-specific and multimodal settings~\citep{jiang2026beyond}, including contract revision~\citep{xu2026rcbsfmultiagent}, self-correcting RAG~\citep{xu2026selfcorrectingrag}, personalized long-form generation~\citep{wang2026thinkwhilegenerating}, structured spatial reasoning~\citep{ma2026thinkingblueprintsassistingvisionlanguage}, CAD editing~\citep{macadmorph}, and 3D understanding~\citep{10.1145/3731715.3733405}. Related efforts also study tool-using reasoning, including pattern-aware tool-integrated reasoning~\citep{xu2025learning} and structured mid-level supervision for tool-using language models~\citep{jiang2026scribestructuredmidlevelsupervision}. At the model level, prior studies improve reasoning capability or efficiency through task-aware data augmentation~\citep{Ma_Chen_Zhang_Wu_Ding_2025}, sparse computation~\citep{chen2026sparse}, distillation~\citep{chen2025improving}, reasoning-oriented pruning~\citep{jiang2025drp}, and multi-teacher data synthesis~\citep{zhang2025find}, while another line emphasizes more reliable and better-directed reasoning via causal abstention~\citep{sun2025causalabstain}, causality-inspired evaluation~\citep{sun2026factecausalityinspiredevaluationtrustworthy}, robust uncertainty quantification for self-evolving LLMs~\citep{zhou2025robust}, and curiosity-driven exploration~\citep{dai2026cde}. In contrast, we focus on reasoning dynamics during RL training and study how exploration and exploitation can be characterized and improved in semantic space.

\subsection{Reinforcement Learning with Verifiable Rewards}
Compared with supervised fine-tuning~\citep{xue2026supervisedfinetuningfailslearn} or in-context learning~\citep{wang2025incomplete}, RLVR-based training is often viewed as more capable of eliciting advanced reasoning behaviors \citep{zeng2025simplerlzoo,yu2025dapo}. This approach is exemplified by DeepSeek-R1-Zero~\citep{guo2025deepseek} and OpenAI o1~\citep{jaech2024openai}, which execute complex reasoning processes through reflection and validation. Following the success of RLVR, a significant body of research has investigated the efficacy of RLVR on popular open-source LLMs, including Qwen~\citep{qwen2}, Mistral~\citep{jiang2024mixtral}, and LLaMA~\citep{grattafiori2024llama}. 

This has fostered an optimistic view that RLVR can not only enhance existing model capabilities but also enable the acquisition of novel reasoning knowledge, facilitating a path toward continuous self-improvement \citep{zeng2025simplerlzoo,yu2025dapo,wen2026reinforcement}. RLVR training has been shown to grant LLMs controllable output length for efficient inference~\citep{yan2025mur,cheng2025optimizing,li2025tl}, deepen their reasoning pathways~\citep{bensal2025reflect}, mitigate their weaknesses~\citep{liang2025sws,liang2026beyond}, enable the use of external tools~\citep{rainone2025replacing,jin2025search}, and even facilitate unsupervised reasoning~\citep{zuo2025ttrl}. However, some studies~\citep{yue2025does} argue that despite improving reasoning confidence and reliability, RLVR may inadvertently suppress exploration. By maximizing expected rewards, RLVR tends to reinforce high-reward trajectories while underweighting novel but uncertain ones. This tension between exploration and exploitation remains a core challenge in RLVR.

\subsection{Exploration and Exploitation in RLVR for LLMs}

Recent studies on exploration and exploitation in RLVR for LLMs have largely relied on token-level analyses of the predictive distribution~\citep{wang2025beyond,cui2025entropy}. In this line of work, next-token entropy is commonly treated as a proxy for exploration, motivating methods such as entropy regularization to encourage diverse reasoning paths~\citep{deng2025trial,cheng2026reasoning}. Conversely, lower entropy or higher token-level confidence is often viewed as a sign of exploitation~\citep{fu2025deep}. Building on this view, several approaches use confidence-related signals to assess, filter, or reinforce reasoning trajectories~\citep{damani2026beyond,qiao2025concise,li2025confidence,xue2026reasonneededefficientgenerative}. However, these measures remain closely tied to token-level uncertainty, which tends to entangle exploration and exploitation under the same measurement framework. 

In contrast, we move beyond token-level action statistics and study exploration and exploitation in semantic space at the response level, aiming to decouple and improve both during RL training.

\subsection{Representation Dynamics in Deep Reinforcement Learning}
Beyond RLVR for LLMs, classical deep RL has long studied how neural representations evolve during training and how they affect exploration, sample efficiency, and generalization. State representation learning for control aims to learn compact latent states that preserve task-relevant dynamics and improve the stability and efficiency of downstream RL algorithms~\citep{Lesort2018StateRL}. More recent work directly analyzes representation dynamics in deep RL, showing that feature collapse or capacity loss can emerge during training and harm performance~\citep{Kumar2020ImplicitUI,Lyle2022UnderstandingAP}. Other studies further examine broader representational properties, such as dynamics-awareness and orthogonality, and relate them to transfer performance across agents and tasks~\citep{Wang2022InvestigatingTP}. These findings suggest that representation geometry is closely tied to RL performance, motivating our focus on semantic-space dynamics in RLVR.


\section{Details of Theorems}

\subsection{Proof of Theorem~\ref{the_erank}}

\textit{Suppose we have a matrix of embeddings $\mathbf{Z}\in\mathbb{R}^{T \times D}$, which is nonzero. Then the effective rank of $\mathbf{Z}$ is a lower bound of $\mathrm{rank} (\mathbf{Z})$:}

\begin{equation}
1 \leq \mathrm{erank} (\mathbf{Z}) \leq \mathrm{rank} (\mathbf{Z}) \leq \text{min}\{T, D\}
\end{equation}

\noindent \textit{Proof.} Let the singular value distribution of the matrix $\mathbf{Z}$ be $\mathbf{p}=(p_1,p_2,\ldots,p_{min\{T,D\}})$. The Shannon entropy of this distribution $H(\mathbf{p})$ is bounded. Its minimum is 0, which occurs when only one element of $p$ is 1 and all others are 0. Its maximum is $\log k$, where $k$ is the number of non-zero singular values, and this occurs when the distribution is uniform ($p_j=1/k$ for all non-zero values). The lower bound is established from the minimum entropy value:

\begin{equation}
\mathrm{erank}(\mathbf{Z})=\exp(H(\mathbf{p}))\geq\exp(0)=1
\end{equation}

Equality holds if and only if the singular value distribution is $(1,0,\ldots,0)$, meaning $\mathbf{Z}$ has only one non-zero singular value. For the upper bound, let $k=\mathrm{rank}(\mathbf{Z})$ be the number of non-zero singular values of $\mathbf{Z}$. The entropy of the distribution $p$ is calculated only over these $k$ values and is maximized when they are uniform. Therefore

\begin{equation}
H(\mathbf{p})\leq\log k
\end{equation}

Applying the exponential function to this inequality gives: 

\begin{equation}
\begin{aligned}
\mathrm{erank}(\mathbf{Z})
&=\exp\!\bigl(H(\mathbf{p})\bigr) \\
&\le \exp(\log k)=k=\mathrm{rank}(\mathbf{Z}) \, .
\end{aligned}
\end{equation}

This establishes that the effective rank is upper-bounded by the conventional rank. The final inequality, $\operatorname{rank}(\mathbf{Z})\leq\min\{T,D\}$, is a standard property of matrix rank. Equality for $\mathrm{erank}(\mathbf{Z})=\mathrm{rank}(\mathbf{Z})$ holds if and only if the non-zero singular values are all equal, corresponding to a uniform singular value distribution over its support.

\subsection{Proof of Proposition~\ref{pro_ON}}

\textit{Assume a hidden-state matrix has effective support on $k$ approximately orthogonal directions. Then \ourER{} and its corresponding \ourV{} are upper-bounded linearly by $k$:
$\ourER{}=\mathcal{O}(k)$ and $\Delta_{\ourER{}}^{(1)}=\mathcal{O}(k)$.
In contrast, the corresponding \ourAcc{} does not scale with $k$:
$\Delta_{\ourER{}}^{(2)}=\mathcal{O}(1)$.}

\noindent \textit{Proof.} Without loss of generality, we take the effective rank for example. We adopt the provided definition of effective rank for a representation matrix $\mathbf{Z}$ with singular values $\{\sigma_i\}$:

\begin{equation}
\begin{split}
\mathrm{erank}(\mathbf{Z})
&= \exp\!\left(-\sum_{j} p_{j}\log(p_{j})\right), \\
\text{where}\quad
p_{j}
&= \frac{\sigma_{j}}{\sum_{k} \sigma_{k}} .
\end{split}
\end{equation}

Our analysis focuses on the dataset matrix $\overline{\mathbf{Z}} \in \mathbb{R}^{N\times D}$, whose rows $\{\mathbf{q}_i\}_{i=1}^N$ are the mean token embeddings of $N$ responses. The singular values $\sigma_i(\overline{\mathbf{Z}})$ of $\overline{\mathbf{Z}}$ are the square roots of the eigenvalues of the Gram matrix $\overline{\mathbf{K}}=\overline{\mathbf{Z}} \overline{\mathbf{Z}}^\top;\mathrm{i.e.,~}\sigma_j(\overline{\mathbf{Z}})=\sqrt{\lambda_j(\overline{\mathbf{K}})}$. Given that the rows of $\overline{\mathbf{Z}}$ are nearly orthogonal, the Gram matrix $\overline{\mathbf{K}}$ is strongly diagonal-dominant. Its eigenvalues can be approximated by its diagonal entries, $\mathrm{for~}j=1,\ldots,N$:

\begin{equation}
\lambda_j(\overline{\mathbf{K}})\approx \overline{\mathbf{K}}_{jj}=\|\mathbf{q}_j\|^2=\frac{1}{T}
\end{equation}

The matrix has $N$ significant eigenvalues, each approximately equal to $1/T$. The singular values of $\overline{\mathbf{Z}}$ are the square roots of the eigenvalues of $\overline{\mathbf{K}}$, $\mathrm{for~}j=1,\ldots,N$:

\begin{equation}
\sigma_j(\overline{\mathbf{Z}})=\sqrt{\lambda_j(\overline{\mathbf{K}})}\approx\sqrt{\frac{1}{T}}=\frac{1}{\sqrt{T}}
\end{equation}

To calculate the effective rank, we first normalize these singular values to form a probability distribution $\{p_j\}$. The sum of singular values is:

\begin{equation}
\sum_{k=1}^N\sigma_k(\overline{\mathbf{Z}})\approx\sum_{k=1}^N\frac{1}{\sqrt{T}}=\frac{N}{\sqrt{T}}
\end{equation}

The individual probabilities are therefore:

\begin{equation}
p_j=\frac{\sigma_j}{\sum_k\sigma_k}\approx\frac{1/\sqrt{T}}{N/\sqrt{T}}=\frac{1}{N}
\end{equation}

The distribution $\mathbf{p}=\{p_1,p_2,\dots,p_N\}$ is a uniform distribution over $N$ states. The Shannon entropy of this distribution is maximal:

\begin{equation}
\begin{aligned}
H(\mathbf{p})
&= -\sum_{j=1}^N p_j \log(p_j)
 = -\sum_{j=1}^N \frac{1}{N}\log\left(\frac{1}{N}\right) \\
&= -N\left(\frac{-\log(N)}{N}\right)
 = \log(N) \, .
\end{aligned}
\end{equation}

The effective rank is the exponential of this entropy: $\mathrm{erank}(\overline{\mathbf{Z}})=\exp(H(\mathbf{p}))=\exp(\log(N))=N$. In the maximal prompt entropy regime, the effective rank of the dataset matrix scales as $\mathcal{O}(N)$.

We adapt them to our context by defining the metric's value at ``time" $n$ as the Effective Rank computed on the dataset matrix formed by the first $n$ prompts, denoted $\overline{\mathbf{Z}}_{1:n}$. Let $m_n=\mathrm{erank}(\overline{\mathbf{Z}}_{1:n})$. From our previous analysis, we established a crucial result that forms the basis of this derivation: for maximal cases, the effective rank of a dataset with $n$ prompts scales linearly with $n$.

\begin{equation}
m_n = \mathrm{erank}(\overline{\mathbf{Z}}_{1:n}) \approx n
\end{equation}

We will use this linear approximation to derive the scaling orders of the difference metrics, assuming a stride of $s=1$ for simplicity. The first-order difference quantifies the average ``velocity" of change in the metric relative to its historical mean. Instantaneous First-Order Difference $(\delta_n^{(1)})$ is the value at step $n$ minus the average of all preceding values.

\begin{equation}
\delta_n^{(1)}=m_n-\left(\frac{1}{n-1}\sum_{k=1}^{n-1}m_k\right)
\end{equation}

Substituting our approximation $m_k\approx k$:

\begin{equation}
\delta_n^{(1)}\approx n-\left(\frac{1}{n-1}\sum_{k=1}^{n-1}k\right)
\end{equation}

\begin{equation}
\delta_n^{(1)}\approx n-\left(\frac{1}{n-1}\cdot\frac{(n-1)n}{2}\right)=n-\frac{n}{2}=\frac{n}{2}
\end{equation}

The historical-baseline deviation grows linearly with $n$. Overall First-Order Difference $\Delta_\mathrm{erank}^{(1)}$: This is the average of the historical-baseline deviation over the entire dataset of size $N$.

\begin{equation}
\Delta_\mathrm{erank}^{(1)}=\frac{1}{N-1}\sum_{n=2}^N \delta_n^{(1)}\approx\frac{1}{N-1}\sum_{n=2}^N\frac{n}{2}
\end{equation}

\begin{equation}
\begin{aligned}
\Delta_\mathrm{erank}^{(1)}
&\approx \frac{1}{2(N-1)}\left(\left(\sum_{n=1}^N n\right)-1\right) \\
&= \frac{1}{2(N-1)}\left(\frac{N(N+1)}{2}-1\right)\, .
\end{aligned}
\end{equation}

For large $N$, the expression is dominated by the highest power of $N$:

\begin{equation}
\Delta_\mathrm{erank}^{(1)} \sim \frac{N^2/4}{N}=\frac{N}{4}
\end{equation}

The first-order difference of the effective rank scales linearly with the number of prompts, $\Delta_\mathrm{erank}^{(1)}=\mathcal{O}(N)$. As for second-order difference, we compute the change in historical-baseline deviation between consecutive values of $\delta_n^{(1)}$.

\begin{equation}
\delta_n^{(1)} - \delta_{n-1}^{(1)}\approx\frac{n}{2}-\frac{n-1}{2}=\frac{1}{2}
\end{equation}

This change is a constant, indicating a linear increase in the first-order difference. Overall Second-Order Difference $\Delta_\mathrm{erank}^{(2)}$:

\begin{equation}
\begin{aligned}
\Delta_\mathrm{erank}^{(2)}
&= \frac{1}{N-2}\sum_{n=3}^N\left(d_n^{(1)}-d_{n-1}^{(1)}\right) \\
&\approx \frac{1}{N-2}\sum_{n=3}^N\frac{1}{2}\, .
\end{aligned}
\end{equation}

\begin{equation}
\Delta_\mathrm{erank}^{(2)} \approx \frac{1}{N-2}\cdot(N-2)\cdot\frac{1}{2}=\frac{1}{2}
\end{equation}

The second-order difference of the effective rank is constant and does not depend on $N$, yielding a scaling order of $\Delta_\mathrm{erank}^{(2)}=\mathcal{O}(1)$.

\section{Additional Theoretical Support for Exploration and Exploitation Metrics}
\label{app:theory_exploit_explore}

In this section we formalize the relationship between our proposed hidden-state metrics (\emph{Effective Rank} and \emph{Effective Rank Velocity}) and the classical notions of exploration and exploitation in reinforcement learning. We first show that the token-level metrics in action space (average log probability and response entropy) are algebraically coupled, whereas the hidden-state metrics in semantic space are not. We then provide a representation-level justification for interpreting Effective Rank as a measure of semantic exploration, and Effective Rank Velocity as a measure of representation-level exploitation that is strongly correlated with greedy value improvement under the PPO-style architecture used in RLVR. Throughout, we consider a conditional language model $p_\theta(y \mid x)$ and a Transformer backbone that produces hidden states $z_t \in \mathbb{R}^D$ at each time step $t$ for a given prompt $x$ and generated response $y_{1:T}$.

\subsection{Action vs.\ Semantic Exploration--Exploitation Metrics}
\label{app:prop_old_new}

In this subsection, we formalize the difference between the token-level \emph{action} metrics used in prior RLHF/RLVR work and the hidden-state \emph{semantic} metrics proposed in this paper. For a given prompt $x$ and generated response $y_{1:T}$, let $\pi_\theta(\cdot \mid x,y_{<t})$ denote the model's token-level policy distribution at step $t$, i.e.\ the softmax over the vocabulary induced by the logits at that position.

\paragraph{Token-level action space, log-probability, and entropy.} We define the \emph{average log probability} of a response and the \emph{response entropy} as

\begin{equation}
\begin{aligned}
\mathrm{AvgLogProb}(x,y_{1:T})
&:= \frac{1}{T}\sum_{t=1}^{T} \log \pi_\theta\big(y_t \mid x,y_{<t}\big),\\
\mathrm{RespEnt}(x,y_{1:T})
&:= \frac{1}{T}\sum_{t=1}^{T} H\!\left(\pi_\theta(\cdot \mid x,y_{<t})\right),\\
H(p)
&:= -\sum_{v} p(v)\log p(v),\\
&\quad p(v):=\pi_\theta\!\left(v \mid x,y_{<t}\right).
\end{aligned}
\end{equation}

Thus $\mathrm{RespEnt}$ is the \emph{token-level} entropy averaged over the response: at each step we compute the Shannon entropy of the vocabulary distribution and then average over time. In the response-level semantic space, we consider $x$ drawn from a prompt distribution $\mathcal{P}_{x}$ and, for the purpose of analysis, responses $y_{1:T}$ drawn \emph{on-policy} from the model $p_\theta(\cdot \mid x)$:

\begin{equation}
\begin{aligned}
&\mathcal{L}_{\text{avg-log}}(\theta) := \\
&\mathbb{E}_{x \sim p(x)} \;
  \mathbb{E}_{y_{1:T} \sim p_\theta(\cdot \mid x)}
  \Big[
    \mathrm{AvgLogProb}(x,y_{1:T})
  \Big],\\
&\mathcal{H}_{\text{resp}}(\theta) := \\
&\mathbb{E}_{x \sim p(x)} \;
  \mathbb{E}_{y_{1:T} \sim p_\theta(\cdot \mid x)}
  \Big[
    \mathrm{RespEnt}(x,y_{1:T})
  \Big].
\end{aligned}
\end{equation}

\begin{proposition}[Token-level exploitation and exploration are tightly coupled]
\label{prop:old_coupled_strong}
Under on-policy sampling $y_{1:T} \sim p_\theta(\cdot \mid x)$, the corpus-level average log probability $\mathcal{L}_{\text{avg-log}}(\theta)$ and response entropy $\mathcal{H}_{\text{resp}}(\theta)$ satisfy
\begin{equation}
    \mathcal{L}_{\text{avg-log}}(\theta)
    \;=\;
    -\mathcal{H}_{\text{resp}}(\theta).
    \label{eq:avglog_vs_respH_exact}
\end{equation}
In particular, under the same sampling distribution, any change of the model that increases token-level exploitation in action space as measured by $\mathcal{L}_{\text{avg-log}}$ necessarily decreases $\mathcal{H}_{\text{resp}}$ by the same amount (and vice versa).
\end{proposition}

\begin{proof}
Fix a prompt $x$ and a time step $t$. Conditioned on $x$ and the history $y_{<t}$, the next token $y_t$ is drawn from $\pi_\theta(\cdot \mid x,y_{<t})$. Taking the expectation of $\log \pi_\theta(y_t \mid x,y_{<t})$ under this distribution yields:

\begin{flalign}
& \mathbb{E}_{y_t \sim \pi_\theta(\cdot \mid x,y_{<t})}
\Big[
\log \pi_\theta(y_t \mid x,y_{<t})
\Big]
\nonumber\\
&= \sum_{v}
\pi_\theta(v \mid x,y_{<t})
\log \pi_\theta(v \mid x,y_{<t})
\nonumber\\
&= - H\!\left(\pi_\theta(\cdot \mid x,y_{<t})\right).
\label{eq:per_step_log_vs_entropy}
\end{flalign}

Now consider a full response $y_{1:T} \sim p_\theta(\cdot \mid x)$. By the law of iterated expectations,

\begin{flalign}
& \mathbb{E}_{y_{1:T} \sim p_\theta(\cdot \mid x)}
\Big[
\log \pi_\theta(y_t \mid x,y_{<t})
\Big]
\nonumber\\
&= \mathbb{E}_{y_{<t} \sim p_\theta(\cdot \mid x)}
\Big[
\mathbb{E}_{y_t \sim \pi_\theta(\cdot \mid x,y_{<t})}
\big[
\log \pi_\theta(y_t \mid x,y_{<t})
\big]
\Big]
\nonumber\\
&= -\mathbb{E}_{y_{<t} \sim p_\theta(\cdot \mid x)}
\Big[
H\!\left(\pi_\theta(\cdot \mid x,y_{<t})\right)
\Big].
\label{eq:per_step_on_policy}
\end{flalign}

where we used Eq.~\ref{eq:per_step_log_vs_entropy} in the last step. Averaging over $t=1,\dots,T$ and dividing by $T$ gives
\begin{flalign}
& \mathbb{E}_{y_{1:T} \sim p_\theta(\cdot \mid x)}
\Big[
\mathrm{AvgLogProb}(x,y_{1:T})
\Big]
\nonumber\\
&= \mathbb{E}_{y_{1:T} \sim p_\theta(\cdot \mid x)}
\Big[
\frac{1}{T}\sum_{t=1}^{T}
\log \pi_\theta(y_t \mid x,y_{<t})
\Big]
\nonumber\\
&= -\mathbb{E}_{y_{1:T} \sim p_\theta(\cdot \mid x)}
\Big[
\frac{1}{T}\sum_{t=1}^{T}
H\!\left(\pi_\theta(\cdot \mid x,y_{<t})\right)
\Big]
\nonumber\\
&= -\mathbb{E}_{y_{1:T} \sim p_\theta(\cdot \mid x)}
\Big[
\mathrm{RespEnt}(x,y_{1:T})
\Big].
\label{eq:onpolicy_avglog_vs_respH}
\end{flalign}

Finally, taking expectation over prompts $x \sim p(x)$ on both sides of Eq.~\ref{eq:onpolicy_avglog_vs_respH} yields
\begin{equation}
    \mathcal{L}_{\text{avg-log}}(\theta)
    =
    -\mathcal{H}_{\text{resp}}(\theta),
\end{equation}
which is exactly Eq.~\ref{eq:avglog_vs_respH_exact}. This shows that under on-policy sampling, the two token-level metrics are related by a fixed negative sign and thus cannot be decoupled in the action space.
\end{proof}

\paragraph{Hidden-state Effective Rank and velocity.}
The next proposition shows that these two hidden-state metrics in Sec.~\ref{sec:static metric} and~\ref{def_velo} are \emph{structurally decoupled} at the level of trajectories: knowing the final Effective Rank alone does not determine ERV, and conversely.

\begin{proposition}[Hidden-state metrics are structurally decoupled]
\label{prop:new_decoupled_strong}
Fix $K \ge 3$ evaluation steps. Consider the map that associates to each Effective Rank trajectory $m = (m_1,\dots,m_K) \in \mathbb{R}^K$ its final value
\begin{equation}
    \mathrm{ER}_{\text{final}}(m) := m_K
\end{equation}
and its Effective Rank velocity
\begin{equation}
    \mathrm{ERV}(m)
    :=
    \frac{1}{K-1}\sum_{j=2}^{K}
        \Big(
            m_j - \frac{1}{j-1}\sum_{k=1}^{j-1} m_k
        \Big).
\end{equation}
Then:
\begin{enumerate}
    \item There is no function $f : \mathbb{R} \to \mathbb{R}$ such that
    $\mathrm{ERV}(m) = f(\mathrm{ER}_{\text{final}}(m))$ for all
    trajectories $m \in \mathbb{R}^K$.
    \item There is no function $g : \mathbb{R} \to \mathbb{R}$ such that
    $\mathrm{ER}_{\text{final}}(m) = g(\mathrm{ERV}(m))$ for all
    trajectories $m \in \mathbb{R}^K$.
\end{enumerate}
Equivalently, $\mathrm{ER}_{\text{final}}$ and $\mathrm{ERV}$ are not functionally dependent: they capture genuinely different aspects of the Effective Rank sequence.
\end{proposition}

\begin{proof}
We view $\mathrm{ER}_{\text{final}}$ and $\mathrm{ERV}$ as real-valued functions on $\mathbb{R}^K$. The proof is purely algebraic and does not rely on any monotonicity of $m_j$.

Step 1: ERV is a non-trivial linear functional. Introduce the shorthand
\begin{equation}
    \Delta m_j := m_j - m_{j-1},
    \quad j \ge 2.
\end{equation}
A direct calculation shows that each increment $\delta_j$ can be written as
\begin{equation}
    \delta_j
    =
    \frac{1}{j-1}\sum_{r=2}^{j} (r-1)\,\Delta m_r,
    \quad j \ge 2,
\end{equation}
so that ERV is a linear functional of $m$:
\begin{equation}
    \mathrm{ERV}(m)
    =
    \sum_{j=1}^{K} \alpha_j m_j,
\end{equation}
for some fixed coefficients $\alpha_1,\dots,\alpha_K$ that depend only on $K$ (and $s$) and satisfy $\sum_{j=1}^{K} \alpha_j = 0$ and $\alpha_j \neq 0$ for at least two indices $j$ (e.g.\ $\alpha_1 \neq 0$ and $\alpha_K \neq 0$). In particular, $\mathrm{ERV}$ is \emph{not} proportional to the projection onto any single coordinate $m_j$.

Step 2: No functional dependence of ERV on the final ER. Suppose, for contradiction, that there exists a function
$f : \mathbb{R} \to \mathbb{R}$ such that
\begin{equation}
\begin{aligned}
\mathrm{ERV}(m) &= f(\mathrm{ER}_{\text{final}}(m)) \\
&= f(m_K)\quad\text{for all } m \in \mathbb{R}^K .
\end{aligned}
\end{equation}

Fix any constant $c \in \mathbb{R}$. Consider the affine subspace
\begin{equation}
    \mathcal{A}_c := \{m \in \mathbb{R}^K : m_K = c\}.
\end{equation}
On this subspace, $\mathrm{ER}_{\text{final}}(m) \equiv c$ is
constant, so by assumption $\mathrm{ERV}(m) \equiv f(c)$ must also
be constant. However, $\mathrm{ERV}$ is a non-trivial linear functional
that depends on at least one coordinate $m_j$ with $j < K$. Therefore,
restricted to $\mathcal{A}_c$, the map $m \mapsto \mathrm{ERV}(m)$
varies with those coordinates and cannot be constant. This yields a
contradiction. Hence no such $f$ exists.

Step 3: No functional dependence of final ER on ERV. The argument is symmetric. Suppose there exists
$g : \mathbb{R} \to \mathbb{R}$ such that
\begin{equation}
\begin{aligned}
m_K &= \mathrm{ER}_{\text{final}}(m) \\
&= g(\mathrm{ERV}(m))\quad\text{for all } m \in \mathbb{R}^K .
\end{aligned}
\end{equation}
Fix any constant $c \in \mathbb{R}$ and consider the affine subspace
\begin{equation}
    \mathcal{B}_c := \{m \in \mathbb{R}^K : \mathrm{ERV}(m) = c\}.
\end{equation}
Since $\mathrm{ERV}$ is a non-trivial linear functional, $\mathcal{B}_c$
is an affine hyperplane of codimension 1, and $m_K$ can vary freely
among its points. Yet the assumed relation
$m_K = g(\mathrm{ERV}(m)) = g(c)$ would force $m_K$ to be constant on
$\mathcal{B}_c$, which is impossible. Thus no such $g$ exists.

Combining the two steps, we conclude that $\mathrm{ER}_{\text{final}}$
and $\mathrm{ERV}$ are not functionally dependent on each other.
\end{proof}

\paragraph{Summary.}
Proposition~\ref{prop:old_coupled_strong} establishes that two classical token-level metrics, average log probability and response entropy, are \emph{algebraically coupled} under on-policy sampling and therefore cannot be varied independently. In contrast, Proposition~\ref{prop:new_decoupled_strong} shows that our hidden-state metrics, terminal Effective Rank and Effective Rank Velocity, are structurally decoupled. They depend on different functionals of the Effective Rank trajectory and capture complementary aspects of exploration (semantic diversity level) and exploitation (the rate of semantic diversity gain).

\subsection{Effective Rank as Semantic Exploration}
\label{app:er_exploration}
We now formalize the interpretation of Effective Rank as a measure of semantic diversity and uncertainty in the hidden-state space, and hence as a representation-level proxy for exploration in LLM reasoning. We assume that the hidden states are semantic representations in the sense that downstream semantic properties can be approximately recovered as linear functionals of the hidden vectors. This is standard in representation learning and probing work on large language models.

\begin{assumption}[Semantic linear decodability]
\label{assump:linear_semantics}
There exists a collection of $K$ semantic features $s^{(1)},\dots,s^{(K)}$ (e.g., semantic roles, entity identities, factual attributes, intermediate reasoning states) such that for each time step $t$ and feature index $k$ we have
\begin{equation}
    s^{(k)}_t \;\approx\; w_k^\top h_t,
    \quad
    w_k \in \mathbb{R}^D.
\end{equation}
Semantic features are therefore approximately linearly decodable from hidden states.
\end{assumption}

\begin{assumption}[Bounded energy and finite support]
\label{assump:bounded_energy}
For a given trajectory $Z_{1:T}$, there exists an orthonormal basis of semantic directions $\{e_1,\dots,e_D\}$ such that each hidden state admits a decomposition
\begin{equation}
    z_t = \sum_{i=1}^{D} a_{t,i}\,e_i,
\end{equation}
with $\sum_{t=1}^{T} a_{t,i}^2 < \infty$ for all $i$, and only finitely many coordinates $a_{t,i}$ carry task-relevant semantic variation.
\end{assumption}

\begin{proposition}[Effective Rank as semantic diversity and uncertainty]
\label{prop:er_exploration}
Let $Z_{1:T}$ have singular values $\{\sigma_i\}_{i=1}^{\min(T,D)}$ (i.e., the diagonal of $\Sigma$ in its SVD), and let $\mathrm{ER}(Z_{1:T})$ denote its Effective Rank.
\begin{enumerate}
    \item If the trajectory uses exactly $k$ orthogonal semantic directions with equal energy and no others, i.e.\ the singular values satisfy $\sigma_1 = \dots = \sigma_k > 0$ and $\sigma_{k+1} = \dots = \sigma_r = 0$, then $\mathrm{ER}(Z_{1:T}) = k$.
    \item More generally, if the singular value spectrum of $Z_{1:T}$ becomes more \emph{spread out} over more directions in the sense of majorization (i.e.\ the normalized singular value vector becomes more uniform over a larger support), then $\mathrm{ER}(Z_{1:T})$ increases.
\end{enumerate}
Consequently, $\mathrm{ER}(Z_{1:T})$ is a basis-invariant, strictly increasing measure of the number of independent semantic directions that are effectively used by the hidden states, and thus a natural representation-level proxy for semantic exploration and uncertainty.
\end{proposition}

\begin{proof}
We proceed in two parts.

\paragraph{(1) Equal-energy $k$-dimensional semantic subspace.}
Suppose $Z_{1:T}$ uses exactly $k$ orthogonal semantic directions with equal energy. Then, up to permutation, the non-zero singular values satisfy
\begin{equation}
    \sigma_1 = \dots = \sigma_k = c > 0,
    \qquad
    \sigma_{k+1} = \dots = \sigma_r = 0.
\end{equation}
The normalized singular values are thus
\begin{equation}
    q_i =
    \begin{cases}
        1/k, & i=1,\dots,k,\\
        0,   & i>k,
    \end{cases}
\end{equation}
and the entropy of $q$ is
\begin{equation}
    H(q) = -\sum_{i=1}^{k} \frac{1}{k} \log \frac{1}{k}
    = \log k.
\end{equation}
Therefore
\begin{equation}
    \mathrm{ER}(Z_{1:T}) = \exp(H(q)) = \exp(\log k) = k.
\end{equation}
This shows that, in the idealized case of exactly $k$ equi-energic semantic directions, Effective Rank matches the true semantic dimensionality $k$.

\paragraph{(2) Monotonicity under majorization.}
Consider two hidden-state trajectories $Z$ and $\tilde Z$ with normalized
singular value spectra $q$ and $\tilde q$, respectively. Suppose that
$q$ is \emph{majorized} by $\tilde q$ (denoted $q \prec \tilde q$),
meaning intuitively that $\tilde q$ is ``more spread out'' and therefore
more uniform across a larger support.

It is a standard result in information theory that the Shannon entropy
$H(\cdot)$ is Schur-concave: if $q \prec \tilde q$, then
$H(q) \le H(\tilde q)$ with strict inequality whenever $q \neq \tilde q$.
Therefore
\begin{equation}
    \mathrm{ER}(Z) = \exp(H(q))
    \;\le\;
    \exp(H(\tilde q))
    = \mathrm{ER}(\tilde Z),
\end{equation}
with strict inequality when the majorization is strict. In words, whenever
the singular value spectrum becomes more spread out across more directions,
the Effective Rank strictly increases.

\paragraph{Summary.}
Combining the two parts, we see that Effective Rank equals the number of equi-energic semantic directions in the idealized case and increases whenever the representation distributes energy over more orthogonal directions. Since, by Assumption~\ref{assump:linear_semantics}, semantic features are linearly decodable along such directions, $\mathrm{ER}(Z_{1:T})$ provides a basis-invariant measure of how many independent semantic dimensions are explored by the hidden states and how evenly they are used. This justifies its interpretation as a representation-level exploration and uncertainty metric.
\end{proof}

\subsection{Effective Rank Velocity as Semantic Exploitation}
\label{app:erv_exploitation}

Building on Sec.~\ref{sec:static metric}, where Effective Rank (ER) is shown to measure the number and uniform use of semantic directions in the hidden-state space, we now give a representation-only justification for interpreting Effective Rank Velocity (ERV) as \emph{semantic exploitation}.

Throughout this subsection we fix a single trajectory $Z_{1:T}$ and a stride $s$. Let the evaluation positions be $t_j = j s$ for $j = 1,\dots,K$ with $K = \lfloor (T-1)/s \rfloor$, and write 
\begin{equation}
    m_j \;:=\; \text{ER}\bigl(Z_{1:t_j}\bigr), \qquad j=1,\dots,K .
\end{equation}

Thus, $\{m_j\}_{j=1}^K$ is the ER trajectory of the growing prefixes of the same response.

\paragraph{ERV as a recency-weighted sum of ER increments.}

For convenience we recall the notation of Def.~\ref{def_velo} with $M = \text{ER}$. Define the local ER increments

\begin{equation}
    \Delta m_r \;:=\; m_r - m_{r-1},\quad r \ge 2.
\end{equation}

Def.~\ref{def_velo} introduces the ``historical-baseline deviation''

\begin{equation}
    \delta_j \;:=\; m_j - \frac{1}{j-1} \sum_{k=1}^{j-1} m_k,
    \qquad j \ge 2,
\end{equation}

and the first-order temporal difference (ERV) as

\begin{equation}
    \Delta^{(1)}_{\text{ER}}
    \;:=\;
    \frac{1}{K-1}\sum_{j=2}^K \delta_j.
\end{equation}

The following lemma makes explicit that ERV is a recency-weighted average of the consecutive ER increments.

\begin{lemma}[Recency-weighted velocity of ER]
\label{lem:erv_weighted_sum}
For any sequence $(m_j)_{j=1}^K$, the historical-baseline deviation admit the representation
\begin{equation}
\label{eq:delta_weighted_sum}
\delta_j
=
\frac{1}{j-1}\sum_{r=2}^j (r-1)\,\Delta m_r,
\qquad j \ge 2,
\end{equation}
and hence ERV can be written as
\begin{equation}
\label{eq:erv_weighted_sum}
\begin{aligned}
\Delta^{(1)}_{\text{ER}}
&=
\sum_{r=2}^K w_r\,\Delta m_r,\\
w_r
&:=
\frac{r-1}{K-1}\sum_{j=r}^K \frac{1}{j-1} > 0.
\end{aligned}
\end{equation}
In particular, ERV is a positive linear combination of the local ER increments $\Delta m_r$, assigning larger weights to more recent steps.
\end{lemma}

\begin{proof}
Eq.~\ref{eq:delta_weighted_sum} is exactly in Def.~\ref{def_velo} with $M=\text{ER}$, obtained by expressing $\delta_j$ in terms of the increments $\Delta m_r$ via telescoping. Plugging Eq.~\ref{eq:delta_weighted_sum} into the definition of $\Delta^{(1)}_{\text{ER}}$ and exchanging the order of summation yields
\begin{equation}
\begin{aligned}
\Delta^{(1)}_{\text{ER}}
&=
\frac{1}{K-1}
\sum_{j=2}^K
\frac{1}{j-1}\sum_{r=2}^j (r-1)\,\Delta m_r\\
&=
\sum_{r=2}^K \left[
\frac{r-1}{K-1}\sum_{j=r}^K \frac{1}{j-1}
\right]\Delta m_r.
\end{aligned}
\end{equation}
which gives Eq.~\ref{eq:erv_weighted_sum} with $w_r$ as stated. Since $r-1>0$ and the harmonic tail $\sum_{j=r}^K (j-1)^{-1}$ is positive, we have $w_r>0$ for all $r$.
\end{proof}

\paragraph{Semantic exploitation as positive ER drift in a fixed semantic subspace.}

Sec.~\ref{app:prop_old_new} states that, up to an orthonormal change of basis, the hidden states can be written as $h_t = \sum_i a_{t,i} e_i$ with bounded energy along each semantic direction $e_i$, and that ER is a strictly increasing, basis-invariant measure of how many semantic directions are effectively used and how evenly energy is distributed among them. In particular, if the set of active directions (support of the singular value spectrum) is kept fixed and the spectrum becomes more uniform (in the sense of majorization), then ER strictly increases. Motivated by this, we isolate an idealized \emph{semantic exploitation} regime in which the trajectory has already selected a semantic subspace and is refining it.

\begin{definition}[Semantic exploitation regime]
\label{def:semantic_exploitation_regime}
Let $(m_j)_{j=1}^K$ be the ER trajectory of a response, and let $q^{(j)}$ denote the normalized singular value vector of $Z_{1:t_j}$. We say that steps $j=2,\dots,K$ form a \emph{semantic exploitation regime with rate $\mu>0$} if:

\begin{enumerate}
\item (Fixed semantic support) The support of $q^{(j)}$ is independent of $j$, i.e., the set of active semantic directions is fixed.
\item (Uniformization within the support) For every $j\ge 2$, $q^{(j)}$ is more uniform than $q^{(j-1)}$ on this fixed support, in the sense of majorization, so that by Prop.~\ref{prop:er_exploration} we have $m_j - m_{j-1} = \Delta m_j \ge \mu$ for some $\mu>0$.
\end{enumerate}

Intuitively, condition (i) says that the model has committed to a particular semantic subspace (a line of reasoning), and condition (ii) says that it keeps redistributing energy within this subspace to make use of all its semantic directions more evenly. This is precisely the notion of ``refining a promising strategy'' in representation space.
\end{definition}

\paragraph{ERV lower-bounds the semantic exploitation rate.}

Under Def.~\ref{def:semantic_exploitation_regime}, ER experiences a persistent positive drift along the trajectory. The next proposition shows that ERV is a quantitative lower bound on this drift, and thus a natural measure of semantic exploitation strength.

\begin{proposition}[ERV as a lower bound on semantic exploitation rate]
\label{prop:erv_lower_bound}
Assume the hidden states satisfy Assumptions~\ref{assump:linear_semantics} and~\ref{assump:bounded_energy} and that steps $j=2,\dots,K$ form a semantic exploitation regime with rate $\mu>0$ in the sense of Def.~\ref{def:semantic_exploitation_regime}, so that $\Delta m_j \ge \mu$ for all $j\ge 2$. Then
\begin{equation}
\label{eq:erv_lower_bound}
\Delta^{(1)}_{\text{ER}}
\;\ge\;
\frac{\mu K}{4}.
\end{equation}
In particular, ERV is strictly positive and grows linearly with the length $K$ of the exploitation segment.
\end{proposition}

\begin{proof}
By Eq.~\ref{eq:delta_weighted_sum} and the assumption $\Delta m_r \ge \mu$ we obtain, for each $j\ge 2$,
\begin{equation}
\begin{aligned}
\delta_j
&=
\frac{1}{j-1}\sum_{r=2}^j (r-1)\,\Delta m_r\\
&\ge
\frac{1}{j-1}\sum_{r=2}^j (r-1)\,\mu\\
&=
\frac{\mu}{j-1}\sum_{r=2}^j (r-1)
=
\frac{\mu j}{2}.
\end{aligned}
\end{equation}
Averaging over $j$ then yields
\begin{equation}
\begin{aligned}
\Delta^{(1)}_{\text{ER}}
&=
\frac{1}{K-1}\sum_{j=2}^K \delta_j\\
&\ge
\frac{1}{K-1}\sum_{j=2}^K \frac{\mu j}{2}
=
\frac{\mu}{2(K-1)}\sum_{j=2}^K j.
\end{aligned}
\end{equation}
For $K\ge 2$ we haven $\sum_{j=2}^K j \ge \frac{K(K-1)}{2}$, so
\begin{equation}
\Delta^{(1)}_{\text{ER}}
\;\ge\;
\frac{\mu}{2(K-1)}\cdot \frac{K(K-1)}{2}
=
\frac{\mu K}{4},
\end{equation}
which proves Eq.~\ref{eq:erv_lower_bound}.
\end{proof}

Thus, in an idealized regime where the model has already discovered a useful semantic subspace and is consistently enriching it, ERV provides a strictly positive, linearly growing lower bound on the rate at which semantic complexity within that subspace is being exploited.

\section{Implementation Details}
\label{appendix:Implement_Details}
\subsection{Training and Evaluation Details}
We typically use the same set of hyperparameters to train and evaluate all models in the SimpleRL-Zoo series~\citep{zeng2025simplerlzoo} in the default main experiment setting.

\paragraph{Training. } We conduct all experiments with 4 A800-PCIE-80GB GPUs. For GRPO and PPO, we use a prompt batch size of 48 with a maximum rollout length of 1536 tokens. Training is performed using a mini-batch size of 24. For GRPO, we generate 4 rollouts per prompt. For PPO, we use DeepSeek-R1-Distill-Qwen-1.5B~\citep{guo2025deepseek} as the value model and generate 1 rollout per prompt.  The default sampling temperature is set to 1.0, and the clip ratio is 0.2. For all actor models ranging from 3B to 8B parameters, we use a learning rate of 1e-6 and a KL loss coefficient of 1e-4. For critic models in PPO, we use a learning rate of 1e-5. For our training datasets, we follow the same setup as in \citet{zeng2025simplerlzoo}, where the data is filtered from GSM8K~\citep{cobbe2021training} and MATH~\citep{hendrycks2021measuring} configured with different difficulty levels for models of varying capabilities. We tested using the checkpoint model trained up to 120 steps.

\paragraph{Evaluation. } We build our evaluation script based on that of \citet{zeng2025simplerlzoo}, using a temperature of 0.6 and a maximum generation length of 2048 tokens. To ensure consistency, we adopt the same prompt template used during training. For most benchmarks, we report Pass@1 results. However, for benchmarks like AIME 2024, which contains fewer problems, we report both Pass@1 and average accuracy (Pass@256), computed over 256 generated samples per problem.

\paragraph{Base Models. } 
To demonstrate the universality of our insights and methods, we conduct zero RL training experiments on Llama-3.2 (3B), Llama-3.1 (8B)~\citep{grattafiori2024llama}, Mistral-v0.3-7B~\citep{jiang2024mixtral}, and Qwen-2.5 (1.5B, 3B, 7B)~\citep{hui2024qwen2}. For value model in PPO, we use DeepSeek-R1-Distill-Qwen-1.5B~\citep{guo2025deepseek} for all experiments.

\paragraph{Benchmark. }
We evaluate on a diverse suite of mathematical reasoning benchmarks. These include standard benchmarks such as GSM8K~\citep{cobbe2021training}, MATH~\citep{hendrycks2021measuring}, ASDiv~\citep{miao2020diverse}, Carp (English Version)~\citep{zhang2023evaluating}, MAWPS~\citep{koncel2016mawps}, SVAMP~\citep{patel2021nlp}, TabMWP~\citep{lu2023dynamic}, and OlympiadBench~\citep{he2024olympiadbench}; Chinese mathematics collections like CMATH~\citep{wei2023cmath} and Gaokao 2024; and benchmarks from mathematics competitions, including the 2024/2025 AIME and the 2023/2024 AMC.

\begin{figure*}[t]
    
    \centering 
    
    \includegraphics[width=\textwidth]{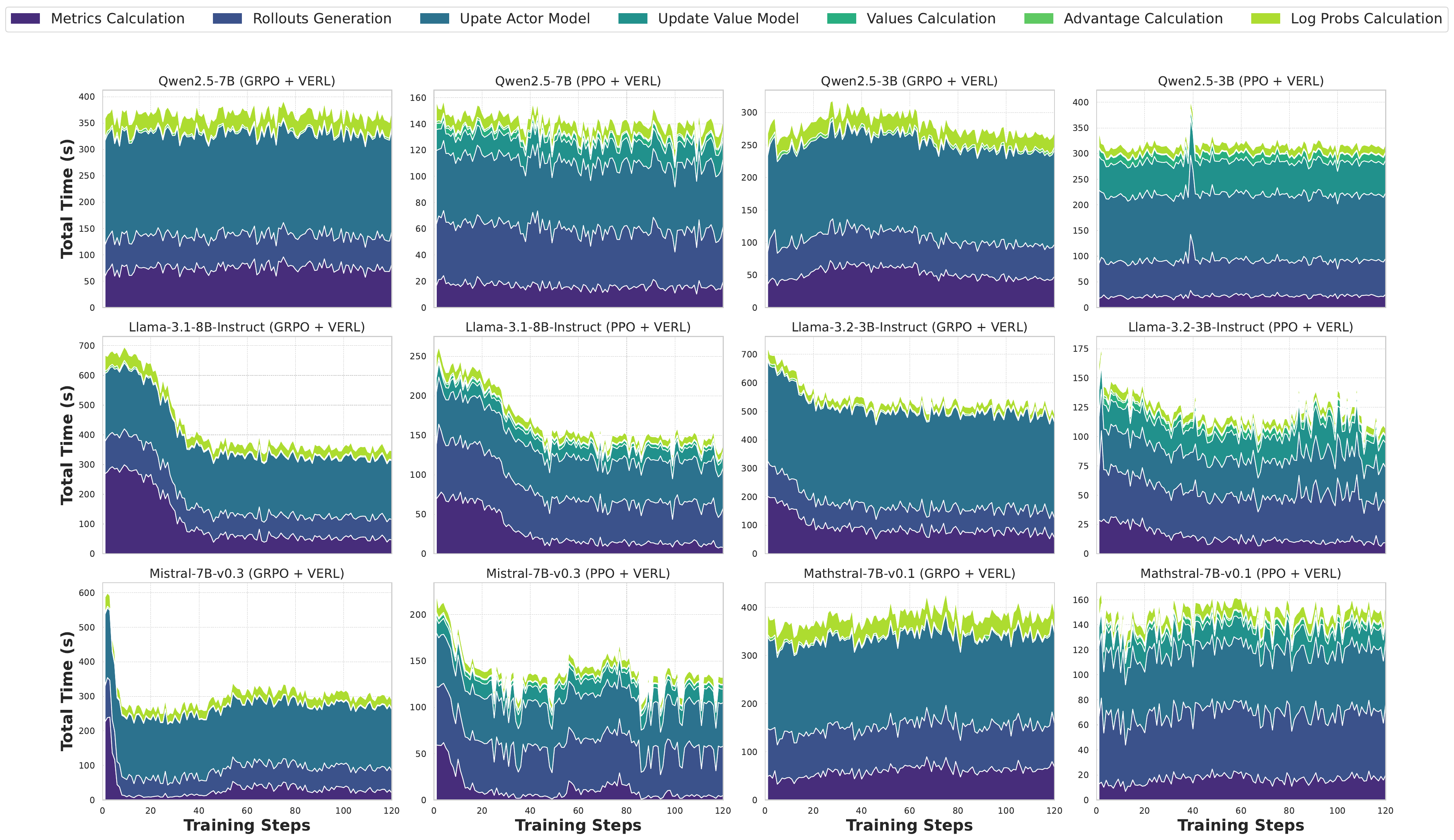}
    
    \caption{Time overhead of the main computation of RL Training.}
    
    \label{time_overhead}
\end{figure*}

\begin{figure*}[htbp]
    
    \centering 
    
    \includegraphics[width=\textwidth]{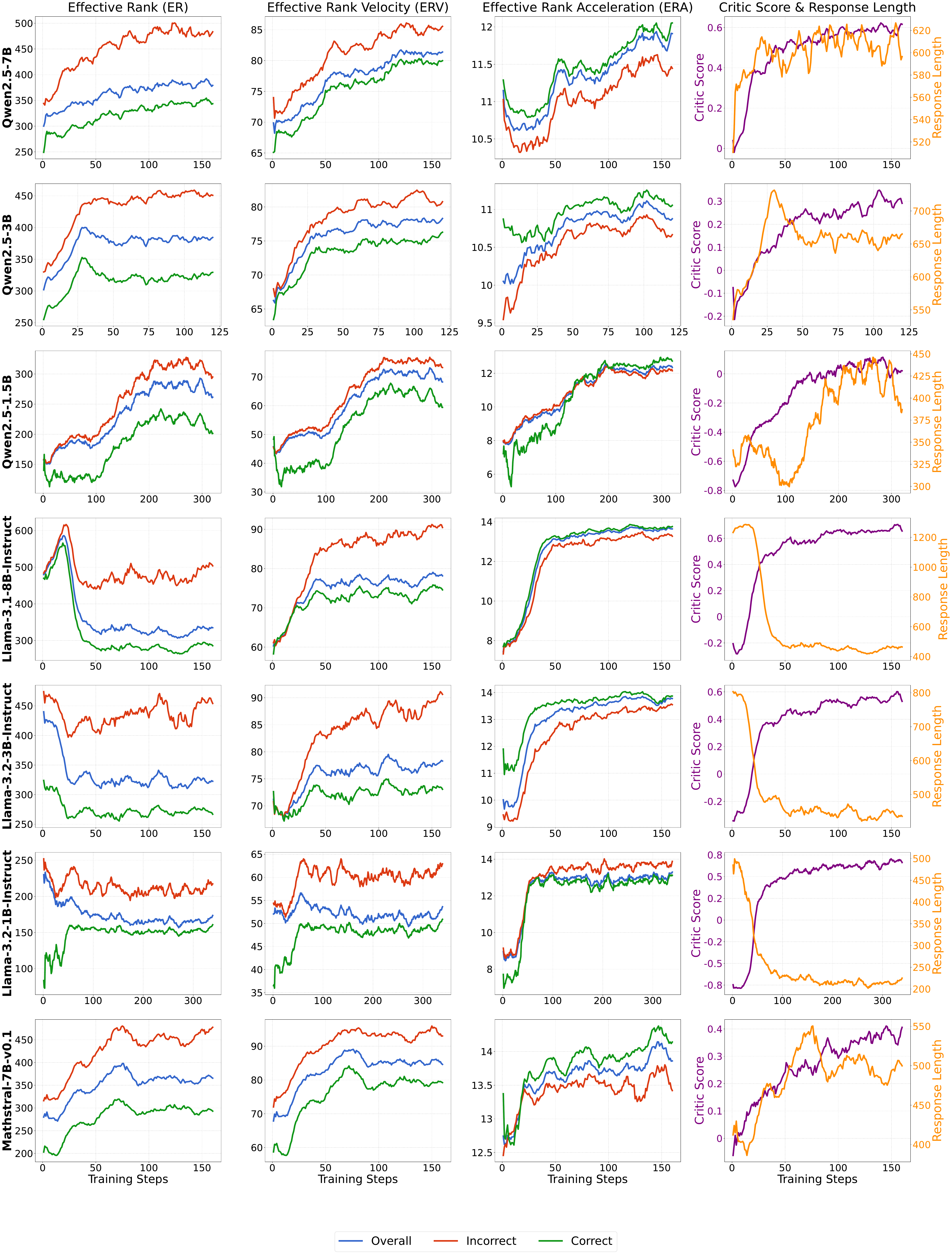}
    
    \caption{Response-level metrics during GRPO post-training, smoothed with a 10-step rolling window. Metrics are shown for the \textcolor{blue}{Overall} batch, as well as for subsets of \textcolor[RGB]{0,140,0}{Correct} and \textcolor{red}{Incorrect} samples. The rightmost column displays the average \textcolor[RGB]{107,30,120}{Critic Score} (reward) and \textcolor{orange}{Response Length} per batch.}
    
    \label{Appendix_observation1_grpo}
\end{figure*}

\begin{figure*}[htbp]
    
    \centering 
    
    \includegraphics[width=\textwidth]{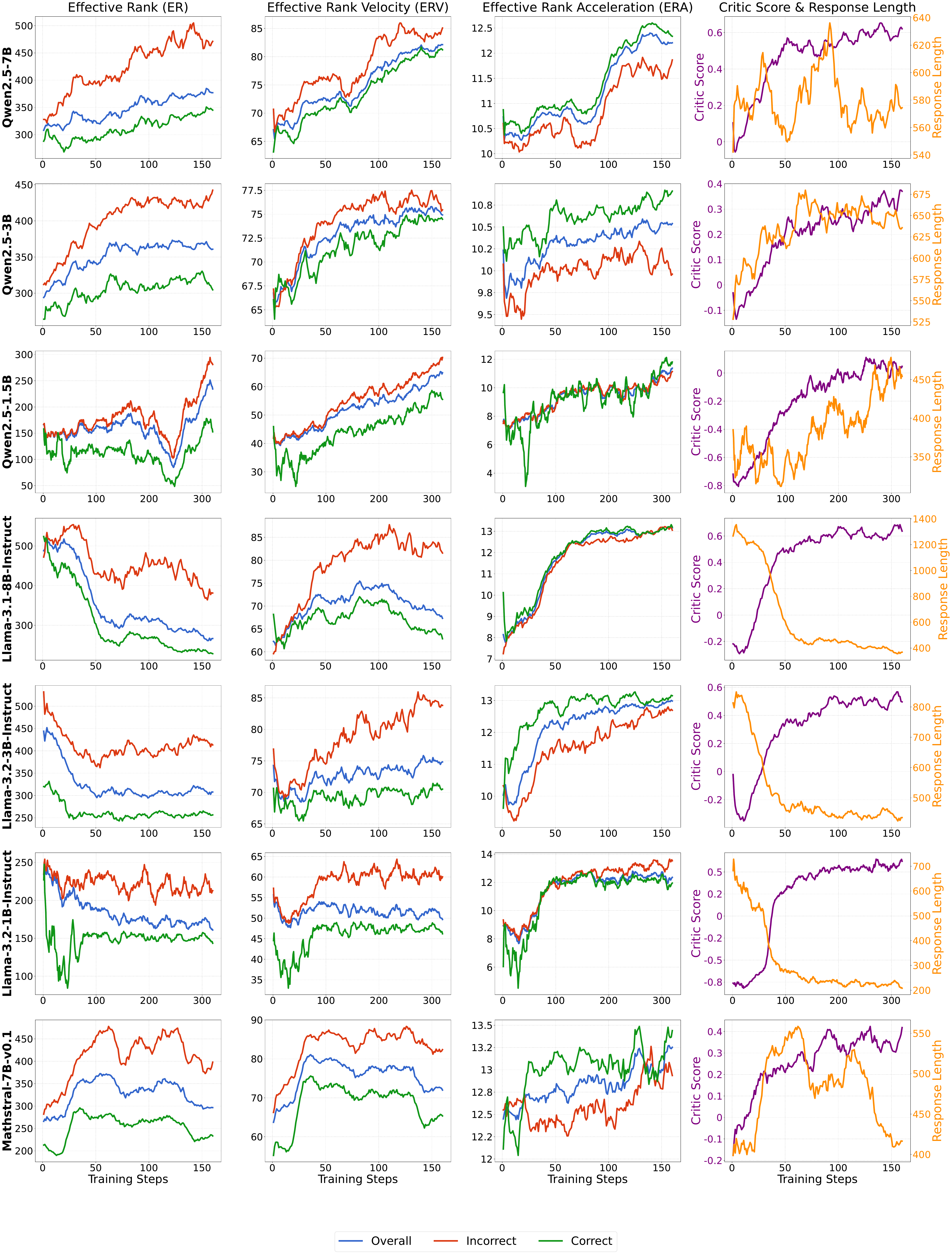}
    
    \caption{Response-level metrics during PPO post-training, smoothed with a 10-step rolling window. Metrics are shown for the \textcolor{blue}{Overall} batch, as well as for subsets of \textcolor[RGB]{0,140,0}{Correct} and \textcolor{red}{Incorrect} samples. The rightmost column displays the average \textcolor[RGB]{107,30,120}{Critic Score} (reward) and \textcolor{orange}{Response Length} per batch.}
    
    \label{Appendix_observation1_ppo}
\end{figure*}

\begin{figure*}[htbp]
    
    \centering 
    
    \includegraphics[width=\textwidth]{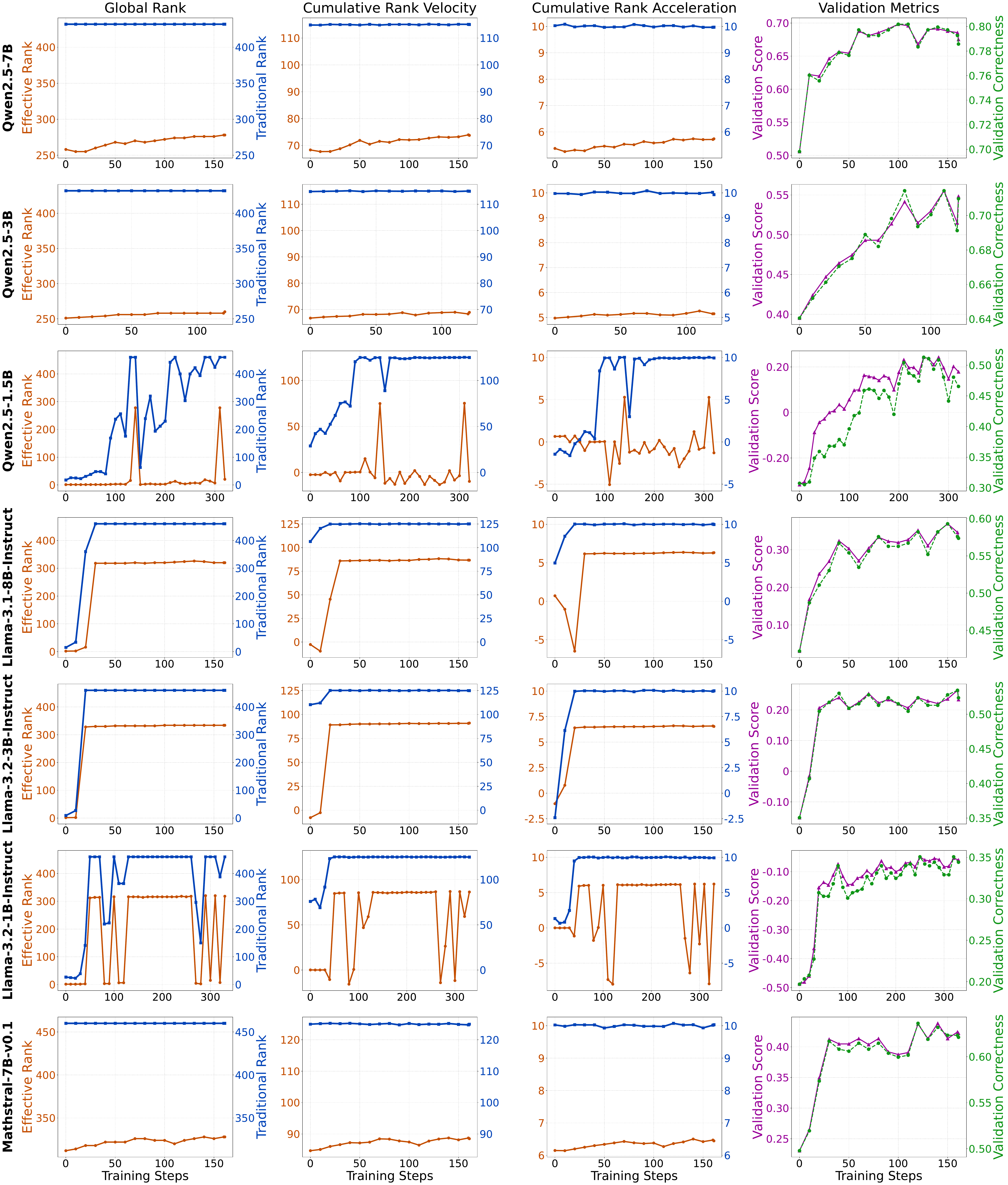}
    
    \caption{Visualization of dataset-level metrics during GRPO post-training. The figure compares \textcolor[RGB]{36,65,171}{Traditional} metrics with our \textcolor{brown}{proposed} metrics. Also shown are the \textcolor[RGB]{107,30,120}{Validation Score} and sample \textcolor[RGB]{0,140,0}{Correctness}, both averaged over the validation dataset.}
    
    \label{Appendix_observation2_grpo}
\end{figure*}

\begin{figure*}[htbp]
    
    \centering 
    
    \includegraphics[width=\textwidth]{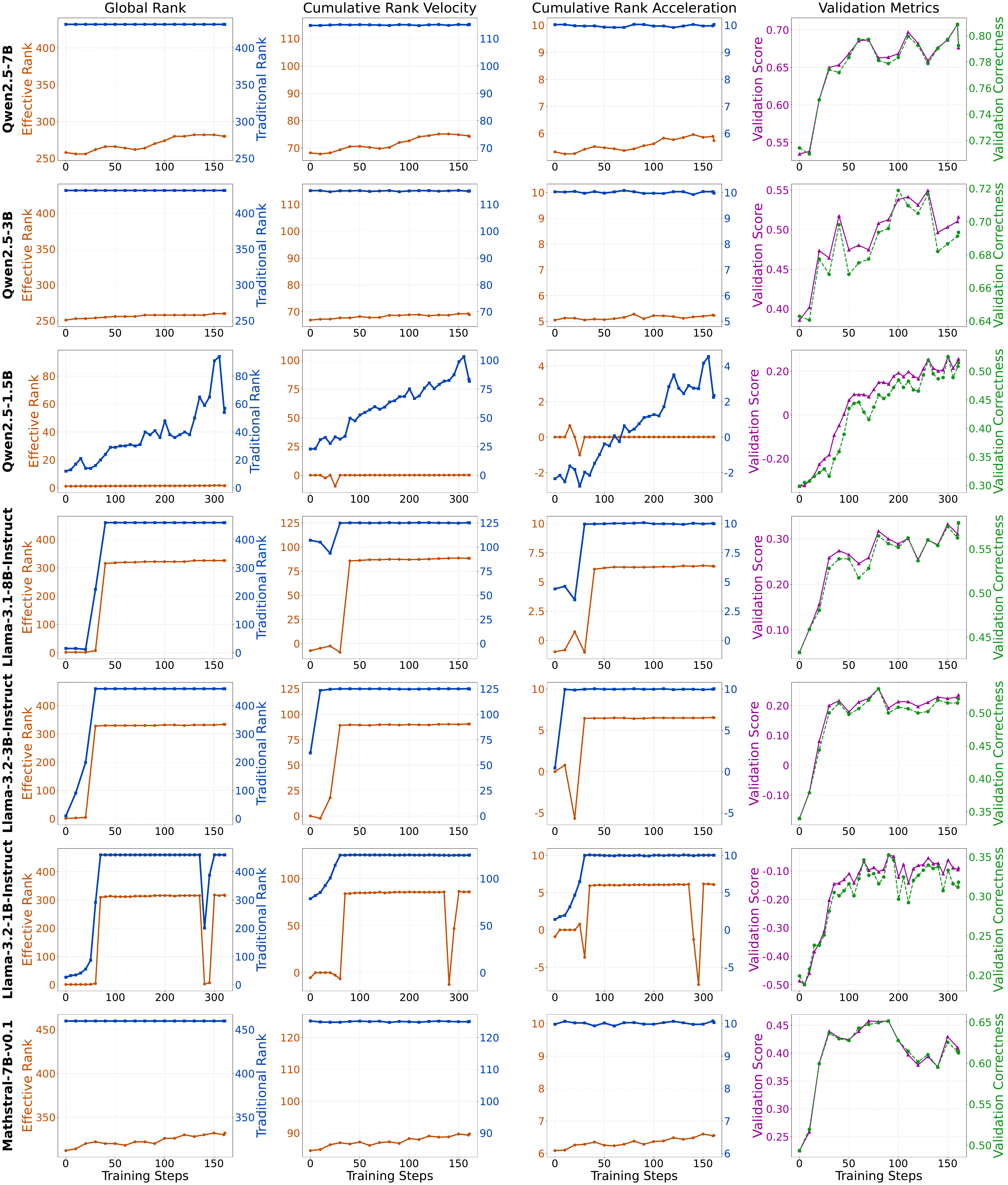}
    
    \caption{Visualization of dataset-level metrics during PPO post-training. The figure compares \textcolor[RGB]{36,65,171}{Traditional} metrics with our \textcolor{brown}{proposed} metrics. Also shown are the \textcolor[RGB]{107,30,120}{Validation Score} and sample \textcolor[RGB]{0,140,0}{Correctness}, both averaged over the validation dataset.}
    
    \label{Appendix_observation2_ppo}
\end{figure*}

\begin{table*}[ht]
\centering
\caption{Performance comparison of instruction-tuned models on mathematical reasoning benchmarks (Pass@1). 
“+ GRPO” and “+ PPO” denote reinforcement learning fine-tuning from the base model using GRPO and PPO, respectively. 
“w/ \ourmethod{}.” indicates the application of our \ourmethod{}-based advantage to the corresponding RL algorithm. 
$\Delta$ represents the performance difference between the baseline RL method and its \ourmethod{}-advanced variant. 
All results are reported in percentage (\%).}
\begin{adjustbox}{width=\textwidth,center}
\begin{tabular}{lcccccccccccccccc}
\toprule
\textbf{Model} & 
\makecell[c]{\textbf{AIME24}} & 
\makecell[c]{\textbf{AIME25}} & 
\makecell[c]{\textbf{AMC23}} & 
\makecell[c]{\textbf{AMC24}} & 
\makecell[c]{\textbf{ASDiv}} & 
\makecell[c]{\textbf{Carp\_En}} & 
\makecell[c]{\textbf{CMATH}} & 
\makecell[c]{\textbf{Gaokao}\\ \textbf{2024\_I}} & 
\makecell[c]{\textbf{Gaokao}\\ \textbf{2024\_Mix}} & 
\makecell[c]{\textbf{Gaokao}\\ \textbf{MathCloze}} &
\makecell[c]{\textbf{GSM8K}} & 
\makecell[c]{\textbf{MAWPS}} & 
\makecell[c]{\textbf{Olympiad}\\ \textbf{Bench}} & 
\makecell[c]{\textbf{SVAMP}} & 
\makecell[c]{\textbf{TabMWP}} & 
\makecell[c]{\textbf{Avg.}} \\
\midrule

\textbf{Llama-3.2-3B-Instruct} 
& 0.0 & 0.0 & 25.0 & 11.1 & 74.6 & 26.5 & 10.2 & 14.3 & 14.3 & 6.8 & 66.6 & 86.9 & 12.7 & 74.1 & 41.4 & 31.0 \\
+ GRPO 
& 3.3 & 0.0 & 27.5 & 8.9 & 88.8 & 45.0 & 28.3 & 21.4 & 20.9 & 23.7 & 80.7 & 96.0 & 16.7 & 87.7 & 71.7 & 41.4 \\
\rowcolor{blue!10} + GRPO w/ \ourmethod{}. 
& 13.3 & 6.7 & 25.0 & 11.1 & 89.3 & 45.4 & 46.2 & 14.3 & 22.0 & 22.9 & 81.7 & 96.0 & 17.6 & 87.8 & 72.3 & 43.4 \\
$\Delta_{\text{GRPO}}$ 
& \textcolor{positive}{+10.0} & \textcolor{positive}{+6.7} & \textcolor{negative}{-2.5} & \textcolor{positive}{+2.2} & \textcolor{positive}{+0.5} & \textcolor{positive}{+0.4} & \textcolor{positive}{+17.9} & \textcolor{negative}{-7.1} & \textcolor{positive}{+1.1} & \textcolor{negative}{-0.8} & \textcolor{positive}{+1.0} & \textcolor{neutral}{+0.0} & \textcolor{positive}{+0.9} & \textcolor{positive}{+0.1} & \textcolor{positive}{+0.6} & \textcolor{positive}{+2.0} \\
+ PPO 
& 10.0 & 3.3 & 22.5 & 13.3 & 87.9 & 46.4 & 21.2 & 7.1 & 16.5 & 20.3 & 81.4 & 95.5 & 17.8 & 86.8 & 71.0 & 40.1 \\
\rowcolor{blue!10} + PPO w/ \ourmethod{}. 
& 10.0 & 3.3 & 25.0 & 11.1 & 88.7 & 46.0 & 30.7 & 14.3 & 19.8 & 27.1 & 82.9 & 95.7 & 17.3 & 85.8 & 71.3 & 41.9 \\
$\Delta_{\text{PPO}}$ 
& \textcolor{neutral}{+0.0} & \textcolor{neutral}{+0.0} & \textcolor{positive}{+2.5} & \textcolor{negative}{-2.2} & \textcolor{positive}{+0.8} & \textcolor{negative}{-0.4} & \textcolor{positive}{+9.5} & \textcolor{positive}{+7.2} & \textcolor{positive}{+3.3} & \textcolor{positive}{+6.8} & \textcolor{positive}{+1.5} & \textcolor{positive}{+0.2} & \textcolor{negative}{-0.5} & \textcolor{negative}{-1.0} & \textcolor{positive}{+0.3} & \textcolor{positive}{+1.9} \\
\midrule

\textbf{Llama-3.1-8B-Instruct} 
& 0.0 & 3.3 & 17.5 & 8.9 & 48.0 & 34.1 & 18.5 & 0.0 & 15.4 & 16.9 & 47.4 & 43.5 & 10.4 & 48.5 & 34.3 & 23.1 \\
+ GRPO 
& 6.7 & 0.0 & 22.5 & 15.6 & 90.3 & 42.4 & 60.7 & 7.1 & 14.3 & 32.2 & 88.4 & 96.4 & 19.7 & 88.5 & 82.7 & 44.5 \\
\rowcolor{blue!10} + GRPO w/ \ourmethod{}. 
& 10.0 & 3.3 & 32.5 & 15.6 & 90.7 & 45.0 & 72.7 & 14.3 & 14.3 & 30.5 & 88.6 & 96.9 & 21.3 & 88.4 & 83.1 & 47.2 \\
$\Delta_{\text{GRPO}}$ 
& \textcolor{positive}{+3.3} & \textcolor{positive}{+3.3} & \textcolor{positive}{+10.0} & \textcolor{neutral}{+0.0} & \textcolor{positive}{+0.4} & \textcolor{positive}{+2.6} & \textcolor{positive}{+12.0} & \textcolor{positive}{+7.2} & \textcolor{neutral}{+0.0} & \textcolor{negative}{-1.7} & \textcolor{positive}{+0.2} & \textcolor{positive}{+0.5} & \textcolor{positive}{+1.6} & \textcolor{negative}{-0.1} & \textcolor{positive}{+0.4} & \textcolor{positive}{+2.7} \\
+ PPO 
& 6.7 & 0.0 & 30.0 & 17.8 & 89.8 & 42.0 & 60.0 & 0.0 & 14.3 & 25.4 & 86.4 & 95.7 & 18.2 & 88.6 & 82.3 & 43.8 \\
\rowcolor{blue!10} + PPO w/ \ourmethod{}. 
& 10.0 & 0.0 & 35.0 & 13.3 & 90.7 & 42.6 & 62.0 & 14.3 & 22.0 & 28.8 & 87.3 & 96.6 & 19.1 & 88.1 & 83.0 & 46.2 \\
$\Delta_{\text{PPO}}$ 
& \textcolor{positive}{+3.3} & \textcolor{neutral}{+0.0} & \textcolor{positive}{+5.0} & \textcolor{negative}{-4.5} & \textcolor{positive}{+0.9} & \textcolor{positive}{+0.6} & \textcolor{positive}{+2.0} & \textcolor{positive}{+14.3} & \textcolor{positive}{+7.7} & \textcolor{positive}{+3.4} & \textcolor{positive}{+0.9} & \textcolor{positive}{+0.9} & \textcolor{positive}{+0.9} & \textcolor{negative}{-0.5} & \textcolor{positive}{+0.7} & \textcolor{positive}{+2.4} \\
\midrule

\textbf{Qwen2.5-3B} 
& 6.7 & 0.0 & 20.0 & 24.4 & 90.7 & 54.7 & 76.7 & 0.0 & 22.0 & 41.5 & 80.7 & 95.1 & 23.0 & 84.3 & 71.3 & 46.1 \\
+ GRPO 
& 3.3 & 0.0 & 40.0 & 22.2 & 92.6 & 56.0 & 82.7 & 7.1 & 27.5 & 42.4 & 82.8 & 96.6 & 23.6 & 89.0 & 81.4 & 49.8 \\
\rowcolor{blue!10} + GRPO w/ \ourmethod{}. 
& 6.7 & 0.0 & 30.0 & 17.8 & 92.6 & 56.9 & 84.8 & 21.4 & 33.0 & 49.2 & 82.2 & 96.4 & 24.4 & 88.5 & 81.0 & 51.0 \\
$\Delta_{\text{GRPO}}$ 
& \textcolor{positive}{+3.4} & \textcolor{neutral}{+0.0} & \textcolor{negative}{-10.0} & \textcolor{negative}{-4.4} & \textcolor{neutral}{+0.0} & \textcolor{positive}{+0.9} & \textcolor{positive}{+2.1} & \textcolor{positive}{+14.3} & \textcolor{positive}{+5.5} & \textcolor{positive}{+6.8} & \textcolor{negative}{-0.6} & \textcolor{negative}{-0.2} & \textcolor{positive}{+0.8} & \textcolor{negative}{-0.5} & \textcolor{negative}{-0.4} & \textcolor{positive}{+1.2} \\
+ PPO 
& 3.3 & 0.0 & 32.5 & 15.6 & 92.8 & 56.5 & 83.2 & 0.0 & 28.6 & 50.0 & 81.7 & 96.6 & 24.4 & 86.0 & 80.8 & 48.8 \\
\rowcolor{blue!10} + PPO w/ \ourmethod{}. 
& 6.7 & 0.0 & 32.5 & 17.8 & 92.6 & 57.0 & 84.3 & 21.4 & 29.7 & 47.5 & 81.8 & 96.5 & 24.6 & 88.3 & 81.4 & 50.8 \\
$\Delta_{\text{PPO}}$ 
& \textcolor{positive}{+3.4} & \textcolor{neutral}{+0.0} & \textcolor{neutral}{+0.0} & \textcolor{positive}{+2.2} & \textcolor{negative}{-0.2} & \textcolor{positive}{+0.5} & \textcolor{positive}{+1.1} & \textcolor{positive}{+21.4} & \textcolor{positive}{+1.1} & \textcolor{negative}{-2.5} & \textcolor{positive}{+0.1} & \textcolor{negative}{-0.1} & \textcolor{positive}{+0.2} & \textcolor{positive}{+2.3} & \textcolor{positive}{+0.6} & \textcolor{positive}{+2.0} \\
\midrule

\textbf{Qwen2.5-7B} 
& 6.7 & 0.0 & 45.0 & 15.6 & 91.4 & 55.8 & 86.7 & 42.9 & 33.0 & 49.2 & 85.8 & 95.4 & 25.8 & 88.5 & 82.8 & 53.6 \\
+ GRPO 
& 10.0 & 6.7 & 55.0 & 26.7 & 94.8 & 60.2 & 91.7 & 14.3 & 34.1 & 64.4 & 90.2 & 97.6 & 36.1 & 92.8 & 91.3 & 57.7 \\
\rowcolor{blue!10} + GRPO w/ \ourmethod{}. 
& 13.3 & 10.0 & 50.0 & 28.9 & 95.0 & 60.8 & 90.7 & 35.7 & 35.2 & 69.5 & 89.2 & 97.7 & 35.4 & 92.9 & 91.9 & 59.8 \\
$\Delta_{\text{GRPO}}$ 
& \textcolor{positive}{+3.3} & \textcolor{positive}{+3.3} & \textcolor{negative}{-5.0} & \textcolor{positive}{+2.2} & \textcolor{positive}{+0.2} & \textcolor{positive}{+0.6} & \textcolor{negative}{-1.0} & \textcolor{positive}{+21.4} & \textcolor{positive}{+1.1} & \textcolor{positive}{+5.1} & \textcolor{negative}{-1.0} & \textcolor{positive}{+0.1} & \textcolor{negative}{-0.7} & \textcolor{positive}{+0.1} & \textcolor{positive}{+0.6} & \textcolor{positive}{+2.1} \\
+ PPO 
& 6.7 & 3.3 & 50.0 & 33.3 & 94.9 & 59.6 & 89.8 & 28.6 & 31.9 & 63.6 & 89.1 & 97.3 & 36.1 & 92.8 & 90.8 & 57.9 \\
\rowcolor{blue!10} + PPO w/ \ourmethod{}. 
& 10.0 & 6.7 & 52.5 & 33.3 & 94.8 & 60.0 & 90.3 & 28.6 & 34.1 & 66.9 & 90.2 & 97.8 & 36.1 & 92.5 & 90.6 & 59.0 \\
$\Delta_{\text{PPO}}$ 
& \textcolor{positive}{+3.3} & \textcolor{positive}{+3.3} & \textcolor{positive}{+2.5} & \textcolor{neutral}{+0.0} & \textcolor{negative}{-0.1} & \textcolor{positive}{+0.4} & \textcolor{positive}{+0.5} & \textcolor{neutral}{+0.0} & \textcolor{positive}{+2.2} & \textcolor{positive}{+3.3} & \textcolor{positive}{+1.1} & \textcolor{positive}{+0.5} & \textcolor{neutral}{+0.0} & \textcolor{negative}{-0.3} & \textcolor{negative}{-0.2} & \textcolor{positive}{+1.1} \\
\midrule

\textbf{Mathstral-7B-v0.1} 
& 0.0 & 0.0 & 12.5 & 8.9 & 87.1 & 51.1 & 74.2 & 28.6 & 33.0 & 31.4 & 81.6 & 93.8 & 17.9 & 87.7 & 54.7 & 44.2 \\
+ GRPO 
& 0.0 & 0.0 & 47.5 & 17.8 & 92.9 & 55.9 & 81.3 & 35.7 & 44.0 & 49.2 & 88.1 & 97.6 & 25.6 & 93.0 & 81.5 & 54.0 \\
\rowcolor{blue!10} + GRPO w/ \ourmethod{}. 
& 6.7 & 0.0 & 45.0 & 20.0 & 93.3 & 55.5 & 81.5 & 50.0 & 40.7 & 46.6 & 89.5 & 97.2 & 29.3 & 90.7 & 83.5 & 55.3 \\
$\Delta_{\text{GRPO}}$ 
& \textcolor{positive}{+6.7} & \textcolor{neutral}{+0.0} & \textcolor{negative}{-2.5} & \textcolor{positive}{+2.2} & \textcolor{positive}{+0.4} & \textcolor{negative}{-0.4} & \textcolor{positive}{+0.2} & \textcolor{positive}{+14.3} & \textcolor{negative}{-3.3} & \textcolor{negative}{-2.6} & \textcolor{positive}{+1.4} & \textcolor{negative}{-0.4} & \textcolor{positive}{+3.7} & \textcolor{negative}{-2.3} & \textcolor{positive}{+2.0} & \textcolor{positive}{+1.3} \\
+ PPO 
& 6.7 & 3.3 & 32.5 & 20.0 & 90.9 & 51.8 & 78.3 & 42.9 & 37.4 & 49.2 & 87.0 & 96.0 & 28.4 & 89.9 & 70.7 & 52.3 \\
\rowcolor{blue!10} + PPO w/ \ourmethod{}. 
& 10.0 & 0.0 & 27.5 & 22.2 & 93.0 & 53.8 & 78.2 & 42.9 & 51.6 & 48.3 & 87.4 & 96.7 & 26.1 & 89.6 & 84.1 & 54.1 \\
$\Delta_{\text{PPO}}$ 
& \textcolor{positive}{+3.3} & \textcolor{negative}{-3.3} & \textcolor{negative}{-5.0} & \textcolor{positive}{+2.2} & \textcolor{positive}{+2.1} & \textcolor{positive}{+2.0} & \textcolor{negative}{-0.1} & \textcolor{neutral}{+0.0} & \textcolor{positive}{+14.2} & \textcolor{negative}{-0.9} & \textcolor{positive}{+0.4} & \textcolor{positive}{+0.7} & \textcolor{negative}{-2.3} & \textcolor{negative}{-0.3} & \textcolor{positive}{+13.4} & \textcolor{positive}{+1.8} \\
\midrule

\textbf{Mistral-7B-v0.3} 
& 0.0 & 0.0 & 10.0 & 0.0 & 40.5 & 12.4 & 21.8 & 14.3 & 13.2 & 3.4 & 24.0 & 50.8 & 1.6 & 39.1 & 30.6 & 17.4 \\
+ GRPO 
& 0.0 & 0.0 & 2.5 & 4.4 & 58.2 & 11.1 & 42.3 & 0.0 & 15.4 & 5.1 & 52.4 & 79.2 & 3.0 & 47.6 & 37.7 & 23.9 \\
\rowcolor{blue!10} + GRPO w/ \ourmethod{}. 
& 0.0 & 0.0 & 7.5 & 2.2 & 59.1 & 15.0 & 43.0 & 0.0 & 6.6 & 4.2 & 40.3 & 69.5 & 2.8 & 57.5 & 53.0 & 24.0 \\
$\Delta_{\text{GRPO}}$ 
& \textcolor{neutral}{+0.0} & \textcolor{neutral}{+0.0} & \textcolor{positive}{+5.0} & \textcolor{negative}{-2.2} & \textcolor{positive}{+0.9} & \textcolor{positive}{+3.9} & \textcolor{positive}{+0.7} & \textcolor{neutral}{+0.0} & \textcolor{negative}{-8.8} & \textcolor{negative}{-0.9} & \textcolor{negative}{-12.1} & \textcolor{negative}{-9.7} & \textcolor{negative}{-0.2} & \textcolor{positive}{+9.9} & \textcolor{positive}{+15.3} & \textcolor{positive}{+0.1} \\
+ PPO 
& 0.0 & 0.0 & 0.0 & 0.0 & 8.9 & 6.6 & 7.7 & 7.1 & 11.0 & 2.5 & 3.3 & 8.6 & 2.1 & 6.9 & 12.0 & 5.1 \\
\rowcolor{blue!10} + PPO w/ \ourmethod{}. 
& 0.0 & 0.0 & 2.5 & 0.0 & 44.7 & 10.6 & 35.7 & 7.1 & 16.5 & 5.1 & 28.8 & 70.7 & 2.4 & 57.5 & 35.1 & 21.1 \\
$\Delta_{\text{PPO}}$ 
& \textcolor{neutral}{+0.0} & \textcolor{neutral}{+0.0} & \textcolor{positive}{+2.5} & \textcolor{neutral}{+0.0} & \textcolor{positive}{+35.8} & \textcolor{positive}{+4.0} & \textcolor{positive}{+28.0} & \textcolor{neutral}{+0.0} & \textcolor{positive}{+5.5} & \textcolor{positive}{+2.6} & \textcolor{positive}{+25.5} & \textcolor{positive}{+62.1} & \textcolor{positive}{+0.3} & \textcolor{positive}{+50.6} & \textcolor{positive}{+23.1} & \textcolor{positive}{+16.0} \\
\bottomrule
\end{tabular}
\end{adjustbox}
\label{tab:pass_1_details}
\end{table*}

\subsection{Efficient Incremental Computation of Higher-Order Metrics}
\label{sec:efficient-computation}

A naive computation of the temporal difference metrics would be computationally prohibitive. Our method's feasibility hinges on an efficient, incremental algorithm that computes the required metrics without redundant operations on the growing hidden state matrix $\mathbf{Z} \in \mathbb{R}^{T\times D}$.

The effective rank is derived from the singular values of the mean-centered hidden state matrix $\mathbf{Z}_c$. These are equivalent to the square roots of the eigenvalues of the centered Gram matrix $\mathbf{K}=\mathbf{Z}_c \mathbf{Z}_c^\top$. Instead of recomputing $\mathbf{K}_t$ from scratch at each time step t, our algorithm incrementally updates two sufficient statistics: the uncentered Gram matrix $\mathbf{U}_t=\mathbf{Z}_{1:t}\mathbf{Z}_{1:t}^\top$ and the sum of hidden state vectors $s_t=\sum_{i=1}^tz_i$. When extending the analysis window, the new uncentered Gram matrix $\mathbf{U}_t$ is constructed from the prior matrix $\mathbf{U}_{t-s}$ and the new chunk of hidden states $\Delta \mathbf{Z}_t=\mathbf{Z}_{t-s+1:t}$. This update follows a recursive block matrix structure:

\begin{equation}
   \mathbf{U}_t = \begin{pmatrix} \mathbf{U}_{t-s} & \mathbf{Z}_{1:t-s}(\Delta \mathbf{Z}_t)^{\top} \\ (\Delta \mathbf{Z}_t)\mathbf{Z}_{1:t-s}^{\top} & (\Delta \mathbf{Z}_t)(\Delta \mathbf{Z}_t)^{\top} \end{pmatrix} 
\end{equation}

From the efficiently updated $\mathbf{U}_t$ and $s_t$, we can directly construct the centered Gram matrix $\mathbf{K}_t$. Letting $\boldsymbol{\mu}_t = s_t/t$ be the mean vector and $\mathbf{1}_t$ be a column vector of ones, $\mathbf{K}_t$ can be expressed as:

\begin{equation}
\begin{aligned}
\mathbf{K}_t
&= \mathbf{U}_t - (\mathbf{Z}_{1:t}\boldsymbol{\mu}_t)\mathbf{1}_t^\top - \mathbf{1}_t(\mathbf{Z}_{1:t}\boldsymbol{\mu}_t)^\top \\
&\quad + (\boldsymbol{\mu}_t^\top\boldsymbol{\mu}_t)\cdot(\mathbf{1}_t\mathbf{1}_t^\top).
\end{aligned}
\end{equation}

This allows for the calculation of $\mathbf{K}_t$ without re-accessing the full history of hidden states. The eigenvalues $\{\lambda_j\}$ of $\mathbf{K}_t$ are then used to derive the effective rank. First, the singular values of the centered matrix are obtained, $\sigma_j = \sqrt{\lambda_j}$. These are normalized to form a probability distribution, $p_j = \sigma_j / \sum_k \sigma_k$. The effective rank is then the exponential of the Shannon entropy of this distribution:$\text{erank}(\mathbf{Z}_{c,t}) = \exp\left(-\sum_j p_j \log p_j\right)$. This pipeline efficiently yields a sequence of effective rank values, $m_{j\cdot s} = \text{erank}(\mathbf{Z}_{c, j\cdot s})$, at each stride $s$. From this sequence, we compute the instantaneous first-order difference $\delta$, which compares the current value to the running average of all preceding values. This is defined recursively as:$\delta_{j \cdot s} = m_{j \cdot s} - \frac{1}{j-1}\sum_{k=1}^{j-1} m_{k \cdot s}$.

The computational advantage of this incremental approach is substantial. While the total cost for the series of eigenvalue decompositions $\mathcal{O}(T^4/s)$, is common to both methods, the cost of matrix construction differs significantly. The naive method's recalculation totals $\mathcal{O}(DT^3/s)$, whereas our incremental update method reduces this to $O(DT^2)$. This reduction of the polynomial dependency on sequence length T from cubic to quadratic is critical, as this term is multiplied by the large hidden dimension $D$, making it the dominant factor in practical performance and rendering the dense calculation of temporal dynamics feasible. In the \emph{worst-case} scenario where the sequence length $T$ exceeds the hidden dimension $D$, and both $D$ and the stride $s$ can be treated as constants. The naïve approach that reconstructs matrices independently at each stride has a matrix-construction cost scaling as $\mathcal{O}(T^2)$, VERL’s incremental Gram/covariance updates scale as $\mathcal{O}(T)$. So asymptotically, our implementation has a strictly better dependency on $T$ than a naïve SVD-based design.

\subsection{Time Overhead of \ourmethod{} Training}

We conducted post-training with Zero RL on several base models. The Fig. \ref{time_overhead} illustrates the time associated with each computational stage. The 'metrics calculation' component, which represents the cost of computing metrics for hidden states, accounts for an insignificant portion of the total processing time. This demonstrates that our method does not introduce substantial time overhead. To further stress-test the \emph{worst-case} scenario, we deliberately compute ER, ERV, and ERA on the CPU rather than the GPU, and still observe that the additional time overhead remains negligible.

    
    
    
    

\section{More Experiments}
\label{sec:More Experiments}
\subsection{Detailed Analysis of Response-Level Hidden States}
\label{Appendix_Response_Analysis}
As shown in Figs.~\ref{Appendix_observation1_grpo} and \ref{Appendix_observation1_ppo}, our analysis of response-level hidden states across additional LLMs confirms that the insights presented in Sec.~\ref{Main_Response_Analysis} hold true for various base models and RL paradigms.

    
    
    
    

\subsection{Detailed Analysis of Dataset-Level Hidden States}
\label{Appendix_Dataset_Analysis}
As shown in Figs.~\ref{Appendix_observation2_grpo} and \ref{Appendix_observation2_ppo}, our analysis of dataset-level hidden states across additional LLMs confirms that the insights presented in Sec.~\ref{Main_Dataset_Analysis} hold for various base models and RL paradigms.

\subsection{Detailed Analysis of Pass@1 Performance}
As shown in Tab.~\ref{tab:pass_1_details}, Pass@1 measures the model's ability to generate a correct answer in a single attempt, which directly reflects its exploitation ability. We fine-tune the base model by integrating our \ourmethod{}-based Advantage method into two reinforcement learning paradigms, GRPO and PPO.

\label{Appendix_Pass@1}


\subsection{Detailed Analysis of Pass@$k$ Performance}
Tab.~\ref{tab:pass_k_details} provides a comprehensive analysis of model performance under Pass@k, a standard measure of success under larger decoding budgets. As a supplement to the main paper, it reports VERL's performance across a variety of mathematical reasoning benchmarks. These results show that VERL improves exploration under larger sampling budgets, especially on harder benchmarks.
\label{Appendix_Pass@k}

\begin{table*}[htbp]
\centering
\caption{Performance comparison of instruction-tuned models under diverse decoding settings (Pass@$k$). All results are reported in percentage (\%).}
\begin{adjustbox}{width=\textwidth,center}
\begin{tabular}{lcccccc}
\toprule
\textbf{Model} & 
\makecell[c]{\textbf{MATH500} \\ \textbf{(Pass@16)}} & 
\makecell[c]{\textbf{AMC23} \\ \textbf{(Pass@128)}} & 
\makecell[c]{\textbf{AMC24} \\ \textbf{(Pass@128)}} & 
\makecell[c]{\textbf{AIME24} \\ \textbf{(Pass@256)}} & 
\makecell[c]{\textbf{AIME25} \\ \textbf{(Pass@256)}} & 
\makecell[c]{\textbf{Avg.}} \\
\midrule

\textbf{Llama-3.2-3B-Instruct} 
& 79.8 & 93.5 & 51.5 & 40.0 & 30.0 & 58.96 \\
+ GRPO 
& 80.2 & 95.4 & 60.6 & 40.0 & 30.0 & 61.24 \\
\rowcolor{blue!10} + GRPO w/ \ourmethod{}. 
& 80.6 & 95.7 & 59.0 & 50.0 & 36.7 & 64.40 \\
$\Delta_{\text{GRPO}}$ 
& \pos{0.4} & \pos{0.3} & \negv{1.6} & \pos{10.0} & \pos{6.7} & \pos{3.16} \\
+ PPO 
& 82.2 & 94.5 & 57.0 & 46.7 & 36.7 & 63.42 \\
\rowcolor{blue!10} + PPO w/ \ourmethod{}. 
& 82.4 & 94.7 & 57.8 & 46.7 & 40.0 & 64.32 \\
$\Delta_{\text{PPO}}$ 
& \pos{0.2} & \pos{0.2} & \pos{0.8} & \zero & \pos{3.3} & \pos{0.90} \\
\midrule

\textbf{Llama-3.1-8B-Instruct} 
& 79.8 & 94.6 & 57.4 & 46.7 & 36.7 & 63.04 \\
+ GRPO 
& 83.4 & 94.9 & 56.9 & 53.3 & 36.7 & 65.04 \\
\rowcolor{blue!10} + GRPO w/ \ourmethod{}. 
& 83.4 & 95.1 & 63.1 & 50.0 & 36.7 & 65.66 \\
$\Delta_{\text{GRPO}}$ 
& \zero & \pos{0.2} & \pos{6.2} & \negv{3.3} & \zero & \pos{0.62} \\
+ PPO 
& 79.2 & 92.4 & 59.0 & 46.7 & 36.7 & 62.80 \\
\rowcolor{blue!10} + PPO w/ \ourmethod{}. 
& 82.4 & 91.9 & 60.0 & 53.3 & 36.7 & 64.86 \\
$\Delta_{\text{PPO}}$ 
& \pos{3.2} & \negv{0.5} & \pos{1.0} & \pos{6.6} & \zero & \pos{2.06} \\
\midrule

\textbf{Qwen2.5-3B} 
& 86.0 & 96.7 & 69.0 & 56.7 & 40.0 & 69.68 \\
+ GRPO 
& 86.6 & 92.2 & 68.5 & 46.7 & 40.0 & 66.80 \\
\rowcolor{blue!10} + GRPO w/ \ourmethod{}. 
& 87.6 & 95.9 & 67.8 & 53.3 & 43.3 & 69.58 \\
$\Delta_{\text{GRPO}}$ 
& \pos{1.0} & \pos{3.7} & \negv{0.7} & \pos{6.6} & \pos{3.3} & \pos{2.78} \\
+ PPO 
& 87.8 & 96.5 & 67.9 & 43.3 & 43.3 & 67.76 \\
\rowcolor{blue!10} + PPO w/ \ourmethod{}. 
& 88.2 & 96.8 & 67.3 & 53.3 & 43.3 & 69.78 \\
$\Delta_{\text{PPO}}$ 
& \pos{0.4} & \pos{0.3} & \negv{0.6} & \pos{10.0} & \zero & \pos{2.02} \\
\midrule

\textbf{Qwen2.5-7B} 
& 90.6 & 98.4 & 73.7 & 60.0 & 60.0 & 76.54 \\
+ GRPO 
& 90.8 & 97.8 & 78.3 & 56.7 & 50.0 & 74.72 \\
\rowcolor{blue!10} + GRPO w/ \ourmethod{}. 
& 91.4 & 98.3 & 79.0 & 63.3 & 60.0 & 78.40 \\
$\Delta_{\text{GRPO}}$ 
& \pos{0.6} & \pos{0.5} & \pos{0.7} & \pos{6.6} & \pos{10.0} & \pos{3.68} \\
+ PPO 
& 91.2 & 98.6 & 74.3 & 53.3 & 56.7 & 74.82 \\
\rowcolor{blue!10} + PPO w/ \ourmethod{}. 
& 91.4 & 98.0 & 74.4 & 56.7 & 66.7 & 77.44 \\
$\Delta_{\text{PPO}}$ 
& \pos{0.2} & \negv{0.6} & \pos{0.1} & \pos{3.4} & \pos{10.0} & \pos{2.62} \\
\midrule

\textbf{Mathstral-7B-v0.1} 
& 80.4 & 88.5 & 60.9 & 43.3 & 36.7 & 61.96 \\
+ GRPO 
& 84.8 & 87.3 & 69.2 & 36.7 & 40.0 & 63.60 \\
\rowcolor{blue!10} + GRPO w/ \ourmethod{}. 
& 87.0 & 97.0 & 76.9 & 50.0 & 50.0 & 72.18 \\
$\Delta_{\text{GRPO}}$ 
& \pos{2.2} & \pos{9.7} & \pos{7.7} & \pos{13.3} & \pos{10.0} & \pos{8.58} \\
+ PPO 
& 82.4 & 91.7 & 70.7 & 53.3 & 40.0 & 67.62 \\
\rowcolor{blue!10} + PPO w/ \ourmethod{}. 
& 84.8 & 93.8 & 69.9 & 53.3 & 46.7 & 69.70 \\
$\Delta_{\text{PPO}}$ 
& \pos{2.4} & \pos{2.1} & \negv{0.8} & \zero & \pos{6.7} & \pos{2.08} \\
\midrule

\textbf{Mistral-7B-v0.3} 
& 36.0 & 73.5 & 39.6 & 20.0 & 16.7 & 37.16 \\
+ GRPO 
& 33.0 & 63.2 & 36.0 & 10.0 & 10.0 & 30.44 \\
\rowcolor{blue!10} + GRPO w/ \ourmethod{}. 
& 34.4 & 64.5 & 38.0 & 16.7 & 13.3 & 33.38 \\
$\Delta_{\text{GRPO}}$ 
& \pos{1.4} & \pos{1.3} & \pos{2.0} & \pos{6.7} & \pos{3.3} & \pos{2.94} \\
+ PPO 
& 21.8 & 46.4 & 25.1 & 6.7 & 6.7 & 21.34 \\
\rowcolor{blue!10} + PPO w/ \ourmethod{}. 
& 19.2 & 46.5 & 30.1 & 3.3 & 13.3 & 22.48 \\
$\Delta_{\text{PPO}}$ 
& \negv{2.6} & \pos{0.1} & \pos{5.0} & \negv{3.4} & \pos{6.6} & \pos{1.14} \\
\bottomrule
\end{tabular}
\end{adjustbox}
\label{tab:pass_k_details}
\end{table*}

\begin{table*}[htbp]
    \centering
    \scriptsize
    \caption{Comparison of GRPO + VERL using an intermediate layer (layer 14) versus the final layer, evaluated by Pass@1 on multiple math benchmarks. Using the last layer yields the strongest average improvement.}
    \label{tab:verl_layer_pass1}
    \resizebox{\textwidth}{!}{
    \begin{tabular}{lcccccccccccccccc}
        \toprule
        Method & aime24 & aime25 & amc23 & amc24 & asdiv & carp\_en & cmath & gaokao24\_I & gaokao24\_mix & gaokao\_math\_cloze & gsm8k & mawps & olympiadbench & svamp & tabmwp & Avg. \\
        \midrule
        Llama-3.2-3B-Instruct
        & 0.0 & 0.0 & 25.0 & 11.1 & 74.6 & 26.5 & 10.2 & 14.3 & 14.3 & 6.8 & 66.6 & 86.9 & 12.7 & 74.1 & 41.4 & 30.97 \\
        GRPO
        & 3.3 & 0.0 & 27.5 & 8.9 & 88.8 & 45.0 & 28.3 & 21.4 & 20.9 & 23.7 & 80.7 & 96.0 & 16.7 & 87.7 & 71.7 & 41.37 \\
        GRPO w/ VERL (layer = 14)
        & 10.0 & 0.0 & 27.5 & 11.1 & 88.6 & 43.6 & 30.7 & 21.4 & 16.5 & 22.0 & 81.9 & 95.5 & 18.1 & 87.0 & 71.4 & 41.69 \\
        GRPO w/ VERL (layer = last)
        & 13.3 & 6.7 & 25.0 & 11.1 & 89.3 & 45.4 & 46.2 & 14.3 & 22.0 & 22.9 & 81.7 & 96.0 & 17.6 & 87.8 & 72.3 & \textbf{43.44} \\
        \bottomrule
    \end{tabular}
    }
\end{table*}

\begin{table*}[htbp]
    \centering
    \scriptsize
    \setlength{\tabcolsep}{4pt}
    \caption{Comparison of GRPO + VERL using an intermediate layer (layer 14) versus the final layer, evaluated by Pass@k on several math benchmarks. Again, using the last layer yields the best average improvement.}
    \label{tab:verl_layer_passk}
    \begin{tabular}{lcccccc}
        \toprule
        Method & math500@16 & amc23@128 & amc24@128 & aime24@256 & aime25@256 & Avg. \\
        \midrule
        Llama-3.2-3B-Instruct
        & 79.8 & 93.5 & 51.5 & 40.0 & 30.0 & 58.96 \\
        GRPO
        & 80.2 & 95.4 & 60.6 & 40.0 & 30.0 & 61.24 \\
        GRPO w/ VERL (layer = 14)
        & 81.0 & 92.9 & 57.0 & 40.0 & 36.7 & 61.52 \\
        GRPO w/ VERL (layer = last)
        & 80.6 & 95.7 & 59.0 & 50.0 & 36.7 & \textbf{64.40} \\
        \bottomrule
    \end{tabular}
\end{table*}

\subsection{Robustness Across Random Seeds}
\label{app:seed_robustness}

To assess robustness to random seed variation, we ran \textbf{five independent training seeds} for \textsc{Llama-3.2-3B-Instruct} on all main benchmarks. We report mean $\pm$ standard deviation over seeds in Tab.~\ref{tab:seed_pass1} (Pass@1) and Tab.~\ref{tab:seed_passk} (Pass@k). Compared to \textsc{GRPO}, adding \ourmethod{} improves the \textbf{average Pass@1} from $39.24 \pm 0.72$ to $45.18 \pm 0.38$ (Tab.~\ref{tab:seed_pass1}), and the \textbf{average Pass@k} from $59.87 \pm 1.37$ to $62.78 \pm 0.63$ (Tab.~\ref{tab:seed_passk}). We further conduct two-sided \textbf{Welch's t-tests} at significance level $\alpha=0.05$ for each benchmark; statistically significant gains are marked with \texttt{*}. Overall, most improvements remain significant, suggesting the gains are not due to random seed noise.

\begin{table*}[!t]
\centering
\setlength{\tabcolsep}{2.4pt}
\renewcommand{\arraystretch}{1.05}
\footnotesize
\caption{Seed robustness on main benchmarks (Pass@1). Results are mean $\pm$ std over 5 seeds. \texttt{*} indicates Welch's t-test significance at $\alpha=0.05$ comparing \textsc{GRPO+}\ourmethod{} against \textsc{GRPO}.}
\label{tab:seed_pass1}
\resizebox{\textwidth}{!}{%
\begin{tabular}{lcccccccccccccccc}
\toprule
Method
& aime24 & aime25 & amc23 & amc24 & asdiv & carp\_en & cmath
& gaokao24\_I & gaokao24\_mix & gaokao\_cloze
& gsm8k & mawps & olympiad & svamp & tabmwp & \textbf{Avg.} \\
\midrule
\textsc{GRPO}
& $10.00 \pm 0.00$ & $0.00 \pm 0.00$ & $17.50 \pm 0.00$ & $11.54 \pm 0.98$
& $88.62 \pm 0.54$ & $44.82 \pm 0.04$ & $15.34 \pm 0.36$
& $5.72 \pm 7.83$ & $19.58 \pm 2.95$ & $22.20 \pm 1.82$
& $80.88 \pm 0.60$ & $95.60 \pm 0.00$ & $17.80 \pm 0.85$
& $87.08 \pm 0.22$ & $71.92 \pm 0.41$
& $\mathbf{39.24 \pm 0.72}$ \\
\textsc{GRPO+}\ourmethod{}
& $12.64 \pm 1.48$\texttt{*} & $3.98 \pm 1.52$\texttt{*} & $25.50 \pm 1.12$\texttt{*} & $16.00 \pm 2.46$\texttt{*}
& $89.50 \pm 0.14$\texttt{*} & $45.20 \pm 0.14$\texttt{*} & $56.62 \pm 0.93$\texttt{*}
& $25.72 \pm 3.94$\texttt{*} & $18.92 \pm 1.97$ & $24.04 \pm 0.76$
& $83.00 \pm 0.59$\texttt{*} & $96.32 \pm 0.16$\texttt{*} & $18.30 \pm 0.71$
& $87.30 \pm 0.28$ & $74.64 \pm 0.75$\texttt{*}
& $\mathbf{45.18 \pm 0.38}$\texttt{*} \\
\bottomrule
\end{tabular}%
}
\end{table*}

\begin{table*}[!t]
\centering
\setlength{\tabcolsep}{4.5pt}
\renewcommand{\arraystretch}{1.08}
\footnotesize
\caption{Seed robustness on harder Pass@k settings. Results are mean $\pm$ std over 5 seeds. \texttt{*} indicates Welch's t-test significance at $\alpha=0.05$.}
\label{tab:seed_passk}
\begin{tabular}{lcccccc}
\toprule
Method & math500@16 & amc23@128 & amc24@128 & aime24@256 & aime25@256 & \textbf{Avg.} \\
\midrule
\textsc{GRPO}
& $80.40 \pm 0.00$ & $90.78 \pm 0.18$ & $56.18 \pm 1.50$ & $39.34 \pm 1.48$ & $32.66 \pm 5.95$
& $\mathbf{59.87 \pm 1.37}$ \\
\textsc{GRPO+}\ourmethod{}
& $80.52 \pm 0.18$ & $94.22 \pm 1.01$\texttt{*} & $58.46 \pm 1.69$ & $43.98 \pm 3.66$\texttt{*} & $36.70 \pm 0.00$
& $\mathbf{62.78 \pm 0.63}$\texttt{*} \\
\bottomrule
\end{tabular}
\end{table*}

Most benchmarks show statistically significant gains (marked with \texttt{*}). On the few tasks without \texttt{*}, the mean still improves, but variance across seeds is slightly larger. On these Pass@k evaluations, \textsc{GRPO+}\ourmethod{} again outperforms \textsc{GRPO}, with a statistically significant improvement in the average Pass@k (see \textbf{Avg.} column).

\subsection{Detailed Analysis of OOD Performance}
\label{app:ood}

\begin{table*}[t!]
\centering
\caption{Out-of-domain (OOD) reasoning performance (Pass@1, \%) across diverse domains. Results are reported for Mathstral-7B-v0.1 and Qwen2.5-7B, comparing vanilla GRPO/PPO and their \ourmethod{}-enhanced variants.}
\setlength{\tabcolsep}{4pt}
\renewcommand{\arraystretch}{0.8}
\begin{adjustbox}{width=\textwidth,center}
\footnotesize
\begin{tabular}{lccc ccccc ccc cccc cccc ccc}
\toprule
\textbf{Model / Method} &
\makecell[c]{\textbf{AstroQA}\\\textbf{EN}} &
\makecell[c]{\textbf{AstroQA}\\\textbf{ZH}} &
\makecell[c]{ } &
\makecell[c]{\textbf{GPQA}\\\textbf{D}} &
\makecell[c]{\textbf{GPQA}\\\textbf{Ext}} &
\makecell[c]{\textbf{GPQA}\\\textbf{Main}} &
\makecell[c]{\textbf{SciBench}} &
\makecell[c]{ } &
\makecell[c]{\textbf{LEXam}} &
\makecell[c]{\textbf{LexEval}} &
\makecell[c]{ } &
\makecell[c]{\textbf{LogiQA}\\\textbf{EN}} &
\makecell[c]{\textbf{LogiQA}\\\textbf{ZH}} &
\makecell[c]{\textbf{ReClor}} &
\makecell[c]{ } &
\makecell[c]{\textbf{MedMCQA}} &
\makecell[c]{\textbf{MedQA}} &
\makecell[c]{\textbf{PubMedQA}} &
\makecell[c]{ } &
\makecell[c]{\textbf{MMLU}} &
\makecell[c]{\textbf{MMLU-Pro}} &
\makecell[c]{\textbf{Avg.}} \\
\midrule

\textbf{Mathstral-7B-v0.1}
& 50.7 & 43.8 && 29.3 & 26.0 & 25.4 & 15.8 && 19.6 & 38.7 && 45.8 & 44.6 & 43.6 && 42.9 & 39.7 & 22.4 && 53.9 & 26.4 & 35.25 \\
+ GRPO
& 57.1 & 49.3 && 28.8 & 29.3 & 29.9 & 22.4 && 22.4 & 42.8 && 47.1 & 45.7 & 46.3 && 44.9 & 43.8 & 22.6 && 56.7 & 29.8 & 39.31 \\
\rowcolor{blue!10} + GRPO w/ \ourmethod{}.
& 58.0 & 51.6 && 31.8 & 31.1 & 34.6 & 27.9 && 22.6 & 47.6 && 50.4 & 44.8 & 49.7 && 47.1 & 46.1 & 24.4 && 58.8 & 30.7 & 41.68 \\
$\Delta_{\text{GRPO}}$
& \pos{0.9} & \pos{2.3} && \pos{3.0} & \pos{1.8} & \pos{4.7} & \pos{5.5} && \pos{0.2} & \pos{4.8} && \pos{3.3} & \negv{0.9} & \pos{3.4} && \pos{2.2} & \pos{2.3} & \pos{1.8} && \pos{2.1} & \pos{0.9} & \pos{2.37} \\
+ PPO
& 52.7 & 47.5 && 28.8 & 29.9 & 27.9 & 22.1 && 17.3 & 44.5 && 47.5 & 45.0 & 46.5 && 44.2 & 45.2 & 23.1 && 56.5 & 29.1 & 37.93 \\
\rowcolor{blue!10} + PPO w/ \ourmethod{}.
& 54.5 & 49.1 && 31.3 & 26.6 & 26.1 & 23.0 && 21.4 & 42.2 && 47.5 & 51.6 & 46.2 && 44.8 & 45.1 & 22.3 && 56.9 & 29.3 & 39.34 \\
$\Delta_{\text{PPO}}$
& \pos{1.8} & \pos{1.6} && \pos{2.5} & \negv{3.3} & \negv{1.8} & \pos{0.9} && \pos{4.1} & \negv{2.3} && \zero & \pos{6.6} & \negv{0.3} && \pos{0.6} & \negv{0.1} & \negv{0.8} && \pos{0.4} & \pos{0.2} & \pos{1.42} \\

\midrule

\textbf{Qwen2.5-7B}
& 61.2 & 58.0 && 27.3 & 26.6 & 24.6 & 15.8 && 21.1 & 50.4 && 50.8 & 49.7 & 56.0 && 50.6 & 62.0 & 43.6 && 61.3 & 36.1 & 43.44 \\
+ GRPO
& 59.6 & 61.3 && 27.3 & 27.8 & 28.8 & 21.5 && 22.0 & 53.6 && 52.1 & 52.1 & 58.8 && 52.1 & 69.2 & 46.6 && 60.6 & 33.8 & 46.15 \\
\rowcolor{blue!10} + GRPO w/ \ourmethod{}.
& 59.4 & 62.4 && 30.8 & 28.9 & 26.8 & 22.8 && 23.0 & 54.6 && 52.8 & 52.4 & 61.6 && 52.0 & 70.3 & 47.8 && 60.9 & 33.4 & 47.09 \\
$\Delta_{\text{GRPO}}$
& \negv{0.2} & \pos{1.1} && \pos{3.5} & \pos{1.1} & \negv{2.0} & \pos{1.3} && \pos{1.0} & \pos{1.0} && \pos{0.7} & \pos{0.3} & \pos{2.8} && \negv{0.1} & \pos{1.1} & \pos{1.2} && \pos{0.3} & \negv{0.4} & \pos{0.94} \\
+ PPO
& 60.0 & 62.1 && 27.3 & 25.3 & 28.3 & 22.1 && 22.8 & 55.2 && 52.1 & 52.1 & 58.6 && 53.6 & 70.4 & 45.8 && 60.9 & 34.1 & 46.20 \\
\rowcolor{blue!10} + PPO w/ \ourmethod{}.
& 60.9 & 61.3 && 34.8 & 30.0 & 27.0 & 23.0 && 23.4 & 57.1 && 53.0 & 52.3 & 61.8 && 52.9 & 71.0 & 45.4 && 61.0 & 34.4 & 47.45 \\
$\Delta_{\text{PPO}}$
& \pos{0.9} & \negv{0.8} && \pos{7.5} & \pos{4.7} & \negv{1.3} & \pos{0.9} && \pos{0.6} & \pos{1.9} && \pos{0.9} & \pos{0.2} & \pos{3.2} && \negv{0.7} & \pos{0.6} & \negv{0.4} && \pos{0.1} & \pos{0.3} & \pos{1.25} \\

\bottomrule
\end{tabular}
\end{adjustbox}
\label{tab:ood_pass1}
\end{table*}

A key motivation of this study is to assess whether our proposed \textsc{VERL} algorithm improves
\emph{out-of-domain (OOD)} generalization for reasoning-centric post-training, rather than merely
overfitting to the in-domain training distribution.
This is aligned with our overall claim that \textsc{VERL} enhances robustness and OOD performance
beyond standard RL fine-tuning baselines. 

To stress-test OOD generalization, we evaluate on a diverse suite of reasoning benchmarks spanning multiple domains:
(i) \textbf{Astronomy}: Astro-QA (English/Chinese)~\citep{li2025astronomical};
(ii) \textbf{Science}: GPQA (Diamond/Extended/Main)~\citep{rein2024gpqa} and SciBench~\citep{wang2024scibench};
(iii) \textbf{Law}: LEXam~\citep{fan2025lexam} and LexEval~\citep{li2024lexeval};
(iv) \textbf{Logical reasoning}: LogiQA (English/Chinese)~\citep{ijcai2020p501} and ReClor~\citep{yu2020reclor};
(v) \textbf{Medical QA}: MedMCQA~\citep{PalUS22MedMCQA}, MedQA~\citep{MedQA}, and PubMedQA~\citep{JinDLCL19PubMedQA};
(vi) \textbf{Broad knowledge}: MMLU~\citep{HendrycksBBZMSS21MMLU} and MMLU-Pro~\citep{WangMZNCGRAHJLK24MMLU-PRO}.
All results are reported as Pass@1 (\%).

Tab.~\ref{tab:ood_pass1} summarizes the OOD Pass@1 results for both Mathstral-7B and Qwen2.5-7B.
Overall, \textsc{VERL} yields consistent average improvements when plugged into either GRPO or PPO:
for Mathstral-7B, \textsc{VERL} improves the OOD average from 39.31 to 41.68 (+2.37) on GRPO
and from 37.93 to 39.34 (+1.42) on PPO;
for Qwen2.5-7B, it improves the OOD average from 46.15 to 47.09 (+0.94) on GRPO
and from 46.20 to 47.45 (+1.25) on PPO.
These gains are broadly distributed across heterogeneous domains, supporting that \textsc{VERL}
enhances OOD reasoning robustness rather than trading off generalization for in-domain gains.

\subsection{Comparison with Regularization Method in Action Space}
\label{app:compare_action_reg}


\begin{table*}[!t]
\centering
\caption{Comparison with action-space regularization baselines on selected reasoning datasets (Pass@1, \%). Best value in each column is in \textbf{bold}; second best is \underline{underlined}.}
\setlength{\tabcolsep}{4pt}
\renewcommand{\arraystretch}{0.8}
\begin{adjustbox}{width=\textwidth,center}
\footnotesize
\begin{tabular}{lcccccccccccccccc}
\toprule
\textbf{Model / Method} &
\makecell[c]{\textbf{AIME24}} &
\makecell[c]{\textbf{AIME25}} &
\makecell[c]{\textbf{AMC23}} &
\makecell[c]{\textbf{AMC24}} &
\makecell[c]{\textbf{ASDiv}} &
\makecell[c]{\textbf{CARP\_EN}} &
\makecell[c]{\textbf{CMath}} &
\makecell[c]{\textbf{Gaokao24\_I}} &
\makecell[c]{\textbf{Gaokao24\_mix}} &
\makecell[c]{\textbf{Gaokao\_math}\\\textbf{\_cloze}} &
\makecell[c]{\textbf{GSM8K}} &
\makecell[c]{\textbf{MAWPS}} &
\makecell[c]{\textbf{OlympiadBench}} &
\makecell[c]{\textbf{SVAMP}} &
\makecell[c]{\textbf{TabMWP}} &
\makecell[c]{\textbf{Avg.}} \\
\midrule
\textbf{Llama-3.2-3B-Instruct}
& 0.0 & 0.0 & 25.0 & 11.1 & 74.6 & 26.5 & 10.2 & \underline{14.3} & 14.3 & 6.8 & 66.6 & 86.9 & 12.7 & 74.1 & 41.4 & 30.97 \\
+ GRPO
& 3.3 & 0.0 & \underline{27.5} & 8.9 & 88.8 & 45.0 & 28.3 & \textbf{21.4} & \underline{20.9} & \underline{23.7} & 80.7 & \underline{96.0} & 16.7 & \underline{87.7} & 71.7 & 41.37 \\
+ GRPO w/ \textsc{Action-LogP} regu
& 3.3 & 0.0 & \underline{27.5} & \textbf{15.6} & 88.8 & 44.6 & \underline{38.2} & 7.1 & 15.4 & 22.0 & 80.9 & \textbf{96.1} & 16.3 & 87.4 & \underline{71.9} & 41.01 \\
+ GRPO w/ \textsc{Action-Entropy} regu
& \underline{10.0} & 0.0 & \textbf{35.0} & 8.9 & \underline{89.2} & \underline{45.3} & 37.5 & 7.1 & 16.5 & \textbf{28.0} & \underline{81.9} & \textbf{96.1} & \textbf{17.9} & 87.0 & 71.6 & \underline{42.13} \\
+ GRPO w/ \textsc{Action-LogP{+}Entropy}
& 3.3 & \underline{3.3} & 25.0 & \underline{13.3} & 88.5 & 44.7 & 37.7 & 0.0 & 18.7 & \underline{23.7} & \textbf{82.0} & \textbf{96.1} & 15.6 & 87.0 & 70.5 & 40.63 \\
\rowcolor{blue!10} + GRPO w/ \ourmethod{} (ours)
& \textbf{13.3} & \textbf{6.7} & 25.0 & 11.1 & \textbf{89.3} & \textbf{45.4} & \textbf{46.2} & \underline{14.3} & \textbf{22.0} & 22.9 & 81.7 & \underline{96.0} & \underline{17.6} & \textbf{87.8} & \textbf{72.3} & \textbf{43.40} \\
\bottomrule
\end{tabular}
\end{adjustbox}
\label{tab:action_reg_pass1}
\end{table*}


\begin{table*}[!t]
\centering
\caption{Comparison with action-space regularization baselines under larger decoding budgets (Pass@k, \%). Best value in each column is in \textbf{bold}; second best is \underline{underlined}.}
\setlength{\tabcolsep}{4pt}
\renewcommand{\arraystretch}{0.8}
\begin{adjustbox}{width=\textwidth,center}
\footnotesize
\begin{tabular}{lcccccc}
\toprule
\textbf{Model / Method} &
\makecell[c]{\textbf{Math500}\\\textbf{(Pass@16)}} &
\makecell[c]{\textbf{AMC23}\\\textbf{(Pass@128)}} &
\makecell[c]{\textbf{AMC24}\\\textbf{(Pass@128)}} &
\makecell[c]{\textbf{AIME24}\\\textbf{(Pass@256)}} &
\makecell[c]{\textbf{AIME25}\\\textbf{(Pass@256)}} &
\makecell[c]{\textbf{Avg.}} \\
\midrule
\textbf{Llama-3.2-3B-Instruct}
& 79.8 & 93.5 & 51.5 & 40.0 & \underline{30.0} & 58.96 \\
+ GRPO
& \underline{80.2} & \underline{95.4} & \textbf{60.6} & 40.0 & \underline{30.0} & 61.24 \\
+ GRPO w/ \textsc{Action-LogP} regu
& 79.6 & 93.8 & 56.9 & 40.0 & \underline{30.0} & 60.06 \\
+ GRPO w/ \textsc{Action-Entropy} regu
& 78.4 & 93.1 & 56.9 & \underline{46.7} & \textbf{36.7} & \underline{62.36} \\
+ GRPO w/ \textsc{Action-LogP{+}Entropy}
& 80.0 & 93.9 & 55.8 & 40.0 & \textbf{36.7} & 61.28 \\
\rowcolor{blue!10} + GRPO w/ \ourmethod{} (ours)
& \textbf{80.6} & \textbf{95.7} & \underline{59.0} & \textbf{50.0} & \textbf{36.7} & \textbf{64.40} \\
\bottomrule
\end{tabular}
\end{adjustbox}
\label{tab:action_reg_passk}
\end{table*}

This experiment aims to compare \ourmethod{} with prior \emph{action-space} regularization strategies used in RL post-training.
Here, ``action space'' refers to the token-level policy over next-token actions during generation.
A common practice is to regularize training using token-level proxy signals computed directly from this policy (e.g., confidence or entropy).
Our goal is to test whether such action-space proxy regularizers can match the performance gains of \ourmethod{},
and to provide evidence that representation-level metrics yield a more effective regularization signal for reasoning.

\paragraph{Setup.}
We keep the post-training pipeline unchanged (same base model, RL algorithm, training data, optimizer, and decoding configuration),
and only replace the auxiliary regularization signal.
We consider three action-space baselines:
(i) \textsc{Action-LogP}, which regularizes using only the \textbf{average log-probability} of the generated response (a confidence-style signal);
(ii) \textsc{Action-Entropy}, which regularizes using only the \textbf{response entropy} (a stochasticity-style signal);
(iii) \textsc{Action-LogP{+}Entropy}, which uses the \textbf{sum} of the above two signals.
In contrast, \ourmethod{} (\textsc{VERL}) employs hidden-state trajectory metrics defined in a semantically richer representation space.
We report Pass@1 results in Tab.~\ref{tab:action_reg_pass1} and larger-budget decoding results (Pass@k) in Tab.~\ref{tab:action_reg_passk}.

\paragraph{Findings.}
Two observations stand out.
First, action-space proxy regularization is often \emph{unstable} under combination:
while a single proxy can sometimes help (e.g., \textsc{Action-Entropy} achieves the strongest or second-strongest performance on several columns),
the naive combination \textsc{Action-LogP{+}Entropy} does not reliably improve further and can even underperform the better single-signal choice
(see the Avg. column in Tab.~\ref{tab:action_reg_pass1} and the Pass@k columns in Tab.~\ref{tab:action_reg_passk}).
This indicates that token-level confidence (LogP) and stochasticity (entropy) may interfere when jointly used as a regularizer.

Second, across both evaluation regimes, \ourmethod{} consistently provides the best overall results.
On Pass@1, \ourmethod{} achieves the highest Avg.\ and attains the best (or second-best) scores on most datasets in Tab.~\ref{tab:action_reg_pass1};
on Pass@k, \ourmethod{} similarly yields the strongest Avg.\ and the top performance on key high-budget reasoning benchmarks
(e.g., AIME24 at Pass@256 and AMC23 at Pass@128 in Tab.~\ref{tab:action_reg_passk}).
Overall, these results suggest that action-space token-level proxies are limited and do not reliably compose,
whereas \ourmethod{} offers a more robust and effective regularization signal for RL post-training.

\subsection{Ablation on the Choice of Hidden Layer}
\label{app:layer_ablation}

We focus on the final hidden layer because our exploration/exploitation metrics are defined in the \emph{semantic} space along the reasoning trajectory, and prior interpretability work~\citep{jing2025lingualens,sajjad2022analyzing,valeriani2023geometry,matthews2024semantics,servedio2025hidden,zhang2025reasoning} suggests that the last layers are most aligned with semantic meaning and model predictions. Empirically, using the final layer gives consistently better performance than using an intermediate layer: for example, GRPO + VERL with the last layer improves the average Pass@1 from 41.69\% (layer 14) to 43.44\%, and the average Pass@k from 61.52\% (layer 14) to 64.40\%. This subsection provides the detailed analysis supporting our design choice to base VERL on the final layer.

Intermediate layers in large language models can encode rich features. However, our notion of exploration and exploitation is explicitly defined in the semantic space of a reasoning trajectory. Existing interpretability studies~\citep{jing2025lingualens,sajjad2022analyzing,valeriani2023geometry,matthews2024semantics,servedio2025hidden,zhang2025reasoning} indicate that hidden states in the last layers are most tightly aligned with token-level semantics and the model's predictive distribution, while mid-layer representations tend to mix morphology, syntax, and other lower-level or task-specific signals. For this reason, we consider the final layer more suitable for semantic diversity metrics.

\paragraph{Intermediate-layer vs.\ final-layer VERL (Pass@1).}
We use Llama-3.2-3B-Instruct as the base model, which has 28 transformer layers. We implemented VERL on an intermediate layer (layer 14) and on the last layer, keeping everything else fixed. The Pass@1 results are reported in Tab.~\ref{tab:verl_layer_pass1}. Both VERL variants improve over GRPO, but the last-layer version clearly gives the strongest overall gains in average Pass@1.


\paragraph{Intermediate-layer vs.\ final-layer VERL (Pass@k).}
A similar pattern holds for Pass@k, summarized in Tab.~\ref{tab:verl_layer_passk}. Using the last layer yields the best average improvement, especially on the more challenging AIME-style benchmarks, further supporting the choice of the final hidden layer as the basis for our semantic exploration and exploitation metrics.




\section{Case Study}
\label{sec:case_study}
All case studies use Qwen2.5-7B as the base model. Outputs in the \textcolor{FigVanilla}{\textbf{gray boxes}} are produced by vanilla GRPO trained for 120 steps, whereas outputs in the \textcolor{FigOurs}{\textbf{purple boxes}} are produced by \ourmethod\ (GRPO + auxiliary shaping) trained for the same 120 steps.

\newcommand{\omitdetails}{\par\textit{(Details omitted as they are not crucial for this case study.)}\par}

\newcommand{\passkQaBSa}{
\omitdetails

1. \textbf{Numerator Condition}: The numerator is \( \sqrt{x-1} \). For the square root to be defined and result in a real number, the expression inside the square root must be non-negative.
   \[
   x - 1 \geq 0 \implies x \geq 1
   \]

2. \textbf{Denominator Condition}: The denominator is \( x - 2 \). For the function to be defined, the denominator must not be zero.
   \[
   x - 2 \neq 0 \implies x \neq 2
   \]

Combining these two conditions, we get:
\[
x \geq 1 \quad \text{and} \quad x \neq 2
\]

The smallest integer that satisfies \( x \geq 1 \) and \( x \neq 2 \) is \( x = 3 \).

Therefore, the smallest possible integer value for \( x \) such that \( f(x) \) has a real number value is \boxed{3}.
}
\newcommand{\passkQaBSb}{
To determine the smallest possible integer value for \( x \) such that \( f(x) = \frac{\sqrt{x-1}}{x-2} \) has a real number value, we need to ensure that both the numerator and the denominator satisfy their respective conditions for real numbers.

1. \textbf{Numerator Condition}: The expression inside the square root must be non-negative.
   \[
   x - 1 \geq 0 \implies x \geq 1
   \]

2. \textbf{Denominator Condition}: The denominator must not be zero to avoid division by zero.
   \[
   x - 2 \neq 0 \implies x \neq 2
   \]

Combining these conditions, we need \( x \geq 1 \) and \( x \neq 2 \). The smallest integer that satisfies these conditions is \( x = 3 \).

Therefore, the smallest possible integer value for \( x \) such that \( f(x) \) has a real number value is \(\boxed{3}\).
}
\newcommand{\passkQaBSc}{
To determine the smallest possible integer value for \( x \) such that \( f(x) = \frac{\sqrt{x-1}}{x-2} \) has a real number value, we need to ensure that both the numerator and the denominator satisfy certain conditions.

1. \textbf{Numerator Condition}: The expression inside the square root must be non-negative because the square root of a negative number is not a real number.
   \[
   x - 1 \geq 0 \implies x \geq 1
   \]

2. \textbf{Denominator Condition}: The denominator must not be zero because division by zero is undefined.
   \[
   x - 2 \neq 0 \implies x \neq 2
   \]

Combining these conditions, we get:
\[
x \geq 1 \quad \text{and} \quad x \neq 2
\]

The smallest integer that satisfies \( x \geq 1 \) and \( x \neq 2 \) is \( x = 3 \).
}
\newcommand{\passkQaBSd}{
\omitdetails

1. \textbf{Square Root Condition}: The expression inside the square root, \( x-1 \), must be greater than or equal to zero for the square root to be a real number.
   \[
   x - 1 \geq 0 \implies x \geq 1
   \]

2. \textbf{Denominator Condition}: The denominator, \( x-2 \), must not be zero to avoid division by zero.
   \[
   x - 2 \neq 0 \implies x \neq 2
   \]

Combining these two conditions, we get:
\[
x \geq 1 \quad \text{and} \quad x \neq 2
\]

The smallest integer that satisfies \( x \geq 1 \) and \( x \neq 2 \) is \( x = 3 \).

Therefore, the smallest possible integer value for \( x \) such that \( f(x) \) has a real number value is \boxed{3}.
}
\newcommand{\passkQaBSe}{

\omitdetails

1. For the square root to be defined, the expression inside it must be greater than or equal to zero:
   \[
   x - 1 \geq 0
   \]
   Solving this inequality:
   \[
   x \geq 1
   \]

2. For the denominator to be non-zero:
   \[
   x - 2 \neq 0
   \]
   Solving this equation:
   \[
   x \neq 2
   \]

\omitdetails

Therefore, the smallest possible integer value for \( x \) such that \( f(x) \) has a real number value is:
\[
\boxed{3}
\]
}
\newcommand{\passkQaBSf}{

\omitdetails

1. For the square root to be defined, the expression inside it must be greater than or equal to zero:
   \[
   x - 1 \geq 0
   \]
   Solving this inequality:
   \[
   x \geq 1
   \]

2. For the function to be defined, the denominator must not be zero:
   \[
   x - 2 \neq 0
   \]
   Solving this equation:
   \[
   x \neq 2
   \]

\omitdetails

Therefore, the smallest possible integer value for \( x \) such that \( f(x) \) has a real number value is \boxed{3}.
}
\newcommand{\passkQaBSg}{


\omitdetails

1. For the square root to be defined, we need:
   \[
   x - 1 \geq 0
   \]
   Solving this inequality:
   \[
   x \geq 1
   \]

2. For the denominator to be non-zero, we need:
   \[
   x - 2 \neq 0
   \]
   Solving this equation:
   \[
   x \neq 2
   \]

Combining these two conditions, \( x \) must be greater than or equal to 1 but not equal to 2. The smallest integer that satisfies this condition is \( x = 3 \).

Therefore, the smallest possible integer value for \( x \) such that \( f(x) \) has a real number value is \boxed{3}.
}
\newcommand{\passkQaBSh}{
\omitdetails

1. \textbf{Square Root Condition}: The expression inside the square root, \(x-1\), must be greater than or equal to zero for the square root to be a real number.
   \[
   x - 1 \geq 0 \implies x \geq 1
   \]

2. \textbf{Denominator Condition}: The denominator, \(x-2\), must not be zero to avoid division by zero.
   \[
   x - 2 \neq 0 \implies x \neq 2
   \]

Combining these two conditions, we get:
\[
x \geq 1 \quad \text{and} \quad x \neq 2
\]

The smallest integer that satisfies \(x \geq 1\) and \(x \neq 2\) is \(x = 3\).

Therefore, the smallest possible integer value for \(x\) such that \(f(x)\) has a real number value is \boxed{3}.
}
\newcommand{\passkQaBSi}{
\omitdetails

1. \textbf{Numerator Condition}: The expression inside the square root, \( x-1 \), must be non-negative because the square root of a negative number is not a real number. Therefore, we need:
   \[
   x - 1 \geq 0 \implies x \geq 1
   \]

2. \textbf{Denominator Condition}: The denominator, \( x-2 \), must not be zero because division by zero is undefined. Therefore, we need:
   \[
   x - 2 \neq 0 \implies x \neq 2
   \]

Combining these two conditions, we get:
\[
x \geq 1 \quad \text{and} \quad x \neq 2
\]

The smallest integer that satisfies \( x \geq 1 \) and \( x \neq 2 \) is \( x = 3 \).

Therefore, the smallest possible integer value for \( x \) such that \( f(x) \) has a real number value is:
\boxed{3}}
\newcommand{\passkQaBSj}{
\omitdetails

1. \textbf{Numerator Condition}: The expression inside the square root, \( x-1 \), must be non-negative because the square root of a negative number is not a real number. Therefore, we need:
   \[
   x - 1 \geq 0 \implies x \geq 1
   \]

2. \textbf{Denominator Condition}: The denominator, \( x-2 \), must not be zero because division by zero is undefined. Therefore, we need:
   \[
   x - 2 \neq 0 \implies x \neq 2
   \]

Combining these two conditions, we get:
\[
x \geq 1 \quad \text{and} \quad x \neq 2
\]

The smallest integer that satisfies \( x \geq 1 \) and \( x \neq 2 \) is \( x = 3 \).

Therefore, the smallest possible integer value for \( x \) such that \( f(x) \) has a real number value is \boxed{3}.}
\newcommand{\passkQaBSk}{



\omitdetails

1. For the square root to be defined, we need:
\[ x - 1 \geq 0 \]
\[ x \geq 1 \]

2. For the denominator to be non-zero, we need:
\[ x - 2 \neq 0 \]
\[ x \neq 2 \]

Combining these two conditions, we get:
\[ x \geq 1 \]
\[ x \neq 2 \]

The smallest integer that satisfies \( x \geq 1 \) and \( x \neq 2 \) is \( x = 3 \).

Therefore, the smallest possible integer value for \( x \) such that \( f(x) \) has a real number value is \boxed{3}.}
\newcommand{\passkQaBSl}{
\omitdetails
1. \textbf{Numerator Condition}: The expression under the square root, \( x-1 \), must be non-negative because the square root of a negative number is not a real number. Therefore, we need:
   \[
   x - 1 \geq 0 \implies x \geq 1
   \]

2. \textbf{Denominator Condition}: The denominator, \( x-2 \), must not be zero because division by zero is undefined. Therefore, we need:
   \[
   x - 2 \neq 0 \implies x \neq 2
   \]

Combining these two conditions, we get:
\[
x \geq 1 \quad \text{and} \quad x \neq 2
\]

The smallest integer that satisfies \( x \geq 1 \) and \( x \neq 2 \) is \( x = 3 \).

Therefore, the smallest possible integer value for \( x \) such that \( f(x) \) has a real number value is:
\boxed{3}}
\newcommand{\passkQaBSm}{
\omitdetails

1. \textbf{Numerator Condition}: The numerator is \( \sqrt{x-1} \). For the square root to be defined and result in a real number, the expression inside the square root must be non-negative.
   \[
   x - 1 \geq 0 \implies x \geq 1
   \]

2. \textbf{Denominator Condition}: The denominator is \( x - 2 \). For the function to be defined, the denominator must not be zero.
   \[
   x - 2 \neq 0 \implies x \neq 2
   \]

Combining these two conditions, we need \( x \) to be greater than or equal to 1, but not equal to 2. Therefore, the smallest possible integer value for \( x \) that satisfies both conditions is 3.

\boxed{3}}
\newcommand{\passkQaBSn}{
\omitdetails
1. \textbf{Square Root Condition}: The expression inside the square root, \( x-1 \), must be greater than or equal to zero for the square root to be a real number.
   \[
   x - 1 \geq 0 \implies x \geq 1
   \]

2. \textbf{Denominator Condition}: The denominator, \( x-2 \), must not be zero to avoid division by zero.
   \[
   x - 2 \neq 0 \implies x \neq 2
   \]

Combining these two conditions, we get:
\[
x \geq 1 \quad \text{and} \quad x \neq 2
\]

The smallest integer that satisfies \( x \geq 1 \) and \( x \neq 2 \) is \( x = 3 \).

Therefore, the smallest possible integer value for \( x \) such that \( f(x) \) has a real number value is \boxed{3}.}
\newcommand{\passkQaBSo}{To determine the smallest possible integer value for \( x \) such that \( f(x) = \frac{\sqrt{x-1}}{x-2} \) has a real number value, we need to ensure that both the numerator and the denominator satisfy their respective conditions for real numbers.

1. \textbf{Numerator Condition}: The expression inside the square root must be non-negative.
   \[
   x - 1 \geq 0 \implies x \geq 1
   \]

2. \textbf{Denominator Condition}: The denominator must not be zero.
   \[
   x - 2 \neq 0 \implies x \neq 2
   \]

Combining these conditions, we need \( x \) to be greater than or equal to 1 but not equal to 2. Therefore, the smallest possible integer value for \( x \) that satisfies both conditions is 3.

\boxed{3}}
\newcommand{\passkQaBSp}{
\omitdetails

1. \textbf{Numerator Condition}: The numerator is \( \sqrt{x-1} \). For the square root to be a real number, the expression inside the square root must be non-negative.
   \[
   x - 1 \geq 0 \implies x \geq 1
   \]

2. \textbf{Denominator Condition}: The denominator is \( x - 2 \). For the function to be defined, the denominator must not be zero.
   \[
   x - 2 \neq 0 \implies x \neq 2
   \]

Combining these two conditions, we get:
\[
x \geq 1 \quad \text{and} \quad x \neq 2
\]

The smallest integer that satisfies \( x \geq 1 \) and \( x \neq 2 \) is \( x = 3 \).

Therefore, the smallest possible integer value for \( x \) such that \( f(x) \) has a real number value is \boxed{3}.}
\newcommand{\passkQaOURa}{
\omitdetails
1. \textbf{Numerator Condition}: The numerator is \( \sqrt{x-1} \). For the square root to be defined and real, the expression inside the square root must be non-negative.
   \[
   x - 1 \geq 0 \implies x \geq 1
   \]

2. \textbf{Denominator Condition}: The denominator is \( x - 2 \). For the function to be defined, the denominator must not be zero.
   \[
   x - 2 \neq 0 \implies x \neq 2
   \]

Combining these conditions, we need \( x \) to be greater than or equal to 1, but not equal to 2. The smallest integer that satisfies these conditions is \( x = 3 \).

Therefore, the smallest possible integer value for \( x \) such that \( f(x) \) has a real number value is \(\boxed{3}\).}
\newcommand{\passkQaOURb}{


\omitdetails
1. For the square root to be defined in the real numbers, the expression inside it must be greater than or equal to zero:
   \[
   x - 1 \geq 0
   \]
   Solving this inequality:
   \[
   x \geq 1
   \]

2. For the function to be defined, the denominator must not be zero:
   \[
   x - 2 \neq 0
   \]
   Solving this equation:
   \[
   x \neq 2
   \]

\omitdetails
Thus, the smallest possible integer value for \( x \) such that \( f(x) \) has a real number value is \(\boxed{3}\).}
\newcommand{\passkQaOURc}{
\omitdetails
1. The expression inside the square root must be non-negative.
2. The denominator must not be zero.



\omitdetails
\#\#\# Step 3: Combine the conditions
From Step 1, we know \( x \geq 1 \). From Step 2, we know \( x \neq 2 \). Therefore, the smallest integer value for \( x \) that satisfies both conditions is the smallest integer greater than 1, which is 3.

\#\#\# Final Answer
The smallest possible integer value for \( x \) such that \( f(x) \) has a real number value is \(\boxed{3}\).}
\newcommand{\passkQaOURd}{



\omitdetails

\#\#\# Combining the Conditions
From Condition 1, we have \( x \geq 1 \).
From Condition 2, we have \( x \neq 2 \).

Combining these, the smallest integer value for \( x \) that satisfies both conditions is \( x = 3 \).

\#\#\# Verification
Let's verify that \( x = 3 \) works:
- For \( x = 3 \), \( \sqrt{x-1} = \sqrt{3-1} = \sqrt{2} \), which is a real number.
- The denominator \( x-2 = 3-2 = 1 \), which is not zero.

Since both conditions are satisfied, the smallest possible integer value for \( x \) is indeed \( 3 \).

Thus, the final answer is:
\[
\boxed{3}
\]}
\newcommand{\passkQaOURe}{
\omitdetails
1. \textbf{Numerator Condition}: The expression under the square root, \( x - 1 \), must be non-negative because the square root of a negative number is not a real number. Therefore, we need:
   \[
   x - 1 \geq 0 \implies x \geq 1
   \]

2. \textbf{Denominator Condition}: The denominator, \( x - 2 \), must not be zero because division by zero is undefined. Therefore, we need:
   \[
   x - 2 \neq 0 \implies x \neq 2
   \]

Combining these two conditions, we get:
\[
x \geq 1 \quad \text{and} \quad x \neq 2
\]

The smallest integer that satisfies \( x \geq 1 \) but not \( x = 2 \) is \( x = 1 \).

Therefore, the smallest possible integer value for \( x \) such that \( f(x) \) has a real number value is \(\boxed{1}\).}
\newcommand{\passkQaOURf}{




\omitdetails

\#\#\# Combining the Conditions
From Condition 1, we know \( x \geq 1 \).
From Condition 2, we know \( x \neq 2 \).

The smallest integer that satisfies \( x \geq 1 \) but not \( x = 2 \) is \( x = 1 \).

\omitdetails
Since \( f(1) \) is a real number, \( x = 1 \) is indeed a valid solution.

Therefore, the smallest possible integer value for \( x \) such that \( f(x) \) has a real number value is \(\boxed{1}\).}
\newcommand{\passkQaOURg}{


\omitdetails

1. For \( \sqrt{x-1} \) to be defined, we need:
   \[
   x - 1 \geq 0 \implies x \geq 1
   \]

2. For the denominator \( x-2 \) to be non-zero, we need:
   \[
   x - 2 \neq 0 \implies x \neq 2
   \]

Combining these two conditions, we get:
\[
x \geq 1 \quad \text{and} \quad x \neq 2
\]

The smallest integer that satisfies \( x \geq 1 \) and \( x \neq 2 \) is \( x = 1 \).

\omitdetails
Since \( f(1) = 0 \) is a real number, our solution is confirmed.

Therefore, the smallest possible integer value for \( x \) such that \( f(x) \) has a real number value is \(\boxed{1}\).}
\newcommand{\passkQaOURh}{

\omitdetails

1. For \( \sqrt{x-1} \) to be defined in the real numbers, we need:
   \[
   x - 1 \geq 0 \implies x \geq 1
   \]

2. For the denominator \( x-2 \) to not be zero, we need:
   \[
   x - 2 \neq 0 \implies x \neq 2
   \]

Combining these two conditions, \( x \) must be greater than or equal to 1 but not equal to 2. The smallest integer that satisfies this condition is \( x = 3 \).

Therefore, the smallest possible integer value for \( x \) such that \( f(x) \) has a real number value is \(\boxed{3}\).}
\newcommand{\passkQaOURi}{
\omitdetails

1. \textbf{Numerator Condition}: The expression inside the square root, \( x-1 \), must be non-negative because the square root of a negative number is not a real number. Therefore, we need:
   \[
   x - 1 \geq 0 \implies x \geq 1
   \]

2. \textbf{Denominator Condition}: The denominator, \( x-2 \), must not be zero because division by zero is undefined. Therefore, we need:
   \[
   x - 2 \neq 0 \implies x \neq 2
   \]

Combining these two conditions, we get:
\[
x \geq 1 \quad \text{and} \quad x \neq 2
\]

The smallest integer that satisfies \( x \geq 1 \) and \( x \neq 2 \) is \( x = 1 \).

Therefore, the smallest possible integer value for \( x \) such that \( f(x) \) has a real number value is \(\boxed{1}\).}
\newcommand{\passkQaOURj}{
\omitdetails

1. \textbf{Numerator Condition}: The expression under the square root must be non-negative.
   \[
   x - 1 \geq 0 \implies x \geq 1
   \]

2. \textbf{Denominator Condition}: The denominator must not be zero.
   \[
   x - 2 \neq 0 \implies x \neq 2
   \]

Combining these two conditions, we get:
\[
x \geq 1 \quad \text{and} \quad x \neq 2
\]

The smallest integer that satisfies \( x \geq 1 \) and \( x \neq 2 \) is \( x = 3 \).

Therefore, the smallest possible integer value for \( x \) such that \( f(x) \) has a real number value is \(\boxed{3}\).}
\newcommand{\passkQaOURk}{
\omitdetails
1. \textbf{Numerator Condition}: The expression inside the square root must be non-negative.
   \[
   x - 1 \geq 0 \implies x \geq 1
   \]
   This ensures that \( \sqrt{x-1} \) is a real number.

2. \textbf{Denominator Condition}: The denominator must not be zero.
   \[
   x - 2 \neq 0 \implies x \neq 2
   \]
   This ensures that the function is defined.

Combining these two conditions, we get:
\[
x \geq 1 \quad \text{and} \quad x \neq 2
\]

The smallest integer that satisfies \( x \geq 1 \) and \( x \neq 2 \) is \( x = 3 \).

Therefore, the smallest possible integer value for \( x \) such that \( f(x) \) has a real number value is \(\boxed{3}\).}
\newcommand{\passkQaOURl}{

\omitdetails

1. For the square root to be defined, we need:
   \[
   x - 1 \geq 0
   \]
   Solving this inequality:
   \[
   x \geq 1
   \]

2. For the denominator to be non-zero, we need:
   \[
   x - 2 \neq 0
   \]
   Solving this equation:
   \[
   x \neq 2
   \]

Combining these two conditions, \( x \) must be greater than or equal to 1 but not equal to 2. The smallest integer that satisfies this condition is \( x = 3 \).

Therefore, the smallest possible integer value for \( x \) such that \( f(x) \) has a real number value is \(\boxed{3}\).}
\newcommand{\passkQaOURm}{
\omitdetails
1. The expression inside the square root must be non-negative.
2. The denominator must not be zero.

Let's address these conditions one by one:

1. For the square root to be defined, the expression inside it must be greater than or equal to zero:
   \[
   x - 1 \geq 0 \implies x \geq 1
   \]

2. For the denominator to be non-zero:
   \[
   x - 2 \neq 0 \implies x \neq 2
   \]

Combining these conditions, we need \( x \) to be greater than or equal to 1 but not equal to 2. The smallest integer that satisfies this is \( x = 3 \).

Therefore, the smallest possible integer value for \( x \) such that \( f(x) \) has a real number value is \(\boxed{3}\).}
\newcommand{\passkQaOURn}{
\omitdetails

1. \textbf{Numerator Condition}: The expression inside the square root must be non-negative because the square root of a negative number is not a real number.
   \[
   x - 1 \geq 0 \implies x \geq 1
   \]

2. \textbf{Denominator Condition}: The denominator must not be zero because division by zero is undefined.
   \[
   x - 2 \neq 0 \implies x \neq 2
   \]

Combining these two conditions, we get:
\[
x \geq 1 \quad \text{and} \quad x \neq 2
\]

The smallest integer that satisfies \( x \geq 1 \) and \( x \neq 2 \) is \( x = 3 \).

Therefore, the smallest possible integer value for \( x \) such that \( f(x) \) has a real number value is \(\boxed{3}\).}
\newcommand{\passkQaOURo}{
\omitdetails
1. \textbf{Numerator Condition}: The expression under the square root, \( x - 1 \), must be non-negative because the square root of a negative number is not a real number.
   \[
   x - 1 \geq 0 \implies x \geq 1
   \]

2. \textbf{Denominator Condition}: The denominator, \( x - 2 \), must not be zero because division by zero is undefined.
   \[
   x - 2 \neq 0 \implies x \neq 2
   \]

Combining these conditions, we need \( x \geq 1 \) and \( x \neq 2 \). The smallest integer that satisfies these conditions is \( x = 1 \).

\omitdetails

Therefore, the smallest possible integer value for \( x \) such that \( f(x) \) has a real number value is \(\boxed{1}\).}
\newcommand{\passkQaOURp}{To determine the smallest possible integer value for \( x \) such that \( f(x) = \frac{\sqrt{x-1}}{x-2} \) has a real number value, we need to ensure that both the numerator and the denominator satisfy their respective conditions for real numbers.

1. \textbf{Numerator Condition}: The expression inside the square root must be non-negative.
   \[
   x - 1 \geq 0 \implies x \geq 1
   \]

2. \textbf{Denominator Condition}: The denominator must not be zero.
   \[
   x - 2 \neq 0 \implies x \neq 2
   \]

Combining these conditions, we need \( x \) to be greater than or equal to 1 but not equal to 2. The smallest integer that satisfies this condition is \( x = 3 \).

Therefore, the smallest possible integer value for \( x \) such that \( f(x) \) has a real number value is \(\boxed{3}\).}

\newcommand{\passkQbBSa}{To solve this problem, we need to use some geometric properties and trigonometry. Let's break it down step by step.

1. \textbf{Understanding the Problem:}
   \omitdetails

2. \textbf{Geometric Insight:}
   \omitdetails

3. \textbf{Using Trigonometry:}
   \omitdetails

4. \textbf{Calculating the Distance:}
   - Let the distance AB be x. Since $\angle$ABC is 60° and we need to maximize AB, we use the Law of Cosines in triangle ABC:
   \omitdetails

Therefore, the square of the distance between Abdul and Bharat is \boxed{432}.}
\newcommand{\passkQbBSb}{
To solve this problem, we can use the Law of Sines in trigonometry. Let's denote the distance between Abdul and Bharat as \( x \) and the distance between Chiang and Bharat as \( y \). We know that the angle between Abdul and Chiang as seen from Bharat is 60°, and the distance between Abdul and Chiang is 48 feet. 

Given:
- \( \angle ABC = 60^\circ \)
- \( AB = 48 \) feet
- We need to find \( x^2 \)

First, we apply the Law of Sines in triangle \( ABC \):
\omitdetails













Thus, the square of the distance between Abdul and Bharat is \boxed{9216}.
}
\newcommand{\passkQbBSc}{To solve this problem, we can use the Law of Sines in trigonometry. Let's denote the distance between Abdul and Bharat as \( AB \), and the distance between Chiang and Bharat as \( BC \). Since Bharat is as far from Abdul as possible to form a 60° angle, triangle \( ABC \) is a right triangle with the right angle at Chiang. This means \( \angle ACB = 90^\circ \) and \( \angle BAC = 60^\circ \).







\omitdetails

Thus, the square of the distance between Abdul and Bharat is \boxed{9216}.}
\newcommand{\passkQbBSd}{To solve this problem, we can use the Law of Sines in trigonometry. Let's break it down step by step:

1. \textbf{Identify the given information:}
\omitdetails
2. \textbf{Set up the triangle:}
\omitdetails
3. \textbf{Use the Law of Sines:}
\omitdetails
4. \textbf{Find the square of the distance:}
   The distance between Abdul and Bharat is 24 feet. The square of this distance is:
   \[
   24^2 = 576
   \]

Therefore, the square of the distance between Abdul and Bharat is \boxed{576}.}
\newcommand{\passkQbBSe}{To solve this problem, we can use the Law of Sines and some geometric properties. Let's break it down step by step.

1. \textbf{Identify the triangle and its properties:}
\omitdetails

2. \textbf{Apply the Law of Sines:}
\omitdetails

3. \textbf{Calculate the distances:}
\omitdetails

4. \textbf{Find the square of the distance:}
\omitdetails

So, the square of the distance between Abdul and Bharat is \boxed{768}.}
\newcommand{\passkQbBSf}{To solve this problem, we need to use some geometric properties and trigonometry. Let's break it down step by step.

1. \textbf{Understand the Problem:}
\omitdetails

2. \textbf{Visualize the Problem:}
\omitdetails

3. \textbf{Use Geometric Properties:}
\omitdetails

4. \textbf{Apply Trigonometry:}
\omitdetails

5. \textbf{Calculate the Distance:}
   - AB = 2 * AC = 2 * 48 = 96 feet.
   - The square of the distance AB is \(96^2\).

6. \textbf{Final Calculation:}
   - \(96^2 = 9216\).

So, the square of the distance between Abdul and Bharat is \boxed{9216}.}
\newcommand{\passkQbBSg}{To solve the problem, we need to determine the distance between Abdul and Bharat, given that the angle between their lines of sight to Chiang is 60° and that Bharat is positioned to be as far from Abdul as possible.


\omitdetails

Using the Law of Cosines in triangle \( ABC \):
\omitdetails
\[ BC^2 = d^2 + 2304 - 48d \]


\omitdetails
The square of the distance between Abdul and Bharat is:
\[ d^2 = 96^2 = 9216 \]

Thus, the square of the distance between Abdul and Bharat is \boxed{9216}.}
\newcommand{\passkQbBSh}{
\omitdetails
1. \textbf{Identify the triangle and the given information:}
\omitdetails
2. \textbf{Use the Law of Sines:}
\omitdetails
3. \textbf{Maximize \( AB \):}
\omitdetails
4. \textbf{Apply the Pythagorean theorem:}

\omitdetails
Thus, the square of the distance between Abdul and Bharat is \boxed{9216}.}
\newcommand{\passkQbBSi}{To solve this problem, we can use the Law of Sines and some geometric properties. Let's break it down step by step.

1. \textbf{Identify the triangle and the given information:}
\omitdetails
2. \textbf{Determine the largest possible angle at A:}
\omitdetails
3. \textbf{Use the Pythagorean theorem:}
\omitdetails
4. \textbf{Find the square of the distance between Abdul and Bharat:}


\omitdetails
So, the square of the distance between Abdul and Bharat is \(\boxed{9216}\).}
\newcommand{\passkQbBSj}{To solve this problem, we can use some geometric properties and trigonometry. Let's break it down step by step:

1. \textbf{Identify the triangle formed}: \omitdetails

2. \textbf{Use the Law of Sines}: \omitdetails

3. \textbf{Maximize the distance}: \omitdetails

4. \textbf{Square the distance}: \omitdetails

Let's calculate it:

\[
48^2 = 2304
\]

So, the square of the distance between Abdul and Bharat is \(\boxed{2304}\).}
\newcommand{\passkQbBSk}{To solve this problem, we need to use some geometric principles, specifically involving the Law of Sines and properties of triangles.

1. \textbf{Identify the triangle and the given information:}
\omitdetails
2. \textbf{Use the Law of Sines to find the distance BC (the distance from Bharat to Chiang):}

\omitdetails
3. \textbf{Find the distance AC (the distance from Abdul to Bharat):}
\omitdetails
Therefore, the square of the distance between Abdul and Bharat is \boxed{11520}.}
\newcommand{\passkQbBSl}{
\omitdetails
1. \textbf{Identify the known values:}
\omitdetails
2. \textbf{Set up the triangle:}
\omitdetails
3. \textbf{Apply the Law of Sines:}
\omitdetails
4. \textbf{Determine the angles:}
\omitdetails
5. \textbf{Simplify the Law of Sines equation:}
\omitdetails
Therefore, the square of the distance between Abdul and Bharat is \boxed{2304}.}
\newcommand{\passkQbBSm}{
\omitdetails
1. \textbf{Identify the triangle and its properties:}
   \omitdetails

2. \textbf{Use the Law of Sines:}
\omitdetails
3. \textbf{Apply the Law of Sines:}
\omitdetails
4. \textbf{Find the square of the distance:}
\omitdetails
Therefore, the square of the distance between Abdul and Bharat is \boxed{2304}.}
\newcommand{\passkQbBSn}{
\omitdetails
1. \textbf{Identify the given information:}
\omitdetails
2. \textbf{Visualize the problem:}
\omitdetails
3. \textbf{Use the Law of Sines:}
\omitdetails
4. \textbf{Properties of an equilateral triangle:}
\omitdetails
5. \textbf{Calculate the square of the distance:}
   
\omitdetails
Thus, the square of the distance between Abdul and Bharat is \boxed{2304}.

 Providethefinalanswerwithin\boxed{}.}
\newcommand{\passkQbBSo}{

\omitdetails
1. Recognize that the triangle \( ABC \) is a right triangle with \( \angle BAC = 90^\circ \) and \( \angle ABC = 60^\circ \). Therefore, \( \angle ACB = 30^\circ \).

\omitdetails
3. Since \( AC = 48 \) feet, and in a 30-60-90 triangle, the side opposite the 60° angle is \( \sqrt{3} \) times the side opposite the 30° angle, we can find \( AB \):
   \[
   AB = 48 \sqrt{3}
   \]

\omitdetails
Therefore, the square of the distance between Abdul and Bharat is \boxed{6912} feet.}
\newcommand{\passkQbBSp}{

First, we need to find the distance \( AB \) using the Law of Sines. In triangle \( ABC \):





\omitdetails
Now, we need to find the square of the distance \( AB \):

\omitdetails
Thus, the square of the distance between Abdul and Bharat is \boxed{1728}.

To verify the accuracy of this result, we can use Python to perform the calculations.
\omitdetails




}
\newcommand{\passkQbOURa}{
\omitdetails
Given that \( \angle ABC = 60^\circ \) and \( AC = 48 \) feet, we can use the Law of Sines to find \( AB \):





\omitdetails
The square of the distance \( AB \) is:

\[
(32\sqrt{3})^2 = 32^2 \times 3 = 1024 \times 3 = 3072
\]

Thus, the square of the distance between Abdul and Bharat is \(\boxed{3072}\).}
\newcommand{\passkQbOURb}{To solve this problem, we can use the Law of Sines in trigonometry.
\omitdetails
Given that \( AC = 48 \) feet, we can use the Law of Sines to find \( AB \):

\[
\frac{AB}{\sin(60^\circ)} = \frac{AC}{\sin(30^\circ)}
\]




\omitdetails
Solving for \( AB \):

\[
AB = 96 \times \frac{\sqrt{3}}{2} = 48\sqrt{3}
\]


\omitdetails
Therefore, the square of the distance between Abdul and Bharat is \(\boxed{6912}\).}
\newcommand{\passkQbOURc}{
\omitdetails
1. \textbf{Identify the given information:}
\omitdetails
2. \textbf{Visualize the problem:}
\omitdetails
3. \textbf{Determine the type of triangle:}
   - Since angle ABC and angle ACB are both 60°, triangle ABC is an equilateral triangle. In an equilateral triangle, all sides are equal.

4. \textbf{Calculate the side length of the equilateral triangle:}
   - Since AC = 48 feet and triangle ABC is equilateral, AB (the distance from Abdul to Bharat) is also 48 feet.

\omitdetails
So, the square of the distance between Abdul and Bharat is \(\boxed{2304}\).}
\newcommand{\passkQbOURd}{To solve this problem, we can use the Law of Sines and some geometric properties. Let's denote the positions of Abdul, Chiang, and Bharat as points \(A\), \(C\), and \(B\) respectively. The distance between \(A\) and \(C\) is given as 48 feet. The angle \( \angle ACB \) is 60°. We need to find the square of the distance \(AB\), which is the farthest distance Bharat can be from Abdul.




\omitdetails
Therefore, the square of the distance between Abdul and Bharat is \(\boxed{6912}\).}
\newcommand{\passkQbOURe}{To solve this problem, we need to use some geometric principles. Let's denote the positions of Abdul, Chiang, and Bharat as points A, C, and B respectively. We are given that the distance between A and C (AC) is 48 feet, and the angle ABC is 60°. We need to find the distance between A and B (AB), and then square it.

\omitdetails
Step 3: Recognize that if triangle ABC is equilateral, then all sides are equal. Therefore, AB = BC = AC = 48 feet.

Step 4: Calculate the square of the distance AB. Since AB = 48 feet, we square this value to get AB².

\[ AB^2 = 48^2 = 2304 \]

So, the square of the distance between Abdul and Bharat is \(\boxed{2304}\).}
\newcommand{\passkQbOURf}{To find the square of the distance between Abdul and Bharat, we can use the Law of Sines in trigonometry. Let's denote the distance between Abdul and Bharat as \( AB \), and the distance between Chiang and Bharat as \( BC \). Since Bharat is as far from Abdul as possible, triangle \( ABC \) is a right triangle with \( \angle ACB = 90^\circ \).

Given:
- \( \angle ABC = 60^\circ \)
- \( AC = 48 \) feet

We want to find \( AB^2 \).

First, we apply the Law of Sines in triangle \( ABC \):


\omitdetails
Thus, the square of the distance between Abdul and Bharat is \(\boxed{3072}\).}
\newcommand{\passkQbOURg}{

\omitdetails
In a right triangle, if one of the angles is 60°, the triangle is a 30-60-90 triangle. In a 30-60-90 triangle, the sides are in the ratio \( 1 : \sqrt{3} : 2 \). Here, \( AC \) is the hypotenuse, so we can set up the relationship as follows:

\[ AC = 2 \times AB \]
\[ 48 = 2 \times AB \]
\[ AB = \frac{48}{2} = 24 \text{ feet} \]

We need to find the square of the distance \( AB \):

\[ AB^2 = 24^2 = 576 \]

Thus, the square of the distance between Abdul and Bharat is \(\boxed{576}\).}
\newcommand{\passkQbOURh}{






\omitdetails
Using the properties of a 30-60-90 triangle, we know that the side opposite the 60° angle is \( \sqrt{3} \) times the side opposite the 30° angle. Let's denote the side opposite the 30° angle (which is half the distance between Abdul and Chiang) as \( x \). Therefore, \( AB = x\sqrt{3} \) and \( BC = 2x \).

















\omitdetails
\[
AB^2 = 2304 (21 - 12\sqrt{3})
\]

Since we are asked to find the square of the distance, and we have already calculated it, we can box the final answer:

\[
\boxed{2304}
\]}
\newcommand{\passkQbOURi}{To solve this problem, we can use the Law of Sines in trigonometry. 


\omitdetails
Using the Law of Sines:
 \[ \frac{AB}{\sin(120^\circ)} = \frac{AC}{\sin(60^\circ)} \]





\omitdetails
Therefore, the square of the distance between Abdul and Bharat is \(\boxed{3072}\).}
\newcommand{\passkQbOURj}{To solve this problem, we need to use some geometric properties and trigonometric relationships. Let's break it down step by step.

1. \textbf{Identify the given information:}
\omitdetails
2. \textbf{Visualize the problem:}
\omitdetails
3. \textbf{Use the Law of Cosines:}
\omitdetails
4. \textbf{Apply the Law of Cosines:}
\omitdetails
5. \textbf{Find the square of the distance:}
   - The square of the distance between Abdul and Bharat is \(d^2 = 2304\).

Therefore, the square of the distance between Abdul and Bharat is \(\boxed{2304}\).}
\newcommand{\passkQbOURk}{To solve this problem, we can use the Law of Sines and some geometric properties.

\omitdetails
Given that the angle ABC is 60° and AB is maximized, the triangle ABC is a right triangle with angle ACB being 90° (since the angle at Bharat is 60° and the sum of angles in a triangle is 180°, the remaining angle at A must be 30°, making AB the hypotenuse of a 30-60-90 triangle).



\omitdetails
Therefore, the square of the distance between Abdul and Bharat is \(\boxed{3072}\).}
\newcommand{\passkQbOURl}{To solve this problem, we need to use some geometric principles. Let's denote the positions of Abdul, Chiang, and Bharat as points \( A \), \( C \), and \( B \) respectively. The distance between \( A \) and \( C \) is 48 feet, and we need to find the distance \( AB \) when the angle \( \angle ABC \) is 60° and \( B \) is as far from \( A \) as possible.

1. \textbf{Identify the configuration}:\omitdetails 

2. \textbf{Use the Law of Cosines}:\omitdetails 

3. \textbf{Substitute into the Law of Cosines}:\omitdetails

So, the square of the distance between Abdul and Bharat is \(\boxed{3840}\) square feet.}
\newcommand{\passkQbOURm}{To solve this problem, we need to use some geometric principles, specifically involving the Law of Sines and properties of triangles. Let's break it down step by step:

1. \textbf{Identify the given information:}
\omitdetails
2. \textbf{Visualize the problem:}
\omitdetails
3. \textbf{Use the Law of Sines:}
\omitdetails
4. \textbf{Determine the angles:}
\omitdetails
5. \textbf{Calculate the distance AB:}

\omitdetails
Therefore, the square of the distance between Abdul and Bharat is \boxed{3072}.}
\newcommand{\passkQbOURn}{To find the square of the distance between Abdul and Bharat, let's denote the distance between Abdul and Bharat as \(d\), and the distance between Chiang and Bharat as \(d'\). Since the angle formed by the lines of sight to Abdul and Chiang is 60°, we can use the Law of Cosines in the triangle formed by Abdul, Chiang, and Bharat.

The Law of Cosines states that for any triangle with sides \(a\), \(b\), and \(c\), and the angle \(\theta\) opposite side \(c\):
\[ c^2 = a^2 + b^2 - 2ab \cos(\theta) \]



\omitdetails
Therefore, the square of the distance between Abdul and Bharat is \(\boxed{2304}\).}
\newcommand{\passkQbOURo}{To solve this problem, we'll use some geometric principles, particularly the Law of Sines and properties of triangles. Let's break down the problem step-by-step:

1. \textbf{Identify the given information:}
\omitdetails
2. \textbf{Visualize the problem:}
\omitdetails
3. \textbf{Use the Law of Sines:}
\omitdetails
4. \textbf{Apply the Law of Sines:}
\omitdetails
5. \textbf{Use trigonometry in the right triangle:}
\omitdetails
Therefore, the square of the distance between Abdul and Bharat is \(\boxed{1728}\).}
\newcommand{\passkQbOURp}{To solve this problem, we need to use some geometric properties and trigonometry. Let's denote the positions of Abdul, Chiang, and Bharat as points A, C, and B respectively in the field. We know that AC = 48 feet and the angle $\angle$ACB = 60°. We want to find the distance AB, which we'll denote as \(d\), and then find \(d^2\).

1. \textbf{Identify the type of triangle:}
\omitdetails
2. \textbf{Apply the Law of Cosines:}
\omitdetails
3. \textbf{Maximize the distance \(d\):}
\omitdetails
Therefore, the square of the distance between Abdul and Bharat is \(\boxed{1728}\).}
\newlength{\BWidth}    \setlength{\BWidth}{\textwidth} 
\newlength{\BRadius}   \setlength{\BRadius}{6pt}           
\newlength{\LRadius}   \setlength{\LRadius}{1pt}           
\newlength{\LSep}   \setlength{\LSep}{3pt}              
\newlength{\LShift}   \setlength{\LShift}{5pt}           
\newlength{\BLine}     \setlength{\BLine}{1pt}             
\newlength{\BannerSep} \setlength{\BannerSep}{8pt}         
\newcommand{\BFont}{\scriptsize}
\newcommand{\LFont}{\footnotesize}

\newcommand{\BannerThemeQuestion}{%
  \def\BFill{yellow!12!white}
  \def\BDraw{orange!70!black}
  \def\LFill{yellow!25}
  \def\LDraw{orange!70!black}
}

\newcommand{\BannerThemeAnswer}{%
  \def\BFill{cyan!10!white}%
  \def\BDraw{blue!60!black}%
  \def\LFill{cyan!20}%
  \def\LDraw{blue!60!black}%
}

\newcommand{\BannerThemeBaseline}{%
  \def\BFill{gray!10}%
  \def\BDraw{gray!65}%
  \def\LFill{gray!20}%
  \def\LDraw{gray!65}%
}

\newcommand{\BannerThemeOurs}{%
  \def\BFill{violet!10!white}%
  \def\BDraw{purple!70!black}%
  \def\LFill{violet!10}%
  \def\LDraw{purple!70!black}%
}

\begin{figure*}[!t]
    \centering
\begin{tikzpicture}
  \BannerThemeQuestion
  \coordinate (C) at (\dimexpr.5\linewidth\relax,0);

  \node[
    draw=\BDraw, fill=\BFill,
    rounded corners=\BRadius, line width=\BLine,
    minimum width=\BWidth,
    text width=\dimexpr\BWidth - \BLine - 2\BannerSep\relax,
    inner sep=\BannerSep,
    align=left,
    font=\BFont
  ] (main) at (C) {
What is the smallest possible integer value of $x$ if $12\div(x+1)$ is an integer?
  };

  \node[
    draw=\LDraw, fill=\LFill,
    rounded corners=\LRadius, inner sep=\LSep,
     anchor=west,
    line width=\BLine, font=\LFont
  ] at ($(main.north west)+(\LShift,0)$) {\textbf{Question}};

\end{tikzpicture}

\vspace{\baselineskip}

\begin{tikzpicture}
  \BannerThemeAnswer
  \coordinate (C) at (\dimexpr.5\linewidth\relax,0);

  \node[
    draw=\BDraw, fill=\BFill,
    rounded corners=\BRadius, line width=\BLine,
    minimum width=\BWidth,
    text width=\dimexpr\BWidth - \BLine - 2\BannerSep\relax,
    inner sep=\BannerSep,
    align=left,
    font=\BFont
  ] (main) at (C) {-13};

  \node[
    draw=\LDraw, fill=\LFill,
    rounded corners=\LRadius, inner sep=\LSep,
     anchor=west,
    line width=\BLine, font=\LFont
  ] at ($(main.north west)+(\LShift,0)$) {\textbf{Ground Truth}};

\end{tikzpicture}

\vspace{\baselineskip}

\begin{tikzpicture}
  \BannerThemeBaseline
  \coordinate (C) at (\dimexpr.5\linewidth\relax,0);

  \node[
    draw=\BDraw, fill=\BFill,
    rounded corners=\BRadius, line width=\BLine,
    minimum width=\BWidth,
    text width=\dimexpr\BWidth - \BLine - 2\BannerSep\relax,
    inner sep=\BannerSep,
    align=left,
    font=\BFont
  ] (main) at (C) {
To find the smallest possible integer value of \(x\) such that \(\frac{12}{x+1}\) is an integer, we need to consider the factors of 12. The expression \(\frac{12}{x+1}\) will be an integer if \(x+1\) is a divisor of 12. Let's list the divisors of 12:\par

1. The divisors of 12 are: \(\pm 1, \pm 2, \pm 3, \pm 4, \pm 6, \pm 12\).\par

2. We need to find the smallest \(x\), so we start with the smallest positive divisor, which is 1:
\[
x + 1 = 1 \implies x = 1 - 1 \implies x = 0
\]\par

3. To confirm, we substitute \(x = 0\) back into the original expression:
\[
\frac{12}{0+1} = \frac{12}{1} = 12
\]
Since 12 is an integer, \(x = 0\) is indeed a valid solution.\par

4. We should also consider the negative divisors to ensure we have the smallest possible integer value. The next smallest divisor is -1:
\[
x + 1 = -1 \implies x = -1 - 1 \implies x = -2
\]\par

5. Substituting \(x = -2\) back into the original expression:
\[
\frac{12}{-2+1} = \frac{12}{-1} = -12
\]
Since -12 is an integer, \(x = -2\) is also a valid solution, but it is not smaller than \(x = 0\).\par

Therefore, the smallest possible integer value of \(x\) is 
\[
\boxed{0}
\]
  };

  \node[
    draw=\LDraw, fill=\LFill,
    rounded corners=\LRadius, inner sep=\LSep,
     anchor=west,
    line width=\BLine, font=\LFont
  ] at ($(main.north west)+(\LShift,0)$) {\textbf{Vanilla GRPO}};

\end{tikzpicture}

\vspace{\baselineskip}

\begin{tikzpicture}
  \BannerThemeOurs
  \coordinate (C) at (\dimexpr.5\linewidth\relax,0);

  \node[
    draw=\BDraw, fill=\BFill,
    rounded corners=\BRadius, line width=\BLine,
    minimum width=\BWidth,
    text width=\dimexpr\BWidth - \BLine - 2\BannerSep\relax,
    inner sep=\BannerSep,
    align=left,
    font=\BFont
  ] (main) at (C) {
To find the smallest possible integer value of \(x\) such that \(\frac{12}{x+1}\) is an integer, we need to determine the divisors of 12. The divisors of 12 are \(\pm 1, \pm 2, \pm 3, \pm 4, \pm 6, \pm 12\). For \(\frac{12}{x+1}\) to be an integer, \(x+1\) must be one of these divisors. We want the smallest possible \(x\), so we should consider the smallest positive divisor, which is 1.\par

If \(x+1 = 1\), then \(x = 0\).\par

We should also consider the negative divisors to ensure we find the smallest possible integer value of \(x\). The smallest negative divisor is \(-12\). If \(x+1 = -12\), then \(x = -13\).\par

Comparing \(x = 0\) and \(x = -13\), the smallest possible integer value of \(x\) is \(-13\).\par

Therefore, the final answer is 
\[
\boxed{-13}
\]
};

  \node[
    draw=\LDraw, fill=\LFill,
    rounded corners=\LRadius, inner sep=\LSep,
     anchor=west,
    line width=\BLine, font=\LFont
  ] at ($(main.north west)+(\LShift,0)$) {\textbf{GRPO with Auxiliary Shaping Term}};
\end{tikzpicture}
    \caption{Case study: Vanilla GRPO fails to exploit numerical commonsense when comparing negative values (-1 vs. -13), while our method gives the correct answer.}
    \label{fig:pass_1_q1}
\end{figure*}

\subsection{Case Study For Pass@1 Setting}
\paragraph{Case Study $\mathrm{I}$.}
For the case in Fig.~\ref{fig:pass_1_q1}, vanilla GRPO incorrectly concludes that $-1<-13$, while GRPO with our auxiliary shaping term produces the correct comparison. This illustrates a weakness in vanilla GRPO’s exploitation: it does not reliably apply basic numerical reasoning, namely that among negative numbers, a larger absolute value implies a smaller number. Adding the auxiliary shaping term improves exploitation, making such comparisons more consistent.

\paragraph{Case Study $\mathrm{II}$.}
For the case in Fig.~\ref{fig:pass_1_q2}, the crucial distinction lies in how the models interpret the condition \emph{“No house faces two different streets”}. The vanilla GRPO–trained model incorrectly infers that each street contains only 10 houses, overlooking the structural implication of the condition. In contrast, the model trained with GRPO augmented by the auxiliary shaping term exploits this commonsense constraint more effectively: it decomposes the problem step by step, correctly deduces that each street has 20 houses, and thereby arrives at the correct total.

\begin{figure*}[!t]
    \centering
\begin{tikzpicture}
  \BannerThemeQuestion
  \coordinate (C) at (\dimexpr.5\linewidth\relax,0);

  \node[
    draw=\BDraw, fill=\BFill,
    rounded corners=\BRadius, line width=\BLine,
    minimum width=\BWidth,
    text width=\dimexpr\BWidth - \BLine - 2\BannerSep\relax,
    inner sep=\BannerSep,
    align=left,
    font=\BFont
  ] (main) at (C) {
In my neighborhood, there are six streets. There are 10 houses on each side of each street. No house faces two different streets. How many houses are in my neighborhood?
  };

  \node[
    draw=\LDraw, fill=\LFill,
    rounded corners=\LRadius, inner sep=\LSep,
     anchor=west,
    line width=\BLine, font=\LFont
  ] at ($(main.north west)+(\LShift,0)$) {\textbf{Question}};

\end{tikzpicture}

\vspace{\baselineskip}

\begin{tikzpicture}
  \BannerThemeAnswer
  \coordinate (C) at (\dimexpr.5\linewidth\relax,0);

  \node[
    draw=\BDraw, fill=\BFill,
    rounded corners=\BRadius, line width=\BLine,
    minimum width=\BWidth,
    text width=\dimexpr\BWidth - \BLine - 2\BannerSep\relax,
    inner sep=\BannerSep,
    align=left,
    font=\BFont
  ] (main) at (C) {120};

  \node[
    draw=\LDraw, fill=\LFill,
    rounded corners=\LRadius, inner sep=\LSep,
     anchor=west,
    line width=\BLine, font=\LFont
  ] at ($(main.north west)+(\LShift,0)$) {\textbf{Ground Truth}};

\end{tikzpicture}

\vspace{\baselineskip}

\begin{tikzpicture}
  \BannerThemeBaseline
  \coordinate (C) at (\dimexpr.5\linewidth\relax,0);

  \node[
    draw=\BDraw, fill=\BFill,
    rounded corners=\BRadius, line width=\BLine,
    minimum width=\BWidth,
    text width=\dimexpr\BWidth - \BLine - 2\BannerSep\relax,
    inner sep=\BannerSep,
    align=left,
    font=\BFont
  ] (main) at (C) {
To determine the total number of houses in the neighborhood, we can follow these steps:\par

1. \textbf{Identify the number of streets and houses per street:}
- There are 6 streets.
- Each street has 10 houses on each side.\par

2. \textbf{Calculate the number of houses on one street:}
Since there are 10 houses on each side of a street, and each house faces only one street, the total number of houses on one street is \(10\) (since they are all on the same street).\par

3. \textbf{Calculate the total number of houses in the neighborhood:}
There are 6 streets, and each street has 10 houses.
Therefore, the total number of houses is \(6 \times 10 = 60\).\par

Putting the final answer within $\boxed{}$, we get:
\[
\boxed{60}
\]
  };

  \node[
    draw=\LDraw, fill=\LFill,
    rounded corners=\LRadius, inner sep=\LSep,
     anchor=west,
    line width=\BLine, font=\LFont
  ] at ($(main.north west)+(\LShift,0)$) {\textbf{Vanilla GRPO}};

\end{tikzpicture}

\vspace{\baselineskip}

\begin{tikzpicture}
  \BannerThemeOurs
  \coordinate (C) at (\dimexpr.5\linewidth\relax,0);

  \node[
    draw=\BDraw, fill=\BFill,
    rounded corners=\BRadius, line width=\BLine,
    minimum width=\BWidth,
    text width=\dimexpr\BWidth - \BLine - 2\BannerSep\relax,
    inner sep=\BannerSep,
    align=left,
    font=\BFont
  ] (main) at (C) {
To find the total number of houses in the neighborhood, let's break down the problem step by step.\par

1. There are 6 streets in the neighborhood.\par
2. Each street has houses on both sides.\par
3. There are 10 houses on each side of every street.\par
4. No house faces two different streets, meaning each house is counted only once.\par

First, calculate the number of houses on one side of a single street:
\[
10 \ \text{houses/side}
\]\par

Since there are houses on both sides of each street, multiply by 2:
\[
10 \ \text{houses/side} \times 2 \ \text{sides/street} = 20 \ \text{houses/street}
\]\par

Now, since there are 6 streets, multiply the number of houses per street by the number of streets:
\[
20 \ \text{houses/street} \times 6 \ \text{streets} = 120 \ \text{houses}
\]\par

So, the total number of houses in the neighborhood is 
\[
\boxed{120}
\]
};

  \node[
    draw=\LDraw, fill=\LFill,
    rounded corners=\LRadius, inner sep=\LSep,
     anchor=west,
    line width=\BLine, font=\LFont
  ] at ($(main.north west)+(\LShift,0)$) {\textbf{GRPO with Auxiliary Shaping Term}};
\end{tikzpicture}
    \caption{Case study: Vanilla GRPO overlooks the constraint and infers 10 houses per street, while our method exploits it correctly and infers 20.}
    \label{fig:pass_1_q2}
\end{figure*}
\paragraph{Case Study $\mathrm{III}$.}
As shown in Fig.~\ref{fig:pass_1_q3}, the task concerns solving a differential equation under the constraint of finding a purely exponential solution rather than a general one. The model trained with vanilla GRPO overlooks this constraint and defaults to the generic procedure for obtaining the full solution. In contrast, the model trained with GRPO augmented by the auxiliary shaping term respects the constraint and produces the correct purely exponential answer. This case illustrates how the auxiliary shaping term enhances exploitation by enforcing adherence to problem-specific constraints.

\begin{figure*}[!t]
    \centering
\begin{tikzpicture}
  \BannerThemeQuestion
  \coordinate (C) at (\dimexpr.5\linewidth\relax,0);

  \node[
    draw=\BDraw, fill=\BFill,
    rounded corners=\BRadius, line width=\BLine,
    minimum width=\BWidth,
    text width=\dimexpr\BWidth - \BLine - 2\BannerSep\relax,
    inner sep=\BannerSep,
    align=left,
    font=\BFont
  ] (main) at (C) {
Find a purely exponential solution of $\frac{d^{4} x}{d t^{4}}-x=e^{-2 t}$.
  };

  \node[
    draw=\LDraw, fill=\LFill,
    rounded corners=\LRadius, inner sep=\LSep,
    anchor=west,
    line width=\BLine, font=\LFont
  ] at ($(main.north west)+(\LShift,0)$) {\textbf{Question}};

\end{tikzpicture}

\vspace{\baselineskip}

\begin{tikzpicture}
  \BannerThemeAnswer
  \coordinate (C) at (\dimexpr.5\linewidth\relax,0);

  \node[
    draw=\BDraw, fill=\BFill,
    rounded corners=\BRadius, line width=\BLine,
    minimum width=\BWidth,
    text width=\dimexpr\BWidth - \BLine - 2\BannerSep\relax,
    inner sep=\BannerSep,
    align=left,
    font=\BFont
  ] (main) at (C) {$\frac{e^{-2t}}{15}$};

  \node[
    draw=\LDraw, fill=\LFill,
    rounded corners=\LRadius, inner sep=\LSep,
     anchor=west,
    line width=\BLine, font=\LFont
  ] at ($(main.north west)+(\LShift,0)$) {\textbf{Ground Truth}};

\end{tikzpicture}

\vspace{\baselineskip}

\begin{tikzpicture}
  \BannerThemeBaseline
  \coordinate (C) at (\dimexpr.5\linewidth\relax,0);

  \node[
    draw=\BDraw, fill=\BFill,
    rounded corners=\BRadius, line width=\BLine,
    minimum width=\BWidth,
    text width=\dimexpr\BWidth - \BLine - 2\BannerSep\relax,
    inner sep=\BannerSep,
    align=left,
    font=\BFont
  ] (main) at (C) {
To find a purely exponential solution of the differential equation ($\frac{d^4 x}{d t^4} - x = e^{-2t}$), we will follow these steps:\par1. \textbf{Find the complementary solution by solving the homogeneous equation:}   \[ \frac{d^4 x}{d t^4} - x = 0\]   The characteristic equation is:\[   r^4 - 1 = 0   \]   Solving for $r$, we get:\[r^4 = 1 \implies r = \pm 1, \pm i   \]Therefore, the complementary solution is:   \[   x_c(t) = C_1 e^t + C_2 e^{-t} + C_3 \cos(t) + C_4 \sin(t)   \]2. \textbf{Find a particular solution using the method of undetermined coefficients:}\par   Since the right-hand side of the non-homogeneous equation is $e^{-2t}$, we assume a particular solution of the form:   \[   x_p(t) = A e^{-2t}   \]   Taking the fourth derivative of $x_p(t)$:\[ \frac{d^4}{d t^4} (A e^{-2t}) = A (-2)^4 e^{-2t} = 16A e^{-2t}\]   Substituting $x_p(t)$ and its fourth derivative into the original differential equation:\[   16A e^{-2t} - A e^{-2t} = e^{-2t}   \]   Simplifying, we get:   \[   (16A - A) e^{-2t} = e^{-2t} \implies 15A e^{-2t} = e^{-2t}  \]   Therefore:   \[   15A = 1 \implies A = \frac{1}{15}   \]   So, the particular solution is:   \[   x_p(t) = \frac{1}{15} e^{-2t}   \]3. \textbf{Combine the complementary and particular solutions:}\par   The general solution to the differential equation is:   \[   x(t) = x_c(t) + x_p(t) = C_1 e^t + C_2 e^{-t} + C_3 \cos(t) + C_4 \sin(t) + \frac{1}{15} e^{-2t}   \]Thus, the final answer is:\[\boxed{C_1 e^t + C_2 e^{-t} + C_3 \cos(t) + C_4 \sin(t) + \frac{1}{15} e^{-2t}}\]
  };

  \node[
    draw=\LDraw, fill=\LFill,
    rounded corners=\LRadius, inner sep=\LSep,
     anchor=west,
    line width=\BLine, font=\LFont
  ] at ($(main.north west)+(\LShift,0)$) {\textbf{Vanilla GRPO}};

\end{tikzpicture}
    \caption{Case study: Vanilla GRPO ignores the constraint of a purely exponential solution, while our method enforces it and yields the correct result.}
    \label{fig:pass_1_q3}
\end{figure*}

\begin{figure*}[!t]\ContinuedFloat
    \centering
\begin{tikzpicture}
  \BannerThemeOurs
  \coordinate (C) at (\dimexpr.5\linewidth\relax,0);

  \node[
    draw=\BDraw, fill=\BFill,
    rounded corners=\BRadius, line width=\BLine,
    minimum width=\BWidth,
    text width=\dimexpr\BWidth - \BLine - 2\BannerSep\relax,
    inner sep=\BannerSep,
    align=left,
    font=\BFont
  ] (main) at (C) {To find a purely exponential solution to the differential equation 
\[
\frac{d^4 x}{dt^4} - x = e^{-2t}
\], we will follow these steps:\par

1. \textbf{Find the complementary solution by solving the homogeneous equation:}
\[
\frac{d^4 x}{dt^4} - x = 0
\]
The characteristic equation is:
\[
r^4 - 1 = 0
\]
Solving for \(r\), we get:
\[
r^4 = 1 \implies r = \pm 1, \pm i
\]
Therefore, the complementary solution is:
\[
x_c(t) = C_1 e^t + C_2 e^{-t} + C_3 \cos(t) + C_4 \sin(t)
\]\par

2. \textbf{Find a particular solution using the method of undetermined coefficients:}
Since the right-hand side of the non-homogeneous equation is \(e^{-2t}\), we assume a particular solution of the form:
\[
x_p(t) = A e^{-2t}
\]
Taking the fourth derivative of \(x_p(t)\):
\[
\frac{d^4 x_p}{dt^4} = 16A e^{-2t}
\]
Substituting \(x_p(t)\) and its fourth derivative into the original differential equation:
\[
16A e^{-2t} - A e^{-2t} = e^{-2t}
\]
Simplifying, we get:
\[
15A e^{-2t} = e^{-2t}
\]
Solving for \(A\):
\[
15A = 1 \implies A = \frac{1}{15}
\]
Therefore, the particular solution is:
\[
x_p(t) = \frac{1}{15} e^{-2t}
\]\par

3. \textbf{Combine the complementary and particular solutions:}
The general solution to the differential equation is:
\[
x(t) = x_c(t) + x_p(t) = C_1 e^t + C_2 e^{-t} + C_3 \cos(t) + C_4 \sin(t) + \frac{1}{15} e^{-2t}
\]\par

Thus, the purely exponential solution is:
\[
\boxed{\frac{1}{15} e^{-2t}}
\]
};

  \node[
    draw=\LDraw, fill=\LFill,
    rounded corners=\LRadius, inner sep=\LSep,
     anchor=west,
    line width=\BLine, font=\LFont
  ] at ($(main.north west)+(\LShift,0)$) {\textbf{GRPO with Auxiliary Shaping Term}};

\end{tikzpicture}
    \caption{Case study: Vanilla GRPO ignores the constraint of a purely exponential solution, while our method enforces it and yields the correct result (continued).}
\end{figure*}

\subsection{Case Study For Pass@16 Setting}
\paragraph{Case Study $\mathrm{I}$.}
As shown in Fig.~\ref{fig:pass_k_q1_1}, the question is a simple elementary-level math problem requiring the integer solution \(x \geq 1, x \neq 2\). While the vanilla GRPO model consistently outputs 3, which is incorrect, the model trained with the auxiliary shaping term produces 1 in 5 out of 16 rollouts, the correct answer. This case highlights that the auxiliary shaping term enhances the model’s exploration ability, enabling it to consider a wider range of possibly valid solutions.

\begin{figure*}[!t]
  \centering
\begin{tikzpicture}
  \BannerThemeQuestion
  \coordinate (C) at (\dimexpr.5\linewidth\relax,0);

  \node[
    draw=\BDraw, fill=\BFill,
    rounded corners=\BRadius, line width=\BLine,
    minimum width=\BWidth,
    text width=\dimexpr\BWidth - \BLine - 2\BannerSep\relax,
    inner sep=\BannerSep,
    align=left,
    font=\BFont
  ] (main) at (C) {
Given $f(x) = \frac{\sqrt{x-1}}{x-2}$, what is the smallest possible integer value for $x$ such that $f(x)$ has a real number value?
  };

  \node[
    draw=\LDraw, fill=\LFill,
    rounded corners=\LRadius, inner sep=\LSep,
     anchor=west,
    line width=\BLine, font=\LFont
  ] at ($(main.north west)+(\LShift,0)$) {\textbf{Question}};

\end{tikzpicture}

\vspace{\baselineskip}

\begin{tikzpicture}
  \BannerThemeAnswer
  \coordinate (C) at (\dimexpr.5\linewidth\relax,0);

  \node[
    draw=\BDraw, fill=\BFill,
    rounded corners=\BRadius, line width=\BLine,
    minimum width=\BWidth,
    text width=\dimexpr\BWidth - \BLine - 2\BannerSep\relax,
    inner sep=\BannerSep,
    align=left,
    font=\BFont
  ] (main) at (C) {1};

  \node[
    draw=\LDraw, fill=\LFill,
    rounded corners=\LRadius, inner sep=\LSep,
     anchor=west,
    line width=\BLine, font=\LFont
  ] at ($(main.north west)+(\LShift,0)$) {\textbf{Ground Truth}};

\end{tikzpicture}
    \caption{Case study demonstrating improved exploration: the auxiliary shaping term enables the model to explore and reach the correct solution.}
    \label{fig:pass_k_q1_1}
\end{figure*}

\newlength{\LabelH} \setlength{\LabelH}{2.3em}
\newlength{\BHeight} \setlength{\BHeight}{\dimexpr 0.93\textheight - \LabelH \relax}
\newlength{\CellW}   \setlength{\CellW}{\dimexpr \BWidth * 24/100 \relax}
\newlength{\Hsep}    \setlength{\Hsep}{\dimexpr \BWidth * 1/75 \relax}
\newlength{\CellH}   \setlength{\CellH}{\dimexpr \BHeight * 99/200 \relax}
\newlength{\Vsep}    \setlength{\Vsep}{\dimexpr \BHeight * 1/100 \relax}

\tikzset{
  BannerCell/.style={
    draw=\BDraw, fill=\BFill,
    rounded corners=\BRadius, line width=\BLine,
    minimum width=\dimexpr \CellW - \BLine \relax, text width=\dimexpr \CellW - \BLine - 2\BannerSep \relax,
    minimum height=\dimexpr \CellH - \BLine \relax,
    inner sep=\BannerSep,
    align=left, font=\BFont,
    outer sep=0pt,
    anchor=north west
  }
}

\begin{figure*}[!t]\ContinuedFloat
  \centering
\begin{tikzpicture}
  \BannerThemeBaseline

  \node[
    draw=\LDraw, fill=\LFill,
    rounded corners=\BRadius,
    line width=\BLine,
    inner sep=\BannerSep, outer sep=0pt,
    anchor=north west, font=\LFont,
    minimum height=\LabelH,
    align=left
  ] (toplabel) at (0,0) {\textbf{Vanilla GRPO\#1-\#8}};
  \matrix (grid) [matrix of nodes,
    anchor=north west,
    nodes={BannerCell},
    column sep=\Hsep,  
    row sep=\Vsep,      
    inner sep=0pt
  ] at ([yshift=\dimexpr  - \LabelH - \Vsep\relax]0,0)
  {
    {
    \passkQaBSa
    } & 
    {
    \passkQaBSb
    } & 
    {
    \passkQaBSc
    } & 
    {
    \passkQaBSd
    } \\
    {
    \passkQaBSe
    } & 
    {
    \passkQaBSf
    } & 
    {
    \passkQaBSg
    } & 
    {
    \passkQaBSh
    } \\
  };
\end{tikzpicture}
  \caption{Case study demonstrating improved exploration: the auxiliary shaping term enables the model to explore and reach the correct solution (continued).}
\end{figure*}

\begin{figure*}[!t]\ContinuedFloat
  \centering
\begin{tikzpicture}
  \BannerThemeBaseline
  \node[
    draw=\LDraw, fill=\LFill,
    rounded corners=\BRadius,
    line width=\BLine,
    inner sep=\BannerSep, outer sep=0pt,
    anchor=north west, font=\LFont,
    minimum height=\LabelH,
    align=left
  ] (toplabel) at ([yshift=\topskip]0,0) {\textbf{Vanilla GRPO\#9-\#16}};
  \matrix (grid) [matrix of nodes,
    anchor=north west,
    nodes={BannerCell},
    column sep=\Hsep,  
    row sep=\Vsep,      
    inner sep=0pt
  ] at ([yshift=\dimexpr \topskip - \LabelH - \Vsep\relax]0,0)
  {
    {
    \passkQaBSi
    } & 
    {
    \passkQaBSj
    } & 
    {
    \passkQaBSk
    } & 
    {
    \passkQaBSl
    } \\
    {
    \passkQaBSm
    } & 
    {
    \passkQaBSn
    } & 
    {
    \passkQaBSo
    } & 
    {
    \passkQaBSp
    } \\
  };
\end{tikzpicture}
  \caption{Case study demonstrating improved exploration: the auxiliary shaping term enables the model to explore and reach the correct solution (continued).}
\end{figure*}

\begin{figure*}[!t]\ContinuedFloat
  \centering
\begin{tikzpicture}
  \BannerThemeOurs
  \node[
    draw=\LDraw, fill=\LFill,
    rounded corners=\BRadius,
    line width=\BLine,
    inner sep=\BannerSep, outer sep=0pt,
    anchor=north west, font=\LFont,
    minimum height=\LabelH,
    align=left
  ] (toplabel) at (0,0) {\textbf{GRPO with Auxiliary Shaping Term\#1-\#8}};
  \matrix (grid) [matrix of nodes,
    anchor=north west,
    nodes={BannerCell},
    column sep=\Hsep,  
    row sep=\Vsep,      
    inner sep=0pt
  ] at ([yshift=\dimexpr  - \LabelH - \Vsep\relax]0,0)
  {
    {
    \passkQaOURa
    } & 
    {
    \passkQaOURb
    } & 
    {
    \passkQaOURc
    } & 
    {
    \passkQaOURd
    } \\
    {
    \passkQaOURe
    } & 
    {
    \passkQaOURf
    } & 
    {
    \passkQaOURg
    } & 
    {
    \passkQaOURh
    } \\
  };
\end{tikzpicture}
  \caption{Case study demonstrating improved exploration: the auxiliary shaping term enables the model to explore and reach the correct solution (continued).}
\end{figure*}

\begin{figure*}[!t]\ContinuedFloat
  \centering
\begin{tikzpicture}
  \BannerThemeOurs
  \node[
    draw=\LDraw, fill=\LFill,
    rounded corners=\BRadius,
    line width=\BLine,
    inner sep=\BannerSep, outer sep=0pt,
    anchor=north west, font=\LFont,
    minimum height=\LabelH,
    align=left
  ] (toplabel) at ([yshift=\topskip]0,0) {\textbf{GRPO with Auxiliary Shaping Term\#9-\#16}};
  \matrix (grid) [matrix of nodes,
    anchor=north west,
    nodes={BannerCell},
    column sep=\Hsep,  
    row sep=\Vsep,      
    inner sep=0pt
  ] at ([yshift=\dimexpr \topskip - \LabelH - \Vsep\relax]0,0)
  {
    {
    \passkQaOURi
    } & 
    {
    \passkQaOURj
    } & 
    {
    \passkQaOURk
    } & 
    {
    \passkQaOURl
    } \\
    {
    \passkQaOURm
    } & 
    {
    \passkQaOURn
    } & 
    {
    \passkQaOURo
    } & 
    {
    \passkQaOURp
    } \\
  };
\end{tikzpicture}
  \caption{Case study demonstrating improved exploration: the auxiliary shaping term enables the model to explore and reach the correct solution (continued).}
\end{figure*}

\paragraph{Case Study $\mathrm{II}$.}
As shown in Fig.~\ref{fig:pass_k_q2}, this high-school level geometry problem illustrates the contrast between vanilla GRPO and GRPO with the auxiliary shaping term. The vanilla GRPO model fails to produce the correct answer in any rollout. In contrast, the model trained with the auxiliary shaping term succeeds in 5 out of 16 rollouts. Moreover, its reasoning traces display higher exploratory diversity: it applies the law of sines 9 times, the law of cosines 4 times, and other general geometric properties 3 times. By comparison, the vanilla GRPO model relies more narrowly on the law of sines (11 times), with only limited use of other strategies. This case highlights that the auxiliary shaping term enhances the model’s exploration capacity, enabling it to leverage a broader range of problem-solving strategies rather than over-committing to a single path.

\setlength{\BHeight}{\dimexpr 0.78\textheight - \LabelH \relax}
\setlength{\CellW}{\dimexpr \BWidth * 24/100 \relax}
\setlength{\Hsep}{\dimexpr \BWidth * 1/75 \relax}
\setlength{\CellH}{\dimexpr \BHeight * 99/200 \relax}
\setlength{\Vsep}{\dimexpr \BHeight * 1/100 \relax}

\begin{figure*}[!t]
  \centering
\begin{tikzpicture}
  \BannerThemeQuestion
  \coordinate (C) at (\dimexpr.5\linewidth\relax,0);

  \node[
    draw=\BDraw, fill=\BFill,
    rounded corners=\BRadius, line width=\BLine,
    minimum width=\BWidth,
    text width=\dimexpr\BWidth - \BLine - 2\BannerSep\relax,
    inner sep=\BannerSep,
    align=left,
    font=\BFont
  ] (main) at (C) {
Abdul and Chiang are standing 48 feet apart in a field. Bharat is standing in the same field as far from Abdul as possible so that the angle formed by his lines of sight to Abdul and Chiang measures $60^\circ$. What is the square of the distance (in feet) between Abdul and Bharat?
  };

  \node[
    draw=\LDraw, fill=\LFill,
    rounded corners=\LRadius, inner sep=\LSep,
     anchor=west,
    line width=\BLine, font=\LFont
  ] at ($(main.north west)+(\LShift,0)$) {\textbf{Question}};

\end{tikzpicture}

\vspace{\baselineskip}

\begin{tikzpicture}
  \BannerThemeAnswer
  \coordinate (C) at (\dimexpr.5\linewidth\relax,0);

  \node[
    draw=\BDraw, fill=\BFill,
    rounded corners=\BRadius, line width=\BLine,
    minimum width=\BWidth,
    text width=\dimexpr\BWidth - \BLine - 2\BannerSep\relax,
    inner sep=\BannerSep,
    align=left,
    font=\BFont
  ] (main) at (C) {3072};

  \node[
    draw=\LDraw, fill=\LFill,
    rounded corners=\LRadius, inner sep=\LSep,
     anchor=west,
    line width=\BLine, font=\LFont
  ] at ($(main.north west)+(\LShift,0)$) {\textbf{Ground Truth}};

\end{tikzpicture}

\vspace{\baselineskip}

\begin{tikzpicture}
  \BannerThemeBaseline

  \node[
    draw=\LDraw, fill=\LFill,
    rounded corners=\BRadius,
    line width=\BLine,
    inner sep=\BannerSep, outer sep=0pt,
    anchor=north west, font=\LFont,
    minimum height=\LabelH,
    align=left
  ] (toplabel) at (0,0) {\textbf{Vanilla GRPO\#1-\#8}};
  \matrix (grid) [matrix of nodes,
    anchor=north west,
    nodes={BannerCell},
    column sep=\Hsep,  
    row sep=\Vsep,      
    inner sep=0pt
  ] at ([yshift=\dimexpr  - \LabelH - \Vsep\relax]0,0)
  {
    {
    \passkQbBSm
    } & 
    {
    \passkQbBSb
    } & 
    {
    \passkQbBSc
    } & 
    {
    \passkQbBSp
    } \\
    {
    \passkQbBSe
    } & 
    {
    \passkQbBSo
    } & 
    {
    \passkQbBSk
    } & 
    {
    \passkQbBSh
    } \\
  };
\end{tikzpicture}
  \caption{Case study: Geometry problem. GRPO with auxiliary shaping term not only achieves higher accuracy but also explores more diverse reasoning strategies than vanilla GRPO.}
  \label{fig:pass_k_q2}
\end{figure*}

\setlength{\BHeight}{\dimexpr 0.93\textheight - \LabelH \relax}
\setlength{\CellW}{\dimexpr \BWidth * 24/100 \relax}
\setlength{\Hsep}{\dimexpr \BWidth * 1/75 \relax}
\setlength{\CellH}{\dimexpr \BHeight * 99/200 \relax}
\setlength{\Vsep}{\dimexpr \BHeight * 1/100 \relax}

\begin{figure*}[!t]\ContinuedFloat
  \centering
\begin{tikzpicture}
  \BannerThemeBaseline
  \node[
    draw=\LDraw, fill=\LFill,
    rounded corners=\BRadius,
    line width=\BLine,
    inner sep=\BannerSep, outer sep=0pt,
    anchor=north west, font=\LFont,
    minimum height=\LabelH,
    align=left
  ] (toplabel) at (0,0) {\textbf{Vanilla GRPO\#9-\#16}};
  \matrix (grid) [matrix of nodes,
    anchor=north west,
    nodes={BannerCell},
    column sep=\Hsep,  
    row sep=\Vsep,      
    inner sep=0pt
  ] at ([yshift=\dimexpr  - \LabelH - \Vsep\relax]0,0)
  {
    {
    \passkQbBSi
    } & 
    {
    \passkQbBSj
    } & 
    {
    \passkQbBSg
    } & 
    {
    \passkQbBSl
    } \\
    {
    \passkQbBSa
    } & 
    {
    \passkQbBSn
    } & 
    {
    \passkQbBSf
    } & 
    {
    \passkQbBSd
    } \\
  };
\end{tikzpicture}
  \caption{Case study: Geometry problem. GRPO with auxiliary shaping term not only achieves higher accuracy but also explores more diverse reasoning strategies than vanilla GRPO (continued).}
\end{figure*}

\begin{figure*}[!t]\ContinuedFloat
  \centering
\begin{tikzpicture}
  \BannerThemeOurs

  \node[
    draw=\LDraw, fill=\LFill,
    rounded corners=\BRadius,
    line width=\BLine,
    inner sep=\BannerSep, outer sep=0pt,
    anchor=north west, font=\LFont,
    minimum height=\LabelH,
    align=left
  ] (toplabel) at (0,0) {\textbf{GRPO with Auxiliary Shaping Term\#1-\#8}};
  \matrix (grid) [matrix of nodes,
    anchor=north west,
    nodes={BannerCell},
    column sep=\Hsep,  
    row sep=\Vsep,      
    inner sep=0pt
  ] at ([yshift=\dimexpr  - \LabelH - \Vsep\relax]0,0)
  {
    {
    \passkQbOURa
    } & 
    {
    \passkQbOURb
    } & 
    {
    \passkQbOURc
    } & 
    {
    \passkQbOURd
    } \\
    {
    \passkQbOURe
    } & 
    {
    \passkQbOURf
    } & 
    {
    \passkQbOURg
    } & 
    {
    \passkQbOURh
    } \\
  };
\end{tikzpicture}
  \caption{Case study: Geometry problem. GRPO with auxiliary shaping term not only achieves higher accuracy but also explores more diverse reasoning strategies than vanilla GRPO (continued).}
\end{figure*}

\begin{figure*}[!t]\ContinuedFloat
  \centering
\begin{tikzpicture}
  \BannerThemeOurs
  \node[
    draw=\LDraw, fill=\LFill,
    rounded corners=\BRadius,
    line width=\BLine,
    inner sep=\BannerSep, outer sep=0pt,
    anchor=north west, font=\LFont,
    minimum height=\LabelH,
    align=left
  ] (toplabel) at (0,0) {\textbf{GRPO with Auxiliary Shaping Term\#9-\#16}};
  \matrix (grid) [matrix of nodes,
    anchor=north west,
    nodes={BannerCell},
    column sep=\Hsep,  
    row sep=\Vsep,      
    inner sep=0pt
  ] at ([yshift=\dimexpr  - \LabelH - \Vsep\relax]0,0)
  {
    {
    \passkQbOURi
    } & 
    {
    \passkQbOURj
    } & 
    {
    \passkQbOURk
    } & 
    {
    \passkQbOURl
    } \\
    {
    \passkQbOURm
    } & 
    {
    \passkQbOURn
    } & 
    {
    \passkQbOURo
    } & 
    {
    \passkQbOURp
    } \\
  };
\end{tikzpicture}
  \caption{Case study: Geometry problem. GRPO with auxiliary shaping term not only achieves higher accuracy but also explores more diverse reasoning strategies than vanilla GRPO (continued).}
\end{figure*}

\end{document}